\pdfoutput=1

\documentclass[11pt]{article}

\usepackage[preprint]{acl}

\usepackage{times}
\usepackage{latexsym}

\usepackage[T1]{fontenc}

\usepackage[utf8]{inputenc}

\usepackage{microtype}

\usepackage{inconsolata}

\usepackage{booktabs}
\usepackage{multirow}
\usepackage{mathtools}
\mathtoolsset{showonlyrefs}
\usepackage{amssymb}

\usepackage{listings}
\usepackage{xcolor}
\lstdefinelanguage{json}{
  morestring=[b]",
  morecomment=[l]{//},
  literate=
   *{0}{{{\color{black}0}}}{1}
    {1}{{{\color{black}1}}}{1}
    {2}{{{\color{black}2}}}{1}
    {3}{{{\color{black}3}}}{1}
    {4}{{{\color{black}4}}}{1}
    {5}{{{\color{black}5}}}{1}
    {6}{{{\color{black}6}}}{1}
    {7}{{{\color{black}7}}}{1}
    {8}{{{\color{black}8}}}{1}
    {9}{{{\color{black}9}}}{1}
    {:}{{{\color{black}:}}}{1}
    {,}{{{\color{black},}}}{1}
    {\{}{{{\color{black}\{}}}{1}
    {\}}{{{\color{black}\}}}}{1}
    {[}{{{\color{black}[}}}{1}
    {]}{{{\color{black}]}}}{1},
}
\usepackage{graphicx}

\newcommand{\opti}{\textsc{Optimization}\xspace}
\newcommand{\spread}{\textsc{Spread}\xspace}
\newcommand{\underedit}{UnderEdit\xspace}
\newcommand{\overedit}{OverEdit\xspace}

\usepackage{relsize}
\usepackage{multirow}
\usepackage{amsmath}
\usepackage{amsfonts}
\usepackage{xspace}

\title{Resolving \underedit \& \overedit with \\ Iterative \& Neighbor-Assisted Model Editing}

\author{
    Bhiman Kumar Baghel \quad
    Emma Jordan \quad
    Zheyuan Ryan Shi \quad
    Xiang Lorraine Li \\
    Department of Computer Science, University of Pittsburgh, PA, USA \\
    \texttt{\{bkb45, xianglli\}@pitt.edu}
}

\begin{document}
\maketitle
\begin{abstract}
Large Language Models (LLMs) are widely deployed in downstream tasks, but keeping their knowledge up-to-date via retraining or fine-tuning is often computationally expensive.
Model editing provides a more efficient alternative by updating a targeted subset of parameters, which often follows the locate-and-edit paradigm.
Despite this efficiency, existing methods are limited: edits may fail to inject knowledge (\underedit) or unintentionally disrupt unrelated neighboring knowledge (\overedit).
To address these challenges, we propose two complementary methods: \textbf{iterative model editing}, which applies successive edits to mitigate \underedit, and \textbf{neighbor-assisted model editing}, which incorporates neighboring knowledge during editing to reduce \overedit.
Our extensive experiments show that these techniques improve editing performance across multiple LLMs, algorithms, and benchmarks, reducing \underedit by up to 38 percentage points and \overedit by up to 6, while remaining broadly applicable to any locate-and-edit method. We release our code at \url{https://github.com/bhimanbaghel/ResolveUnderOverEdit}.
\end{abstract}

\section{Introduction}
\label{intro}
\begin{figure}[ht]
    \centering
    \includegraphics[width=\columnwidth]{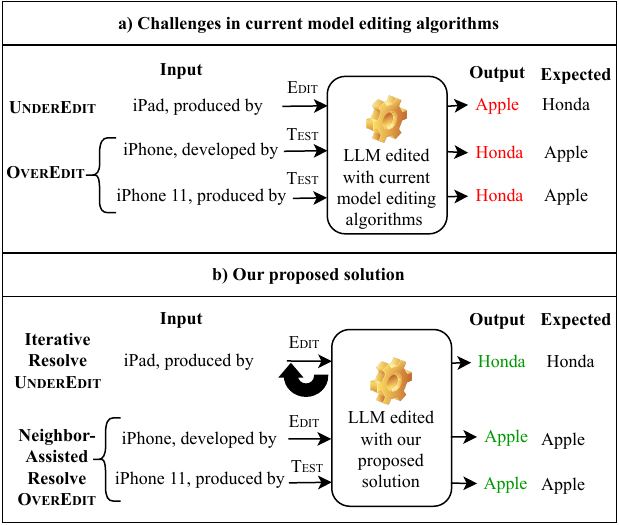}
    \caption{
    The example from C{\smaller OUNTER}F{\smaller ACT}
    updates \textit{iPad} producer from \textit{Apple} to \textit{Honda}. \underedit fails to make the desired update in the \textsc{Edit} sentence, while \overedit introduces the undesired change in the \textsc{Test} sentences as shown in (a). 
    The proposed iterative model editing mitigated \underedit and neighbor-assisted model editing reduced \overedit by incorporating related knowledge in \textsc{edit} stage as shown in (b).
    }
    \label{fig:edit-example}
\end{figure}
LLMs have been widely used as repositories of factual and specialized knowledge \citep{petroni2020context, jiang-etal-2021-know, roberts-etal-2020-much, youssef-etal-2023-give}. However, the world is constantly changing, with knowledge and information evolving rapidly, such as significant government policy changes and their wide impacts across various domains. Thus, it is essential for many NLP applications, such as text generation, question answering, and knowledge retrieval, to have models that can adapt to knowledge changes both effectively and efficiently. Re-training an LLM is resource-intensive \citep{patterson2021carbon}. Standard supervised fine-tuning is data hungry and less effective~\citep{meng2023memit}. 
Model-editing, which directly modifies important model parameters for making the prediction, has emerged as a more efficient alternative for updating outdated information~\citep{meng2022locating, meng2023memit, Li_Li_Song_Yang_Ma_Yu_2024, fang2025alphaedit}.
These methods adopt a ``locate-and-edit'' approach, where they first identify the parameter locations associated with outdated knowledge and then update the parameters to enable the model to incorporate and predict the new knowledge.

The effectiveness of the methods is evaluated from two perspectives. The first is whether the method successfully updates the knowledge, failure on this leaves certain facts unedited, causing \underedit. Secondly, whether the update introduces unintended modifications to neighboring knowledge — a phenomenon we call \overedit. Existing methods suffer from both \underedit and \overedit as shown in Figure~\ref{fig:edit-example}.

To address this, we propose methods to mitigate both \underedit and \overedit. For \underedit, we hypothesize that the parameter update is insufficient to achieve the desired knowledge change. The editing process performed a rank-one update on the layer parameters to achieve the desired update. We empirically showed that the approximation introduces errors, leading to \underedit.
To this end, we proposed iterative model editing, wherein editing is performed multiple times. For \overedit,
we hypothesize that model editing can benefit from including neighboring knowledge during the editing stage. We thus introduce neighbor-assisted model editing, a procedure that integrates neighboring knowledge during the editing process to keep the test neighboring knowledge unchanged. 

In summary, we propose solutions to two fundamental challenges in model editing: \underedit, where edits fail, and \overedit, where neighboring knowledge is erroneously modified. We evaluate our approach using four “locate and edit” model editing algorithms, ROME~\citep{meng2022locating}, MEMIT~\citep{meng2023memit}, PMET~\citep{Li_Li_Song_Yang_Ma_Yu_2024} and AlphaEdit~\citep{fang2025alphaedit}, and applied to four LLMs: GPT-2 XL (1.5B) \citep{radford2019language}, GPT-J (6B) \citep{gpt-j}, Llama-2 (7B) \citep{touvron2023llama2openfoundation}, and Llama-3.1 (8B) \citep{llama3_8b}. Our experiments are conducted on two widely used factual knowledge editing benchmarks: C{\smaller OUNTER}F{\smaller ACT} \citep{meng2022locating} and ZsRE \citep{DBLP:journals/corr/LevySCZ17}. Our results show that iterative model editing improves edit success while also reducing the approximation error introduced by the rank-one update. Furthermore, we demonstrate that incorporating even a single neighboring knowledge during model editing reduces unintended modifications to neighboring knowledge at test time, resulting in stronger edit performance. Overall, our proposed methods are broadly applicable and consistently effective across current locate-and-edit approaches that adopt the two-stage editing framework, and are readily applicable to future methods built on this foundation.

\section{The Locate-and-Edit Framework}
\label{sec:prelim}
In this section, we provide background on the locate-and-edit model editing framework along with the notation used throughout the paper.

An autoregressive LLM is a function $f_\theta \colon \mathcal{X}^T \to \Delta(\mathcal{X})$, that takes as input a sequence of tokens $x = (x_1,x_2,\cdots,x_T)$ of length $T$ with $x_i$ in the dictionary $\mathcal{X}$, and uses model parameters $\theta$ to return a probability distribution $\Delta(\mathcal{X})$ to model the next token $x'$, i.e,. $f_\theta(x)[{x'}] \approx \Pr(X'=x'|X=x)$, where $X$ and $X'$ are random variables representing the sequence of input tokens and the next token, respectively. 
The internal computations of an LLM relies on a grid of hidden states  $h_{t}^{l}$, where $l$ corresponds to the layer and $t$ corresponds to the token position in the sequence (using tokens $x_1,x_2,\cdots,x_t$). Each layer is a standard transformer block with the self-attention module, MLP module, etc \citep{vaswani2017attention}.

Prior work focuses on editing the factual knowledge within the LLM. Factual knowledge is represented as a triplet $(s, r, o)$, where subject $s \in \mathcal{X}^{T_s}$, relation $r \in \mathcal{X}^{T_r}$, and object $o \in \mathcal{X}$ are sequences of tokens, e.g., The \textit{iPad} $[s]$ is \textit{produced by} $[r]$ \textit{Apple} $[o]$. We consider only single token objects in this representation, following previous work~\cite{meng2022locating, meng2023memit, Li_Li_Song_Yang_Ma_Yu_2024}.
The model editing task is to make the model place a higher likelihood on a new object $o^*$ than an old object $o$ when presented with $x = (s,r)$, i.e., find new parameters $\theta'$, such that $f_{\theta'}(x)[o^*] > f_{\theta'}(x)[o]$. Model editing is not limited to a single edit, but can encompass a batch of $m$ desired edits $D=\{(s_i,r_i,o_i,o^*_i)\}_{i=1}^m$.

Locate and edit model editing algorithms hypothesize that factual knowledge locates within specific layers of the LLM, and updating parameters in these layers is sufficient to induce the desired change in object \citep{pearl2013direct, NEURIPS2020_92650b2e, meng2022locating}. These methods employ causal tracing~to identify these layers responsible for the factual knowledge, referred to as causal layers $\{l_1, \ldots, l_c\}$, where $c$ denotes the number of layers in this range.
The last MLP in these layers has been found to have a major impact on the object token distribution when presented with the subject tokens \citep{meng2022locating, meng2023memit, geva-etal-2021-transformer}.
Due to this major impact, locate and edit methods focus on only updating these MLP weights. 

The weight update is performed in two stages: \opti stage finds ideal values for the network's hidden state in certain transformer layer to make $o^*$ likely and \spread stage updates the weights of the last MLP in casual layer(s) to approximate this ideal hidden state. We detail these stages below for MEMIT~\citep{meng2023memit} and discuss its differences to PMET~\citep{Li_Li_Song_Yang_Ma_Yu_2024}, AlphaEdit~\citep{fang2025alphaedit} and ROME~\citep{meng2022locating}. 
\begin{table*}[t]
\centering
\resizebox{\textwidth}{!}{%
\begin{tabular}{@{}l|l|l@{}}
\toprule
\textbf{Algorithm} & \textbf{\opti Stage} & \textbf{\spread Stage} \\ 
\midrule
\textbf{MEMIT~\citep{meng2023memit}}      & $h^{l_c}$ (hidden state) & $W^{l_1} \cdots W^{l_c}$ (least-squares update) \\
\textbf{PMET~\citep{Li_Li_Song_Yang_Ma_Yu_2024}}       & $\textit{Attn}^{l_c}$ \& $\textit{z}^{l_c}$ (ideal attention + MLP) & $W^{l_1} \cdots W^{l_c}$ (attention-free update) \\
\textbf{AlphaEdit~\citep{fang2025alphaedit}}  & $h^{l_c}$ (projected to null space) & $W^{l_1} \cdots W^{l_c}$ (null space-constrained update) \\
\textbf{ROME~\citep{meng2022locating}}       & $z^{l_*}$ (ideal value) & $W^{l_*}$ (rank-one equality-constrained update) \\
\textbf{R-ROME~\citep{gupta-etal-2024-rebuilding}}     & $z^{l_*}$, $k_*$ (averaged over context) & $W^{l_*}$ (stabilized rank-one update) \\
\textbf{EMMET~\citep{gupta-etal-2024-unified}}      & $h^{l_c}$ (batched hidden states) & $W^{l_1} \cdots W^{l_c}$ (equality-constrained batch update) \\
\textbf{ENCORE~\citep{gupta2025lifelongsequentialknowledgeediting}}     & $h^{l_c}$ (with MPES early stopping) & $W^{l_1} \cdots W^{l_c}$ (norm-constrained update) \\
\textbf{EVOKE (LTI)~\cite{zhang2025uncovering}}& $h^{l_c}$ (constrained via multi-stage loss) & Follows base method (e.g., ROME or MEMIT) \\
\bottomrule
\end{tabular}
}
\caption{We present a unifying overview of existing locate-and-edit algorithms within the two-stage framework of \opti and \spread. This abstraction allows our proposed methods—iterative and neighbor-assisted editing—to be applied broadly across all listed algorithms, as well as to future methods built upon this framework. For detailed descriptions of each algorithm, see Appendix~\ref{sec:MPR}.}
\label{tab:algos}
\end{table*}
\paragraph{\opti Stage: Learning the Ideal State.}
The goal of the \opti stage is to find what outputs in the causal layers would lead to a high likelihood on $o^*$. The methods we investigate search for ideal outputs in different locations. MEMIT searches for an ideal output, $\bar h^{l_c}_{t}$, for the last casual layer at $t$, the last token index of the subject $s$. The search is performed by finding a vector $\delta$ to add to current hidden state value $h^{l_c}_{t}$.
We represent the output probability distribution of the model using $h^{l_c}_t$ and $\delta$ as $f_\theta(x,h^{l_c}_t+\delta)$.

To make the hidden state $\delta$ change robust to diverse contexts, these methods add a random prefix to the prompt, i.e., the network takes as input $x_i = (\xi_i,s,r$), where $\xi_i$ is one of $n$ random prefixes. The loss function for $\delta$ is to minimize the average negative log likelihood of $o^*$, i.e.,

\begin{align}
g(\delta) \dot = &\ -\frac{1}{n}\sum_{i=1}^n  \ln f_\theta \left (x_i,h^{l_c}_t+\delta \right )[o^*] \\
&\ + D_\text{KL} \left (f_\theta \left ( s, h^{l_c}_t+\delta \right ) || f_\theta \left (s \right ) \right ) 
\end{align}

where $D_\text{KL}$ is the Kullback–Leibler divergence, which is added to constrain the model's output to be close to the original. 

The ideal hidden state for the prompt $x=(s,r)$ is $\bar h_t^{l_c} = h_t^{l_c} + \delta^*$, where $\delta^*$ is found by performing gradient descent on $g$. This ideal hidden state is then used in computing weight update in the next stage. AlphaEdit is the same as MEMIT in this stage, whereas PMET differs from MEMIT by searching for ideal outputs for the attention module and MLP modules in layer $l_c$. ROME searches for an ideal output for the MLP module of a single layer in the set of causal layers.  



\begin{figure}[t]
    \centering



    \includegraphics[width=\columnwidth]{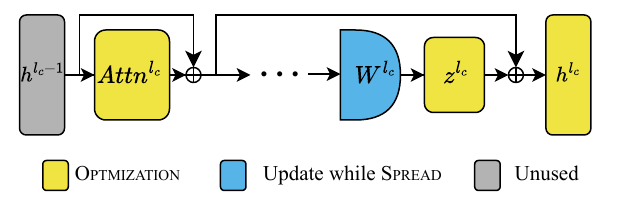}

    \caption{
    The diagram shows a simplified transformer layer to complement Table~\ref{tab:algos}, composed of attention and MLP modules. Only the last MLP is shown, as all methods modify its parameters.
    }
    \label{fig:algodiff}
\end{figure}

\paragraph{\spread Stage: Propagating the Change.}

The goal of the \spread stage is to find new weights $\theta'$ such that the hidden state after the update $\hat h^{l_c}_t$ is close to the ideal hidden state $\bar h^{l_c}_t$ for all desired edits in $D$. 
Not all weights in the network are updated, only the weights $W^{l}$ corresponding to the weights of the last MLP layer in causal layers are updated. 
The weight update methods are derived from a rank-one approximation to make $\hat h^{l_c}_t \approx \bar h^{l_c}_t$, which can lead to failure. 
The different algorithms update different set of weights: MEMIT, PMET, and AlphaEdit update $W^{l}$ for all causal layers, whereas ROME only updates one $W^{l}$. Among these, AlphaEdit differs by computing $\theta'$ using the null space projection method \citep{Wang_2021_CVPR}.
The method differences are shown in Table~\ref{tab:algos}, with Figure \ref{fig:algodiff} complementing it.

Building on the preceding discussion of the primary methods, several additional locate-and-edit algorithms are instances of the same two-stage framework: R-ROME~\citep{gupta-etal-2024-rebuilding}, which conditions on more context prompts than ROME to improve update stability; EMMET~\citep{gupta-etal-2024-unified}, which extends ROME to batch edits; EVOKE (LTI)\citep{zhang2025uncovering}, which adds constraints in the \opti stage to regularize the latent search; and ENCORE\citep{gupta2025lifelongsequentialknowledgeediting}, which uses early stopping in \opti to limit overconfident hidden states and normalized updates in \spread to control update magnitude, thereby mitigating overconfident editing and instability. Table~\ref{tab:algos} provides an overview of how these methods implement the two stages, and Appendix~\ref{sec:MPR} details the algorithms. We next examine inherent limitations of this two-stage framework and propose a general, method-agnostic remedy that applies broadly across such approaches.

\section{Method}
The memory-editing algorithms mentioned above face challenges, such as failing to edit certain knowledge i.e., \underedit or changing neighbor knowledge that should remain unchanged i.e., \overedit. In this section, we present our proposed method to address these issues. Specifically, we introduce iterative model editing (\ref{sec:iterative-edit}) to mitigate \underedit and neighbor-assisted model editing (\ref{sec:neighbor-edit}) to reduce \overedit.

\subsection{Iterative Model Editing}
\label{sec:iterative-edit}
There are two possible reasons for \underedit to occur. The first is that $\bar h^{l_c}_t$ does not reflect a hidden state for a successful edit. The second is that the weight update results in $\hat h^{l_c}_t \not \approx \bar h^{l_c}_t$. We hypothesize that both of these potential problems can be addressed by running the memory edit process multiple times because: 1) it allows for potentially finding better $\bar h^{l_c}_t$ after updating the model parameters so that $\|\hat h_t^{l_c} - \bar h^{l_c}_t\| \le \|h^{l_c}_t - \bar h^{l_c}_t \|$, and 2) on the next iteration, the approximation used in the \spread stage for the weight update will be better since $\hat h^t_{l_c}$ is closer to $\bar h_{t}^{l_c}$ than $h_t^{l_c}$.
Our approach involves whole-pipeline iteration—executing \opti followed by \spread across multiple iterations until convergence. In contrast, editing algorithms (Table~\ref{tab:algos}) perform optimization only once within the \opti stage using gradient descent, which finds the target representation but does not by itself yield a better edit. Our proposed iterative refinement ensures that both the target representation and the subsequent parameter update improve in tandem.
We detail this iterative process below for MEMIT, but it can also be adapted to ROME, PMET and AlphaEdit by replacing $\bar h_{t}^{l_c}$ with the targets of the optimization procedure for those algorithms.

Iterative model editing works as follows. 
At iteration $k$,
\opti computes the ideal hidden state $\bar h_{t,k}^{l_c}$ based on the hidden state produced with the model parameters $\theta_k$, i.e., $\bar h_{t,k}^{l_c} = h_{t,k}^{l_c} + \delta_k^*$, where $\delta_k^*$ is obtained by optimizing $g(\delta)$ using $\theta_k$ as the model parameters. 
\spread stage updates the model parameters to $\theta_{k+1}$ based on the computed $\bar h_{t,k}^{l_c}$, producing a new hidden state $\hat h_{t,k}^{l_c}$. Note that $\hat h_{t,k}^{l} = h_{t,k+1}^{l}$.

The iterations end when model perplexity using $\hat h_{t,k}^{l_c}$ is within $\epsilon$ of the perplexity using $\bar h_{t,k}^{l_c}$, i.e.,
\begin{align}
    |p(\theta_{k+1}, \hat h^{l_c}_{t,k}) - p(\theta_k, \bar h_{t,k}^{l_c})| \le  \epsilon,
\end{align}
where $p$ is the perplexity of the target token over the $m$ edits in $D$ 
\begin{align}
    p(\theta, h) \dot=  \frac{1}{m}\sum_{i=1}^n e^{-\ln f_{\theta}(x_i, h)[o_i^*]}. 
\end{align}
For brevity, we use $\Delta p_k$ to denote the above difference in perplexity in iteration $k$. Empirically, we found $\epsilon=1$ to be a sufficient threshold for the data sets used in this paper. We set $\epsilon=1$ because, across all datasets considered, this threshold consistently delivered the best (or statistically indistinguishable) performance while minimizing the number of iterations (Table~\ref{tab:mcf-results}). Table \ref{tab:mcf-results-all} further justifies this choice, showing that smaller thresholds yield negligible gains at substantially higher compute cost, whereas larger thresholds can degrade accuracy.

The algorithm aims to reduce the perplexity difference between the two stages mainly because we found a mismatch between these two perplexity values. The \opti stage seeks to learn a hidden representation $\bar h_{t,k}^{l_c}$ that minimizes the perplexity of the target token, however, we found that after \spread this perplexity increases, leading to \underedit (as shown in Table \ref{tab:mcf-results-all} and \ref{tab:neighbor-all}). Thus we directly minimize this discrepancy and thereby mitigate \underedit. We provide further empirical verification of this relation in Appendix~\ref{sec:analysis_underedit}.

Figure \ref{fig:perplexity-example} illustrates how iterative model editing progressively brings $\hat h^{l_c}_{t,k}$ closer in perplexity to $\bar h_{t,k}^{l_c}$. The figure also highlights that most of the improvement stems from applying \spread multiple times, as the perplexity of $\bar h_{t,k}^{l_c}$ changes relatively little across iterations. However, the stability in perplexity does not imply that $\bar h_{t,k}^{l_c}$ remains identical at each iteration. We verified this through an additional experiment using iterative \spread, where \opti was performed only in the first iteration. We found that keeping $\bar h_{t,k}^{l_c}$ fixed across iterations leads to overfitting and performance degradation, underscoring the necessity of running \opti at every step. This confirms that, despite similar perplexities, the hidden states evolve across iterations and must be recomputed to ensure stable and effective edits. A detailed explanation of why the hidden states evolve across iterations is provided in Appendix~\ref{sec:spreadonly}.

\begin{figure}
    \centering
    \includegraphics[width=0.95\columnwidth]{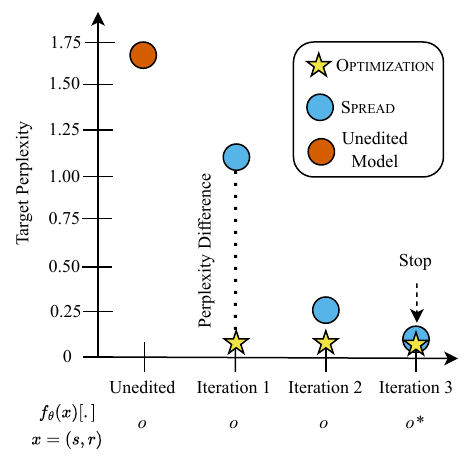}
    \caption{An editing example of using MEMIT to edit GPT-J.  Iterative model editing resolving \underedit. As the iteration proceeds the perplexity differences eventually reduces to $\leq \epsilon$, leading to the model predicting new object. The perplexity values are Box-Cox transformed to better visualize extreme high and low values.
    }
    \label{fig:perplexity-example}
\end{figure}

\subsection{Neighbor-Assisted Model Editing}
\label{sec:neighbor-edit}
Model editing must not only change the model’s output from $o$ to $o^*$ given $(s, r)$, but also preserve outputs for neighboring knowledge, i.e $(\tilde{s}, r, o)$, where $\tilde{s}$ is a new subject sharing the same relation $r$.
A preservation example is shown in Figure~\ref{fig:edit-example}a: iPhone 11 [$\tilde{s}$] is still produced by [$r$]  Apple [$o$] despite iPad is edited to produced by Honda\footnote{In the C{\smaller OUNTER}F{\smaller ACT} dataset}. 

Existing model editing algorithms struggle to preserve neighboring knowledge because \opti is designed solely to maximize the likelihood of the new knowledge, $(s, r, o^*)$. Moreover, iterative model editing can exacerbate this \overedit issue, as each iteration continues to reinforce the new knowledge without explicitly preserving neighboring knowledge. 

\citet{gangadhar-stratos-2024-model} argue that incorporating neighboring knowledge while learning new facts through fine-tuning is more effective at preserving such neighbors compared to conventional model editing. Inspired by this observation, we hypothesize that incorporating neighboring knowledge into the \opti stage can help to mitigate \overedit.

We propose neighbor-assisted model editing,  which optimizes $\bar h_{t}^{l_c}$ to maximize the likelihood of the new knowledge $(s,r,o^*)$ and the neighboring knowledge $(\tilde s, r,o)$. To accomplish this we define the loss function for $\delta$ as:
\begin{align}
\tilde g(\delta) \dot = &\ -\frac{1}{n}\sum_{i=1}^n  \ln f_\theta \left (x_i,h^{l_c}_t+\delta \right )[o^*] \\
&\ + D_\text{KL} \left (f_\theta \left ( s, h^{l_c}_t+\delta \right ) || f_\theta \left (s \right ) \right ) \\
&\ - \ln f_\theta \left (\tilde x,h^{l_c}_t+\delta \right )[o],
\end{align}
where $\tilde x= (\tilde{s}, r)$  without any prefix. In our experiments we only included a single neighboring knowledge fact $(\tilde s, r, o)$, but it should be extensible to multiple neighboring knowledge facts. We omit the iteration notation $k$ here for simplicity. The proposed loss function change could be easily applied in different iterations.

\begin{table*}[t]
\resizebox{\textwidth}{!}{%
\begin{tabular}{@{}llrrrrrrrrrrr@{}}
\toprule
\multirow{2}{*}{\textbf{Model}}                                                     & \multirow{2}{*}{\textbf{Algo}}  & \multicolumn{1}{l}{\multirow{2}{*}{\textbf{k}}} & \multicolumn{2}{l}{\textbf{Efficacy ($\uparrow$)}}                                          & \multicolumn{2}{l}{\textbf{Generalization ($\uparrow$)}}                                   & \multicolumn{2}{l}{\textbf{Specificity ($\uparrow$)}}                                       & \multicolumn{2}{l}{\textbf{Score ($\uparrow$)}}                                            & \multicolumn{1}{l}{\textbf{Perplexity ($\downarrow$)}} & \multicolumn{1}{l}{$|\Delta p_k|(\downarrow)$}      \\ \cmidrule(l){4-13} 
                                                                                    &                                 & \multicolumn{1}{l}{}                            & \multicolumn{1}{l}{\textbf{Accuracy}} & \multicolumn{1}{l}{\textbf{Success}}   & \multicolumn{1}{l}{\textbf{Accuracy}} & \multicolumn{1}{l}{\textbf{Success}}  & \multicolumn{1}{l}{\textbf{Accuracy}} & \multicolumn{1}{l}{\textbf{Success}}   & \multicolumn{1}{l}{\textbf{Accuracy}} & \multicolumn{1}{l}{\textbf{Success}}  & \multicolumn{1}{l}{\textbf{ME-PPL-50}}   & \multicolumn{1}{l}{\textbf{}} \\ \midrule
\multirow{8}{*}{\textbf{\begin{tabular}[c]{@{}l@{}}GPT-2 XL\\ (1.5B)\end{tabular}}} & \textbf{Unedited}                   & \multicolumn{1}{r|}{0}                          & 0.00\%                                & \multicolumn{1}{r|}{21.67\%}           & 0.00\%                                & \multicolumn{1}{r|}{31.33\%}          & \multicolumn{1}{l}{}                  & \multicolumn{1}{r|}{56.67\%}           & \multicolumn{1}{l}{}                  & \multicolumn{1}{r|}{31.33\%}          & 54.66                                                  & \multicolumn{1}{l}{}                           \\ \cmidrule(l){2-13} 
                                                                                    & \textbf{ROME}                       & \multicolumn{1}{r|}{1}                          & 1.00\%                                & \multicolumn{1}{r|}{56.00\%}           & 1.00\%                                & \multicolumn{1}{r|}{55.00\%}          & 0.00\%                                & \multicolumn{1}{r|}{92.00\%}           & 1.00\%                                & \multicolumn{1}{r|}{64.00\%}          & 7.07E+03                                               & 7.15E+05                                       \\ \cmidrule(l){2-13} 
                                                                                    & \multirow{2}{*}{\textbf{MEMIT}}     & \multicolumn{1}{r|}{1}                          & 79.00\%                               & \multicolumn{1}{r|}{92.67\%}           & 22.00\%                               & \multicolumn{1}{r|}{65.67\%}          & \textbf{75.00\%}                      & \multicolumn{1}{r|}{\textbf{99.00\%}}  & 42.67\%                               & \multicolumn{1}{r|}{83.00\%}          & 58.66                                                  & 11359.60                                       \\
                                                                                    &                                     & \multicolumn{1}{r|}{4}                          & \textbf{99.67\%}                      & \multicolumn{1}{r|}{\textbf{99.67\%}}  & \textbf{35.67\%}                      & \multicolumn{1}{r|}{\textbf{76.00\%}} & 68.67\%                               & \multicolumn{1}{r|}{\textbf{99.00\%}}  & \textbf{56.67\%}                      & \multicolumn{1}{r|}{\textbf{90.00\%}} & 62.44                                                  & 0.47                                           \\ \cmidrule(l){2-13} 
                                                                                    & \multirow{2}{*}{\textbf{PMET}}      & \multicolumn{1}{r|}{1}                          & 21.67\%                               & \multicolumn{1}{r|}{57.00\%}           & 3.00\%                                & \multicolumn{1}{r|}{42.00\%}          & \textbf{91.67\%}                      & \multicolumn{1}{r|}{\textbf{100.00\%}} & 8.33\%                                & \multicolumn{1}{r|}{58.33\%}          & 55.60                                                  & 103785.71                                      \\
                                                                                    &                                     & \multicolumn{1}{r|}{10}                         & \textbf{99.33\%}                      & \multicolumn{1}{r|}{\textbf{99.67\%}}  & \textbf{24.67\%}                      & \multicolumn{1}{r|}{\textbf{70.33\%}} & 75.33\%                               & \multicolumn{1}{r|}{99.00\%}           & \textbf{47.00\%}                      & \multicolumn{1}{r|}{\textbf{87.33\%}} & 63.92                                                  & 0.42                                           \\ \cmidrule(l){2-13} 
                                                                                    & \multirow{2}{*}{\textbf{AlphaEdit}} & \multicolumn{1}{r|}{1}                          & 98.00\%                               & \multicolumn{1}{r|}{99.00\%}           & 33.00\%                               & \multicolumn{1}{r|}{76.00\%}          & \textbf{63.00\%}                      & \multicolumn{1}{r|}{99.00\%}           & 53.00\%                               & \multicolumn{1}{r|}{90.00\%}          & 59.41                                                  & 5.20E+06                                       \\
                                                                                    &                                     & \multicolumn{1}{r|}{2}                          & \textbf{99.00\%}                      & \multicolumn{1}{r|}{\textbf{100.00\%}} & \textbf{36.00\%}                      & \multicolumn{1}{r|}{\textbf{76.00\%}} & 59.00\%                               & \multicolumn{1}{r|}{\textbf{99.00\%}}  & \textbf{55.00\%}                      & \multicolumn{1}{r|}{\textbf{90.00\%}} & 60.84                                                  & 0.11                                           \\ \midrule
\multirow{8}{*}{\textbf{\begin{tabular}[c]{@{}l@{}}GPT-J\\ (6B)\end{tabular}}}      & \textbf{Unedited}                   & \multicolumn{1}{r|}{0}                          & 9.33\%                                & \multicolumn{1}{r|}{38.00\%}           & 9.00\%                                & \multicolumn{1}{r|}{38.33\%}          & \multicolumn{1}{l}{}                  & \multicolumn{1}{r|}{82.00\%}           & \multicolumn{1}{l}{}                  & \multicolumn{1}{r|}{37.33\%}          & 39.80                                                  & \multicolumn{1}{l}{}                           \\ \cmidrule(l){2-13} 
                                                                                    & \textbf{ROME}                       & \multicolumn{1}{r|}{1}                          & 1.00\%                                & \multicolumn{1}{r|}{57.00\%}           & 1.00\%                                & \multicolumn{1}{r|}{55.00\%}          & 0.00\%                                & \multicolumn{1}{r|}{76.00\%}           & 1.00\%                                & \multicolumn{1}{r|}{61.00\%}          & 2.15E+05                                               & 1.21E+14                                       \\ \cmidrule(l){2-13} 
                                                                                    & \multirow{2}{*}{\textbf{MEMIT}}     & \multicolumn{1}{r|}{1}                          & 99.00\%                               & \multicolumn{1}{r|}{100.00\%}          & 75.00\%                               & \multicolumn{1}{r|}{95.67\%}          & \textbf{69.33\%}                      & \multicolumn{1}{r|}{\textbf{89.33\%}}  & 77.67\%                               & \multicolumn{1}{r|}{94.33\%}          & 42.20                                                  & 1.22                                           \\
                                                                                    &                                     & \multicolumn{1}{r|}{2}                          & \textbf{99.33\%}                      & \multicolumn{1}{r|}{\textbf{100.00\%}} & \textbf{80.67\%}                      & \multicolumn{1}{r|}{\textbf{98.00\%}} & 66.33\%                               & \multicolumn{1}{r|}{88.33\%}           & \textbf{79.00\%}                      & \multicolumn{1}{r|}{\textbf{95.00\%}} & 43.92                                                  & 0.03                                           \\ \cmidrule(l){2-13} 
                                                                                    & \multirow{2}{*}{\textbf{PMET}}      & \multicolumn{1}{r|}{1}                          & 98.00\%                               & \multicolumn{1}{r|}{\textbf{99.67\%}}  & 76.00\%                               & \multicolumn{1}{r|}{95.00\%}          & \textbf{68.33\%}                      & \multicolumn{1}{r|}{\textbf{88.67\%}}  & 77.67\%                               & \multicolumn{1}{r|}{93.67\%}          & 41.27                                                  & 1.15                                           \\
                                                                                    &                                     & \multicolumn{1}{r|}{3}                          & \textbf{99.00\%}                      & \multicolumn{1}{r|}{\textbf{99.67\%}}  & \textbf{76.67\%}                      & \multicolumn{1}{r|}{\textbf{95.67\%}} & \textbf{68.33\%}                      & \multicolumn{1}{r|}{\textbf{88.67\%}}  & \textbf{78.33\%}                      & \multicolumn{1}{r|}{\textbf{94.00\%}} & 41.15                                                  & 0.05                                           \\ \cmidrule(l){2-13} 
                                                                                    & \multirow{2}{*}{\textbf{AlphaEdit}} & \multicolumn{1}{r|}{1}                          & 99.33\%                               & \multicolumn{1}{r|}{100.00\%}          & 68.67\%                               & \multicolumn{1}{r|}{95.33\%}          & 81.00\%                               & \multicolumn{1}{r|}{97.33\%}           & 81.33\%                               & \multicolumn{1}{r|}{97.33\%}          & 42.62                                                  & 24.24                                          \\
                                                                                    &                                     & \multicolumn{1}{r|}{5}                          & 82.00\%                               & \multicolumn{1}{r|}{97.67\%}           & 47.67\%                               & \multicolumn{1}{r|}{88.00\%}          & 39.00\%                               & \multicolumn{1}{r|}{92.67\%}           & 51.00\%                               & \multicolumn{1}{r|}{92.33\%}          & 5.84E+06                                               & \multicolumn{1}{l}{}                           \\ \midrule
\multirow{8}{*}{\textbf{\begin{tabular}[c]{@{}l@{}}Llama-2\\ (7B)\end{tabular}}}    & \textbf{Unedited}                   & \multicolumn{1}{r|}{0}                          & 15.00\%                               & \multicolumn{1}{r|}{13.67\%}           & 15.00\%                               & \multicolumn{1}{r|}{15.00\%}          & \multicolumn{1}{l}{}                  & \multicolumn{1}{r|}{84.33\%}           & \multicolumn{1}{l}{}                  & \multicolumn{1}{r|}{19.67\%}          & 30.63                                                  & \multicolumn{1}{l}{}                           \\ \cmidrule(l){2-13} 
                                                                                    & \textbf{ROME}                       & \multicolumn{1}{r|}{1}                          & 0.00\%                                & \multicolumn{1}{r|}{48.00\%}           & 0.00\%                                & \multicolumn{1}{r|}{49.00\%}          & 0.00\%                                & \multicolumn{1}{r|}{76.00\%}           & 0.00\%                                & \multicolumn{1}{r|}{55.00\%}          & 1.45E+04                                               & 8.10E+05                                       \\ \cmidrule(l){2-13} 
                                                                                    & \multirow{2}{*}{\textbf{MEMIT}}     & \multicolumn{1}{r|}{1}                          & 91.67\%                               & \multicolumn{1}{r|}{98.00\%}           & 70.33\%                               & \multicolumn{1}{r|}{93.33\%}          & 29.33\%                               & \multicolumn{1}{r|}{67.33\%}           & 50.67\%                               & \multicolumn{1}{r|}{83.67\%}          & 42.10                                                  & 198.79                                         \\
                                                                                    &                                     & \multicolumn{1}{r|}{2}                          & 14.33\%                               & \multicolumn{1}{r|}{79.00\%}           & 9.67\%                                & \multicolumn{1}{r|}{73.67\%}          & 6.67\%                                & \multicolumn{1}{r|}{70.67\%}           & 9.00\%                                & \multicolumn{1}{r|}{74.67\%}          & 9.37E+03                                               & 4664.24                                        \\ \cmidrule(l){2-13} 
                                                                                    & \multirow{2}{*}{\textbf{PMET}}      & \multicolumn{1}{r|}{1}                          & 94.33\%                               & \multicolumn{1}{r|}{97.00\%}           & 68.33\%                               & \multicolumn{1}{r|}{86.67\%}          & \textbf{76.33\%}                      & \multicolumn{1}{r|}{\textbf{89.00\%}}  & 77.33\%                               & \multicolumn{1}{r|}{90.33\%}          & 30.73                                                  & 3.32                                           \\
                                                                                    &                                     & \multicolumn{1}{r|}{2}                          & \textbf{95.33\%}                      & \multicolumn{1}{r|}{\textbf{98.33\%}}  & \textbf{70.00\%}                      & \multicolumn{1}{r|}{\textbf{88.67\%}} & 75.33\%                               & \multicolumn{1}{r|}{88.67\%}           & \textbf{78.00\%}                      & \multicolumn{1}{r|}{\textbf{91.67\%}} & 30.76                                                  & 0.09                                           \\ \cmidrule(l){2-13} 
                                                                                    & \multirow{2}{*}{\textbf{AlphaEdit}} & \multicolumn{1}{r|}{1}                          & 94.33\%                               & \multicolumn{1}{r|}{97.00\%}           & 47.67\%                               & \multicolumn{1}{r|}{67.33\%}          & \textbf{59.00\%}                      & \multicolumn{1}{r|}{\textbf{80.00\%}}  & 61.67\%                               & \multicolumn{1}{r|}{79.67\%}          & 30.79                                                  & 18.84                                          \\
                                                                                    &                                     & \multicolumn{1}{r|}{2}                          & \textbf{100.00\%}                     & \multicolumn{1}{r|}{\textbf{100.00\%}} & \textbf{67.67\%}                      & \multicolumn{1}{r|}{\textbf{89.33\%}} & 51.67\%                               & \multicolumn{1}{r|}{77.33\%}           & \textbf{68.00\%}                      & \multicolumn{1}{r|}{\textbf{87.67\%}} & 31.47                                                  & 0.15                                           \\ \midrule
\multirow{8}{*}{\textbf{\begin{tabular}[c]{@{}l@{}}Llama-3.1\\ (8B)\end{tabular}}}  & \textbf{Unedited}                   & \multicolumn{1}{r|}{0}                          & 1.00\%                                & \multicolumn{1}{r|}{7.00\%}            & 1.00\%                                & \multicolumn{1}{r|}{9.33\%}           & \multicolumn{1}{l}{}                  & \multicolumn{1}{r|}{89.67\%}           & \multicolumn{1}{l}{}                  & \multicolumn{1}{r|}{11.33\%}          & 71.73                                                  & \multicolumn{1}{l}{}                           \\ \cmidrule(l){2-13} 
                                                                                    & \textbf{ROME}                       & \multicolumn{1}{r|}{1}                          & 1.00\%                                & \multicolumn{1}{r|}{78.00\%}           & 0.00\%                                & \multicolumn{1}{r|}{68.00\%}          & 0.00\%                                & \multicolumn{1}{r|}{66.00\%}           & 1.00\%                                & \multicolumn{1}{r|}{70.00\%}          & 1.03E+05                                               & 1.44E+08                                       \\ \cmidrule(l){2-13} 
                                                                                    & \multirow{2}{*}{\textbf{MEMIT}}     & \multicolumn{1}{r|}{1}                          & 96.33\%                               & \multicolumn{1}{r|}{98.00\%}           & 52.67\%                               & \multicolumn{1}{r|}{80.33\%}          & \textbf{81.00\%}                      & \multicolumn{1}{r|}{\textbf{98.00\%}}  & 72.33\%                               & \multicolumn{1}{r|}{91.33\%}          & 72.09                                                  & 4,550.10                                       \\
                                                                                    &                                     & \multicolumn{1}{r|}{3}                          & \textbf{100.00\%}                     & \multicolumn{1}{r|}{\textbf{100.00\%}} & \textbf{68.67\%}                      & \multicolumn{1}{r|}{\textbf{93.67\%}} & 74.67\%                               & \multicolumn{1}{r|}{97.00\%}           & \textbf{79.00\%}                      & \multicolumn{1}{r|}{\textbf{96.67\%}} & 72.14                                                  & 0.01                                           \\ \cmidrule(l){2-13} 
                                                                                    & \multirow{2}{*}{\textbf{PMET}}      & \multicolumn{1}{r|}{1}                          & \textbf{2.33\%}                       & \multicolumn{1}{r|}{96.00\%}           & 7.67\%                                & \multicolumn{1}{r|}{79.00\%}          & \textbf{68.67\%}                      & \multicolumn{1}{r|}{\textbf{98.00\%}}  & \textbf{5.00\%}                       & \multicolumn{1}{r|}{90.33\%}          & 76.32                                                  & 1050.33                                        \\
                                                                                    &                                     & \multicolumn{1}{r|}{3}                          & 1.00\%                                & \multicolumn{1}{r|}{\textbf{98.33\%}}  & \textbf{11.00\%}                      & \multicolumn{1}{r|}{\textbf{88.33\%}} & 64.33\%                               & \multicolumn{1}{r|}{97.00\%}           & 3.00\%                                & \multicolumn{1}{r|}{\textbf{94.33\%}} & 78.02                                                  & 0.06                                           \\ \cmidrule(l){2-13} 
                                                                                    & \multirow{2}{*}{\textbf{AlphaEdit}} & \multicolumn{1}{r|}{1}                          & 94.67\%                               & \multicolumn{1}{r|}{97.00\%}           & 50.33\%                               & \multicolumn{1}{r|}{78.33\%}          & \textbf{76.00\%}                      & \multicolumn{1}{r|}{\textbf{97.00\%}}  & 69.00\%                               & \multicolumn{1}{r|}{90.00\%}          & 71.81                                                  & 4,470.02                                       \\
                                                                                    &                                     & 3                                               & \textbf{100.00\%}                     & \textbf{100.00\%}                      & \textbf{68.00\%}                      & \textbf{93.67\%}                      & 67.33\%                               & 95.67\%                                & \textbf{76.00\%}                      & \textbf{96.67\%}                      & 72.30                                                  & 0.01                                           \\ \bottomrule
\end{tabular}
}
\caption{Iterative model editing results on C{\smaller OUNTER}F{\smaller ACT} for at most 10 iterations (denoted by k). We compare the evaluation metrics of iteration that met stopping criterion $|\Delta p_k |\leq 1$ to that of their corresponding first iteration and \textbf{bold} the higher value. PMET on GPT-2 XL require more than 5 iterations to achieve our stopping criteria. Results for all iterations are provided in Table~\ref{tab:mcf-results-all} (Appendix \ref{sec:izsre}). While ROME is known to collapse (results reported in Table~\ref{tab:mcf-results-all}), we observed a unique case of collapse with Llama-2 (7B) specifically when using MEMIT. We discuss this in Section \ref{sec:result}. Note: Unlike EasyEdit~\cite{wang-etal-2024-easyedit}, which reports a single-edit baseline, batch editing with ROME is performed to ensure consistency with the setups of other algorithms.
} 
\label{tab:mcf-results}
\end{table*}
\section{Experimental Details}
In this section, we detail the experiments to demonstrate the effectiveness of iterative and neighbor-assisted model editing with MEMIT, PMET, AlphaEdit and ROME. 
We evaluate these algorithms with our modifications across four LLMs: GPT-2 XL (1.5B), GPT-J (6B), Llama-2 (7B) and Llama-3.1 (8B). We use EasyEdit\footnote{\url{https://github.com/zjunlp/EasyEdit}}\cite{wang-etal-2024-easyedit} with default hyperparameters; implementation details are in Appendix~\ref{sec:implement}. 

\subsection{Datasets}
To evaluate model editing across different datasets, we use the C{\smaller OUNTER}F{\smaller ACT} \cite{meng2022locating} and ZsRE \cite{DBLP:journals/corr/LevySCZ17} datasets. Both datasets consist of approximately 20k factual knowledge instances. Due to hardware limitations, for each model-editing experiment, we ran it on a subset of $m=1,\!000$ edits for each dataset. We repeat the editing task three times each using a different set of $m$ edits sampled from the whole dataset. We ensured that the edits in these trials were mutually exclusive and report the averages across them. 

It is not uncommon for a model to ``collapse" (fail on a downstream task) after editing. To evaluate model collapse we use the ME-PPL-50 dataset \citep{yang-etal-2024-butterfly}. ME-PPL-50 comprises 50 utterances, each averaging 22 tokens, sampled from LLMs’ pre-training corpora. \citet{yang-etal-2024-butterfly} demonstrated that high perplexity on this dataset correlates with failures in various downstream tasks, making it an efficient proxy for evaluating model collapse. They also observed that this behavior remains consistent regardless of dataset size. Thus, we use this smaller set. We analyze the impact of our proposed methods on model collapse in Section \ref{sec:result}.

\begin{table*}[h]
\resizebox{\textwidth}{!}{%
\begin{tabular}{@{}llr|rr|rr|rr|rr|cc@{}}
\toprule
\multirow{2}{*}{\textbf{Model}}                                                             & \multirow{2}{*}{\textbf{Algo}} & \multicolumn{1}{l}{\multirow{2}{*}{\textbf{k}}} & \multicolumn{2}{l}{\textbf{Efficacy ($\uparrow$)}}                                        & \multicolumn{2}{l}{\textbf{Generalization ($\uparrow$)}}                                  & \multicolumn{2}{l}{\textbf{Specificity ($\uparrow$)}}                                     & \multicolumn{2}{l}{\textbf{Score ($\uparrow$)}}                                           & \multicolumn{1}{l}{\textbf{Perplexity ($\downarrow$)}} & \multicolumn{1}{l}{ $|\Delta p_k|(\downarrow)$}      \\ \cmidrule(l){4-13} 
                                                                                            &                                & \multicolumn{1}{l}{}                            & \multicolumn{1}{l}{\textbf{Accuracy}} & \multicolumn{1}{l}{\textbf{Success}} & \multicolumn{1}{l}{\textbf{Accuracy}} & \multicolumn{1}{l}{\textbf{Success}} & \multicolumn{1}{l}{\textbf{Accuracy}} & \multicolumn{1}{l}{\textbf{Success}} & \multicolumn{1}{l}{\textbf{Accuracy}} & \multicolumn{1}{l}{\textbf{Success}} & \multicolumn{1}{l}{\textbf{ME-PPL-50}}   & \multicolumn{1}{l}{\textbf{}} \\ \midrule
\multirow{5}{*}{\textbf{\begin{tabular}[c]{@{}l@{}}GPT-2 XL\\ (1.5B)\\ \#739\end{tabular}}} & \textbf{Unedited}              & 0                                               & 1.00\%                                & 9.00\%                               & 1.00\%                                & 22.00\%                              & \multicolumn{1}{l}{}                  & 100.00\%                             & \multicolumn{1}{l}{}                  & 18.00\%                              & 54.66                                   & \multicolumn{1}{l}{}                 \\ \cmidrule(l){2-13} 
                                                                                            & \textbf{MEMIT}                 & 4                                               & \textbf{99.00\%}                      & \textbf{99.00\%}                     & \textbf{38.00\%}                      & \textbf{74.00\%}                     & 52.00\%                               & 77.00\%                              & 54.00\%                               & 82.00\%                              & 62.27                                   & 0.05                                 \\
                                                                                            & \textbf{NA\_MEMIT}             & 4                                               & \textbf{99.00\%}                      & \textbf{99.00\%}                     & 36.00\%                               & 70.00\%                              & \textbf{86.00\%}                               & \textbf{95.00\%}                              & \textbf{60.00\%}                      & \textbf{86.00\%}                     & 64.89                                   & 0.19                                 \\ \cmidrule(l){2-13} 
                                                                                            & \textbf{PMET}                  & 8                                               & \textbf{99.00\%}& \textbf{99.00\%}& \textbf{31.00\%}& \textbf{68.00\%}& 67.00\%& 86.00\%& 52.00\%& 82.00\%& 63.47                                   & 0.31                                 \\
                                                                                            & \textbf{NA\_PMET}              & 9                                               & 98.00\%& 98.00\%& 29.00\%& 64.00\%& \textbf{85.00\%}& \textbf{97.00\%}& \textbf{53.00\%}& \textbf{83.00\%}& 66.29                                   & 0.29                                 \\ \midrule
\multirow{5}{*}{\textbf{\begin{tabular}[c]{@{}l@{}}GPT-J\\ (6B)\\ \#960\end{tabular}}}      & \textbf{Unedited}              & 0                                               & 0.00\%                                & 8.00\%                               & 1.00\%                                & 10.00\%                              & \multicolumn{1}{l}{}                  & 100.00\%                             & \multicolumn{1}{l}{}                  & 12.00\%                              & 39.80                                   & \multicolumn{1}{l}{}                 \\ \cmidrule(l){2-13} 
                                                                                            & \textbf{MEMIT}                 & 2                                               & \textbf{99.00\%}                      & \textbf{100.00\%}                    & \textbf{79.00\%}                      & \textbf{97.00\%}                     & 63.00\%                               & 83.00\%                              & 78.00\%                               & 93.00\%                              & 44.84                                   & 0.64                                 \\
                                                                                            & \textbf{NA\_MEMIT}             & 2                                               & \textbf{99.00\%}                      & 99.00\%                              & 75.00\%                               & 92.00\%                              & \textbf{81.00\%}                      & \textbf{95.00\%}                     & \textbf{84.00\%}                      & \textbf{95.00\%}                     & 45.24                                   & 0.33                                 \\ \cmidrule(l){2-13} 
                                                                                            & \textbf{PMET}                  & 1                                               & \textbf{99.00\%}                      & \textbf{100.00\%}                    & \textbf{72.00\%}                      & \textbf{93.00\%}                     & 65.00\%                               & 84.00\%                              & 76.00\%                               & 92.00\%                              & 40.79                                   & 0.25                                 \\
                                                                                            & \textbf{NA\_PMET}              & 6                                               & 98.00\%                               & 99.00\%                              & 69.00\%                               & 89.00\%                              & \textbf{80.00\%}                      & \textbf{94.00\%}                     & \textbf{81.00\%}                      & \textbf{94.00\%}                     & 43.54                                   & 0.44                                 \\ \midrule
\multirow{3}{*}{\textbf{\begin{tabular}[c]{@{}l@{}}Llama-2\\ (7B)\\ \#1340\end{tabular}}}   & \textbf{Unedited}              & 0                                               & 38.33\%                               & 57.00\%                              & 37.00\%                               & 56.00\%                              & \multicolumn{1}{l}{}                  & 59.67\%                              & \multicolumn{1}{l}{}                  & 55.67\%                              & 33.69                                   & \multicolumn{1}{l}{}                 \\ \cmidrule(l){2-13} 
                                                                                            & \textbf{PMET}                  & 3                                               & 96.00\%                               & 98.00\%                              & 72.00\%                               & 89.00\%                              & 70.00\%                               & 92.00\%                              & 78.00\%                               & 93.00\%                              & 31                                      & 0.44                                 \\
                                                                                            & \textbf{NA\_PMET}              & 10                                              & 62.00\%                               & 79.00\%                              & 47.00\%                               & 75.00\%                              & 26.00\%                               & 76.00\%                              & 40.00\%                               & 76.00\%                              & 1414.01                                 & \multicolumn{1}{l}{}                 \\ \bottomrule
\end{tabular}
}
\caption{Neighbor-Assisted model editing results on C{\smaller OUNTER}F{\smaller ACT}. We present iteration where our proposed stopping criteria is achieved for both neighbor-assisted (NA\_) and without neighbor runs of the model editing algorithms. We compare their evaluation metrics and \textbf{bold} the higher value. Results among models and from Table~\ref{tab:mcf-results} are not comparable due to difference in neighboring samples as explained in Appendix.~\ref{sec:neighbor}. Hence, we report the no. of examples (\#) used to run experiment for each model. NA\_PMET on Llama-2 (7B) stands as an exception that didn't achieved the stopping criteria within 10 iteration and showed a performance decrease. Results for all iterations are provided in Table~\ref{tab:neighbor-all}.} 
\label{tab:mcf-neighbor-results-all}
\end{table*}

\subsection{Evaluation Metrics}
We are primarily concerned with evaluating how well iterative and neighbor-assisted model editing reduce the frequency of \underedit and \overedit. We measure how successful the editing algorithms were by examining \textit{efficacy} and \textit{generalization} scores. Efficacy measures the success of introducing new knowledge edits in the dataset. Generalization tests whether the edit is generalizable by evaluating the model on paraphrases of the examples in the dataset. To understand how well the algorithms were at avoiding \overedit{s}, we measure the \textit{specificity} of the model, i.e., how much of the neighbor knowledge remained unchanged. 
To summarize the overall performance in a \textit{score}, we use a harmonic mean of efficacy, generalization, and specificity. 

For each of these metrics, we report two evaluation scores: \textit{success} and \textit{accuracy}. 
Success is the percentage of edits where $f_{\bar \theta}(x_i)[o_i^*] > f_{\bar \theta}(x_i)[o_i]$ (or $f_{\bar \theta}(\tilde x_i)[o_i] > f_{\bar \theta}(\tilde x_i)[o^*_i]$ for specificity) with $\bar \theta$ being the final weights after editing. Accuracy is the percentage of edits where $o^*$ (or $o$ in the case of specificity) is the most likely next token.

\section{Results and Discussions}
\label{sec:result}
We show the experimental results for both iterative model editing and neighbor-assisted model editing in this section. The hardware used on running these experiments are detailed in Appendix~\ref{sec:hardware}.

\subsection{Iterative Model Editing Results}
We conducted iterative model editing experiments across all datasets, LLMs, and editing algorithms, running each configuration for at most 10 iterations. The evaluation results are presented in Table \ref{tab:mcf-results} for C{\smaller OUNTER}F{\smaller ACT} and Table \ref{tab:zsre-results} for ZsRE (provided in Appendix due to space constraints). From these experiments results, we drew several conclusions. 

First, iterative model editing consistently improves performance, with the overall success scores increasing across iterations for most models and algorithms. 
The overall success improvement stems from enhanced efficacy and generalization capabilities, which means fewer cases of \underedit. Specifically, we observed an increase in success accuracy of up to 38 percentage points, with a greater improvement in efficacy accuracy of up to 77 percentage points (PMET on GPT-2 XL). We conducted more analysis in Appendix~\ref{sec:analysis_underedit} to showcase the efficacy improvement is mostly coming from \underedit examples. Secondly, as iteration goes, the perplexity difference constantly goes down in most cases.
Finally, the proposed stopping criterion ($|\Delta p_k| \le \epsilon$) consistently halts the process, validating its reliability. Using this criterion also yields better overall scores compared to executing the algorithm only once. 
Additionally, in Section~\ref{sec:analysis_ae}, we compare our stopping criterion against two alternatives and find it to be the most effective.

While efficacy and generalization improve significantly with iterative model editing, specificity decreased in some experiments, indicating an increase in \overedit. We argue that this occurs because maximizing the likelihood of new knowledge through updates to causal layer weight parameters inadvertently affects neighboring knowledge due to shared weights. However, the overall performance increase outweighs the drop in specificity. 
 
Although iterative editing is effective in most cases, we also observed model collapse as editing progresses, indicated by high model perplexity on ME-PPL-50 (6 out of 16 experiment settings). This collapse behavior aligns with the continuous editing failures observed in previous work~\cite{gupta-etal-2024-model, meng2022locating}. This suggests that when the combination of the model and editing algorithm succeeds in continuous editing, an essential experimental setting, iterative editing can further improve model performance. 
Specifically, ROME collapses under iterative \footnote{ ROME does not support batch editing, as it can only modify one fact at a time~\cite{meng2023memit}. We discuss this further in Appendix~\ref{sec:MPR}.} editing, which aligns with prior findings on its inability to support continuous editing~\cite{gupta-etal-2024-model}. We also found that Llama-2 (7B) collapses only when edited with MEMIT, a result consistent with findings in~\citet{yang-etal-2024-butterfly}.
Thus, we conclude that iterative model editing does not inherently lead to collapse. We further establish this is Appendix~\ref{sec:model_collapse}. However, unstable models that fail with sequential editing may not benefit from this approach. Although investigating collapse is not the primary focus of this work, our findings offer a useful foundation for future research in this area.

We also consistently observed a trade-off between generalization and specificity in Table~\ref{tab:mcf-results}. Specifically, we focused on AlphaEdit, which aims to address this trade-off using null space projection to constrain model updates so that they minimally interfere with existing knowledge.
Our results show that although AlphaEdit is designed to mitigate \overedit, its performance still degrades in specificity under iterative model editing—indicating that this method does not fully resolve the problem. This reflects a No-Free-Lunch effect within locate-and-edit methods—improving one objective (e.g., efficacy) inevitably harms another (e.g., specificity). This persistent trade-off affects all locate-and-edit algorithms, and the proposed iterative editing is effective and widely applicable across this family of methods.

\subsection{Neighbor-Assisted Model Editing Results}
To evaluate neighbor-assisted model editing method, we only conducted experiments on C{\smaller OUNTER}F{\smaller ACT} due to data limitations of ZsRE explained in Appendix~\ref{sec:implement}. We perform iterative model editing with the modified neighbor loss, excluding collapsed settings (evaluation results in Table~\ref{tab:mcf-neighbor-results-all}).

We observed consistently higher specificity across all iterations when using neighbor-assisted editing (denoted by \textsc{NA\_}) compared to setups without it. This increase in specificity was accompanied by an overall improvement in score. Specifically, we observed gains of up to 6 percentage points in success accuracy and up to 34 percentage points in specificity accuracy (NA\_MEMIT on GPT-2 XL). Although there was a slight decrease in generalization, the gain in specificity was more substantial, increasing the score. 
Additionally, because no prefix was added to the neighboring knowledge, we investigate its role in Appendix~\ref{sec:analysis_nap}. We found that adding prefixes led to a slightly higher overall score but reduced the specificity score.

Moreover, the proposed stopping criteria $|\Delta p_k| \le \epsilon$, originally defined for iterative model editing, remain effective for neighbor-assisted model editing. We observed one exception in the case of LLaMA-2 (7B), where neighbor-assisted editing with PMET resulted in performance degradation. We attribute this to an increased tendency toward model collapse, as indicated by elevated perplexity (ME-PPL-50). Notably, LLaMA-2 (7B) was again the only model to exhibit such collapse behavior, reinforcing our earlier hypothesis that model-specific factors contribute to instability. However, identifying the precise training-related causes of this behavior requires deeper investigation, which we leave for future work.

\begin{table}[]
\resizebox{\columnwidth}{!}{%
\begin{tabular}{@{}llllllllll@{}}
\toprule
\multicolumn{2}{l}{\textbf{Dataset}}                                                                                                          & \multicolumn{4}{l}{\textbf{C{\smaller OUNTER}F{\smaller ACT}}}                                                                                                                   & \multicolumn{4}{l}{\textbf{ZsRE}}                                                                                                              \\ \midrule
                                                                                      & \multicolumn{1}{l|}{}                                 &                              & \textbf{Score ($\uparrow$)}                           & \textbf{$|\Delta p_k|(\downarrow$)}                  & \multicolumn{1}{l|}{\textbf{$\Delta_{p2}$  ($\downarrow$)}}                  &                              & \textbf{Score ($\uparrow$)}                           & {$|\Delta p_k|(\downarrow)$}                  & {$\Delta_{p2}(\downarrow)$}                           \\ \cmidrule(lr){4-6} \cmidrule(l){8-10} 
\multirow{-2}{*}{\textbf{Model}}                                                      & \multicolumn{1}{l|}{\multirow{-2}{*}{\textbf{Algo}}}  & \multirow{-2}{*}{\textbf{k}} & \textbf{Acc}                        & \textbf{}               & \multicolumn{1}{l|}{\textbf{}}               & \multirow{-2}{*}{\textbf{k}} & \textbf{Acc}                        & \textbf{}               & \textbf{}                        \\ \midrule
                                                                                      & \multicolumn{1}{l|}{}                                 & 4                            & \cellcolor[HTML]{9AFF99}\textbf{56.67\%} & \cellcolor[HTML]{9AFF99}0.47 & \multicolumn{1}{l|}{8.65}                         & 3                            & \cellcolor[HTML]{9AFF99}\textbf{46.67\%} & \cellcolor[HTML]{9AFF99}0.03 & 39.36                                 \\
                                                                                      & \multicolumn{1}{l|}{\multirow{-2}{*}{\textbf{MEMIT}}} &                              &                                          &                              & \multicolumn{1}{l|}{}                             & 4                            & \cellcolor[HTML]{CBCEFB}45.00\%          & 0.01                         & \cellcolor[HTML]{CBCEFB}0.02          \\ \cmidrule(l){2-10} 
                                                                                      & \multicolumn{1}{l|}{}                                 & 10                           & \cellcolor[HTML]{9AFF99}\textbf{47.00\%} & \cellcolor[HTML]{9AFF99}0.42 & \multicolumn{1}{l|}{1.29}                         & 6                            & \cellcolor[HTML]{9AFF99}\textbf{54.00\%} & \cellcolor[HTML]{9AFF99}0.19 & 1.29                                  \\
\multirow{-4}{*}{\textbf{\begin{tabular}[c]{@{}l@{}}GPT-2\\ XL \\ (1.5B)\end{tabular}}} & \multicolumn{1}{l|}{\multirow{-2}{*}{\textbf{PMET}}}  &                              &                                          &                              & \multicolumn{1}{l|}{}                             & 7                            & \cellcolor[HTML]{CBCEFB}53.33\%          & 0.08                         & \cellcolor[HTML]{CBCEFB}0.11          \\ \midrule
                                                                                      & \multicolumn{1}{l|}{}                                 & 2                            & \cellcolor[HTML]{9AFF99}\textbf{79.00\%} & \cellcolor[HTML]{9AFF99}0.03 & \multicolumn{1}{l|}{1.20}                         & 2                            & \cellcolor[HTML]{9AFF99}74.67\%          & \cellcolor[HTML]{9AFF99}0.01 & 1.65                                  \\
                                                                                      & \multicolumn{1}{l|}{\multirow{-2}{*}{\textbf{MEMIT}}} &                              &                                          &                              & \multicolumn{1}{l|}{}                             & 3                            & \cellcolor[HTML]{CBCEFB}\textbf{75.00\%} & 0.00                         & \cellcolor[HTML]{CBCEFB}\textbf{0.02} \\ \cmidrule(l){2-10} 
                                                                                      & \multicolumn{1}{l|}{}                                 & 1                            & \cellcolor[HTML]{FFCE93}77.67\%          & \cellcolor[HTML]{FFCE93}1.15 & \multicolumn{1}{l|}{}                             & 2                            & \cellcolor[HTML]{9AFF99}74.00\%          & \cellcolor[HTML]{9AFF99}0.09 & 4.06                                  \\
                                                                                      & \multicolumn{1}{l|}{}                                 & 3                            & \cellcolor[HTML]{9AFF99}\textbf{78.33\%} & \cellcolor[HTML]{9AFF99}0.05 & \multicolumn{1}{l|}{4.18}                         & 3                            & \cellcolor[HTML]{CBCEFB}\textbf{74.33\%} & 0.02                         & \cellcolor[HTML]{CBCEFB}0.07          \\
\multirow{-5}{*}{\textbf{\begin{tabular}[c]{@{}l@{}}GPTJ \\ (6B)\end{tabular}}}       & \multicolumn{1}{l|}{\multirow{-3}{*}{\textbf{PMET}}}  & 4                            & \cellcolor[HTML]{CBCEFB}78.33\%          & 0.02                         & \multicolumn{1}{l|}{\cellcolor[HTML]{CBCEFB}0.03} &                              &                                          &                              &                                       \\ \midrule
                                                                                      & \multicolumn{1}{l|}{}                                 & 2                            & \cellcolor[HTML]{9AFF99}78.00\%          & \cellcolor[HTML]{9AFF99}0.09 & \multicolumn{1}{l|}{3.18}                         & 2                            & \cellcolor[HTML]{9AFF99}78.00\%          & \cellcolor[HTML]{9AFF99}0.07 & 6.28                                  \\
\multirow{-2}{*}{\textbf{\begin{tabular}[c]{@{}l@{}}Llama-2 \\ (7B)\end{tabular}}}    & \multicolumn{1}{l|}{\multirow{-2}{*}{\textbf{PMET}}}  & 3                            & \cellcolor[HTML]{CBCEFB}\textbf{78.33\%} & 0.16                         & \multicolumn{1}{l|}{\cellcolor[HTML]{CBCEFB}0.07} & 3                            & \cellcolor[HTML]{CBCEFB}\textbf{78.67\%} & 0.02                         & \cellcolor[HTML]{CBCEFB}0.05          \\ \bottomrule
\end{tabular}
}
\caption{Comparing stopping criteria. We compare our proposed stopping criteria (green) to the two alternate stopping criteria, monotonic decrease (orange), and small change  (purple). We \textbf{bold} the higher scores among them. We report results for all iterations in Table~\ref{tab:stopcompare}.}
\label{tab:compare-short}
\end{table}
\subsection{Analysis: How effective is the stopping criterion?}
\label{sec:analysis_ae}
We tested two alternate stopping criteria to the proposed stopping criteria $|\Delta p_k|\le 1$. The first is that $|\Delta p_k|$ should monotonically decrease i.e., $|\Delta p_{k+1}| < |\Delta p_{k}|$, otherwise stop and use $\theta_k$. The second is to stop when the difference in perplexity between consecutive iterations, i.e after \spread stage, is small, i.e., $\Delta_{p2} = |p(\theta_{k+1}, h^{l_c}_{t,k+1}) - p(\theta_{k}, h^{l_c}_{t,k})| \leq 1$.
We found our proposed criteria to be most the most effective in these experiments as shown in Table \ref{tab:compare-short}. 
\section{Related Work}
In this work, we extensively discussed locate-and-edit model editing algorithms. KN~\citep{dai-etal-2022-knowledge} is a related method based on gradient-based neuron selection.
In addition, there is a body of research that employs meta-learners to guide the parameter updates required for specific edits. For example, KE~\cite{decao2021editing} uses a hyper-network to update model parameters, while MEND~\cite{mitchell2022fast} trains gradient-based, lightweight model editor networks. MALMEN~\cite{tan2024massive} builds upon MEND to address scalability challenges.
Another line of research adds new knowledge without altering the model’s parameters. SERAC~\cite{mitchell2022memorybased}, GRACE~\cite{hartvigsen2023aging}, and WISE~\cite{wang2024wise} achieve this by employing additional memory to store new knowledge. A router network is then trained to decide whether to retrieve knowledge from the original model or the additional memory, ensuring the intended knowledge is accessed without modifying the model’s core parameters. We specifically focus on locate-and-edit model editing methods due to their effectiveness and efficiency in updating only the important parameters. Our proposed method introduces simple changes to existing techniques while still demonstrating effectiveness.
\section{Conclusion}
In this work, we addressed key challenges in model editing—\underedit and \overedit—by proposing iterative and neighbor-assisted model editing techniques. Our iterative approach effectively resolves \underedit by reducing the approximation error to ensure sufficient weight updates, while neighbor-assisted editing mitigates \overedit by preserving neighboring knowledge. Extensive experiments across diverse editing algorithms, LLMs, and datasets validate the efficacy of our methods. These contributions pave the way for more reliable model editing, with broad applicability to dynamic knowledge updates in LLMs.
\section{Limitations}
We acknowledge the persistent trade-off between efficacy/generalization and specificity in locate-and-edit algorithms, which our methods do not eliminate.
We believe, these trade-offs reflect a fundamental challenge of direct model editing: LLM parameters are shared across diverse knowledge types, and no existing method to the best of our knowledge can fully isolate the parameters tied to a specific fact. Our findings highlight this limitation and underscore the need for future research into more precise editing.

Our evaluation is limited to structured factual triplets, a constraint inherited from current locate-and-edit algorithms. Editing unstructured or free-form text remains challenging because locating target knowledge is non-trivial. We acknowledge this broader methodological limitation and view support for unstructured inputs as an important direction for future work.

Limited computational resources restricted us from experimenting with larger batch sizes and additional LLMs, such as GPT-NeoX (20B) and larger Llama-2 and Llama-3.1 models. We hypothesize that the experimental result trend will remain the same, and we leave the verification of this hypothesis for future work.
\section{Ethical Statement}
Our methods strengthen locate-and-edit model editing for factual updates. The same techniques could be used to insert false or misleading facts; we do not endorse such use. Similar to the general use of LLMs, models edited with our approach can still produce incorrect information or hallucinated content, so such system should be used with caution.

\section*{Acknowledgments}
This research was supported in part by the University of Pittsburgh Center for Research Computing and Data, RRID:SCR\_022735, through the resources provided. Specifically, this work used the H2P cluster, which is supported by NSF award number OAC-2117681. Bhiman was partly funded by a grant from Pitt Cyber Program\footnote{https://www.cyber.pitt.edu/about}. We are also grateful to Michael Boratko, the Pitt NLP group, and the anonymous reviewers for their constructive feedback and suggestions.

\bibliographystyle{acl_natbib}

\appendix
\section{Locate-and-Edit Algorithms}
\label{sec:MPR}
All locate-and-edit algorithms can be formulated under a unified two-stage framework consisting of an \opti stage, where the target representation is computed, and a \spread stage, where the model weights are updated accordingly. As summarized in Table~\ref{tab:algos}, this abstraction captures a wide range of existing model editing methods, and provides a foundation on which our proposed strategies—iterative and neighbor-assisted model editing—can be broadly applied. While some prior methods, such as AlphaEdit and EVOKE (LTI), specifically target the \overedit problem, they do so using fundamentally different mechanisms from ours and do not generalize across algorithms. Notably, no existing work has directly addressed the \underedit challenge. To our knowledge, we are the first to propose a unified strategy that simultaneously mitigates both \underedit and \overedit. Detailed descriptions of each algorithm are provided below.

As discussed in Section~\ref{sec:prelim}, model editing algorithms operate on the hypothesis that updating the final MLP parameters is sufficient to increase the likelihood of a new object $o^*$ over the original object $o$ when presented with a (subject, relation) pair $x = (s, r)$ as input to the LLM. Specifically, the final MLP weight matrix $W$ functions as a linear associative memory that stores a key-value mapping $[k, v]$\footnote{Distinct from key-value pairs in attention mechanisms}\cite{meng2022locating}. Here, the key encodes the last subject token, while the value represents the relation-object pair $(r, o)$ as a property of the subject $s$. Within the transformer architecture, the key corresponds to the output of the first fully connected MLP layer, whereas the value corresponds to the output of the second fully connected MLP layer $z$, as shown in Figure\ref{fig:algodiff}. This key-value mapping $[k, v]$ is derived by computing the inner product between the key  $k$  and the final MLP weight matrix $W$ as $Wk \approx v$.

Model editing involves modifying a batch of  M  desired edits, represented as  D$ = \{(s_m, r_m, o_m, o^*_m)\}_{m=1}^M$, which translates to inserting $M$ new key-value pairs $[K_M, V_M]$ by updating the final MLP weights $W$ at the causal layers $\{l_1, \ldots, l_c\}$.

\paragraph{MEMIT}\cite{meng2023memit} operates under the assumption that factual edits can be made by modifying the final MLP weight matrix \( W^l \) at each causal layer \( l \) with a small update \( \Delta^l \), yielding new weights \( W^{l^*} = W^l + \Delta^l \). The goal is to insert \( M \gg 1 \) new key-value mappings \( [K_M, V_M] \) while preserving \( E \) existing mappings \( [K_E, V_E] \), where \( K_E = [k_e]_{e=1}^E \) and \( V_E = [v_e]_{e=1}^E \) denote the pre-existing keys and values.

MEMIT formulates this as an optimization problem to find a transformation \( \hat{W} \) that minimizes the sum of squared distances between transformed keys and their target values:
\begin{align*}
\hat{W} &\triangleq \arg\min_{\hat{W}} \sum_{i=1}^{E+M} \left\| \hat{W} k_i - v_i \right\|^2.
\end{align*}

This objective consists of two parts:
\begin{itemize}
    \item \( \sum_{i=1}^{E} \left\| \hat{W} k_i - v_i \right\|^2 \): encourages the preservation of existing knowledge.
    \item \( \sum_{i=E+1}^{E+M} \left\| \hat{W} k_i - v_i \right\|^2 \): enforces the integration of new knowledge.
\end{itemize}

To compute the optimal update \( \Delta^l \), MEMIT uses a closed-form solution derived from the residual matrix \( R^l = V - W K \), where \( K \) and \( V \) stack the relevant key and value vectors across edits. Specifically,
\begin{align*}
\Delta^{l} \leftarrow R^{l}K^{l\top}(C^{l} + K^{l}K^{l\top})^{-1},
\end{align*}
where \( C^{l} \) is a regularization term proportional to the uncentered covariance of the pre-existing keys. This provides an analytical solution for \( \Delta^l \), avoiding iterative optimization such as gradient descent. The resulting update is then applied as:
\[
W^{l^*} \leftarrow W^l + \Delta^l.
\]

In practice, the residual \( R^l \) is computed using the hidden state \( \bar{h}^{l_c}_t \) obtained in the \opti stage. For full derivation and further implementation details, see \citet{meng2023memit}.

\paragraph{PMET}\cite{Li_Li_Song_Yang_Ma_Yu_2024} shares the same optimization objective as MEMIT in the \spread stage but introduces a key difference in the \opti stage. As shown in Figure~\ref{fig:algodiff}, PMET searches for an ideal self-attention output \( \bar{\textit{attn}}^{l_c}_t \) and an ideal MLP output \( \bar{z}^{l_c}_t \). The core insight behind PMET is that the self-attention module captures generalizable patterns, while the MLP is more tightly coupled to fact-specific content. Therefore, PMET assumes that the contribution of self-attention to the hidden state \( h^{l_c}_t \) is not necessary for editing factual knowledge.

Based on this assumption, PMET reconstructs a modified hidden state using only the ideal MLP output \( \bar{z}^{l_c}_t \), effectively omitting the influence of attention. This ideal hidden state is then used to compute the residual matrix \( R^l \), and the update \( \Delta^l \) is computed using the same closed-form solution as MEMIT:
\begin{align*}
\Delta^{l} \leftarrow R^{l}K^{l\top}(C^{l} + K^{l}K^{l\top})^{-1},
\end{align*}
followed by
\[
W^{l^*} \leftarrow W^l + \Delta^l.
\]

By decoupling attention from factual edits, PMET enables more precise updates that reduce unintended interference with unrelated knowledge. Empirically, this leads to improved edit success and generalization compared to MEMIT.

\paragraph{AlphaEdit}\cite{fang2025alphaedit} extends the locate-and-edit paradigm by explicitly constraining model updates to reduce interference with existing knowledge. While it follows a similar two-stage structure (\opti and \spread), AlphaEdit introduces a novel use of null space projection~\citep{Wang_2021_CVPR} to isolate updates from directions associated with pre-existing knowledge.

During the \opti stage, AlphaEdit computes the target hidden representation \( \bar{h}^{l_c}_t \) in a manner similar to MEMIT. However, before computing the update \( \Delta^l \), it projects the residual matrix \( R^l = V - W K \) into the null space of the pre-existing keys \( K_E \). This projection ensures that the update is orthogonal to directions associated with existing key-value mappings, thereby reducing the risk of \overedit.

Formally, let \( N^l \) be a projection matrix that spans the null space of \( K_E \). AlphaEdit applies this projection to both the residual and key matrices:
\begin{align*}
\Delta^{l} \leftarrow (N^l R^{l}) (N^l K^{l})^\top \left(C^{l} + (N^l K^{l})(N^l K^{l})^\top \right)^{-1}.
\end{align*}

The resulting update is applied in the usual manner:
\[
W^{l^*} \leftarrow W^l + \Delta^l.
\]

By constraining updates to directions that are orthogonal to known information, AlphaEdit aims to improve specificity and mitigate OverEdit. However, as we show in Section~\ref{sec:result}, this constraint can limit editing flexibility, and iterative model editing can further improve AlphaEdit’s performance.

\paragraph{ROME}\cite{meng2022locating} is the predecessor of MEMIT and adopts a more constrained approach to model editing. Unlike MEMIT, which apply updates across multiple causal layers, ROME assumes that new knowledge can be fully integrated into a single transformer layer \( l_* \). It follows the same two-stage structure but focuses exclusively on editing one final MLP weight matrix \( W^{l_*} \).

In the \opti stage, ROME identifies an ideal MLP output \( \bar{z}^{l_*}_t \) for a given input \( (s, r) \). This output represents the desired value vector that should be produced by the edited layer for the edited subject. The \spread stage then computes an updated MLP weight matrix \( \hat{W}^{l_*} \) that (1) preserves all existing knowledge and (2) exactly maps a new key \( k_* \) to the target value \( v_* \). Formally, the update is obtained by solving the following constrained optimization problem:
\begin{align*}
\text{minimize} \quad & \left\| \hat{W}^{l_*} K_E - V_E \right\|^2 \\
\text{subject to} \quad & \hat{W}^{l_*} k_* = v_*,
\end{align*}
where \( K_E \) and \( V_E \) are matrices containing the keys and values for existing knowledge.

The resulting solution allows for a rank-preserving edit that satisfies the new constraint while minimizing distortion to previously stored mappings. For a complete derivation, see \citet{meng2022locating}.

While ROME produces highly precise edits—especially for single-fact updates—it does not support true batch editing.\footnote{Batch edits in ROME are executed sequentially, one fact at a time.} This makes it prone to instability when applying many edits, as sequential updates can interfere destructively. Prior work~\citep{gupta-etal-2024-model,yang-etal-2024-butterfly} shows that ROME suffers from model collapse when used iteratively, due to sharp increases in perplexity arising from the accumulation of such updates.

\paragraph{R-ROME}\cite{gupta-etal-2024-rebuilding} builds on ROME with the explicit goal of resolving its tendency to collapse under sequential edits. While ROME applies a constrained rank-one update to a single MLP weight matrix \( W^{l_*} \), it suffers from instability when edits are applied repeatedly. R-ROME attributes this collapse to how ROME computes the key vector \( k \) used in the update rule.

In ROME, the key \( k \) is directly computed from the current subject \( s \), without considering broader contextual signals. R-ROME proposes a revised formulation in which the key \( k_* \) is computed as an average over multiple neighboring prompts \( x_j + s \), resulting in a more stable and robust key representation. This modification aligns the computation of the key \( k_* \) and the value \( v_* \), which are jointly used to derive the update.

The updated weight matrix is then computed via:
\[
\hat{W} = W + \Lambda_* (C^{-1} k_*)^\top,
\]
where \( \Lambda_* \) scales the residual \( v_* - Wk_* \) based on the adjusted key \( k_* \). By unifying the key used in both the value computation and the update direction, R-ROME mitigates the destructive interference observed in ROME and reduces the likelihood of collapse from so-called ``disabling edits''.

Empirically, R-ROME improves stability in sequential editing scenarios while retaining the precision benefits of ROME. However, like ROME, it remains limited to editing one fact at a time and does not natively support batch updates.

\paragraph{EMMET}\cite{gupta-etal-2024-unified} unifies the ROME and MEMIT families of model editing by showing that both optimize a common preservation-memorization objective. While ROME performs single edits using strict equality constraints and MEMIT enables batched edits via a least-squares formulation, EMMET generalizes both by introducing equality-constrained batch editing. This allows EMMET to combine the precision of ROME with the scalability of MEMIT.

Like other methods in the locate-and-edit family, EMMET follows the two-stage editing framework. In the \opti stage, it identifies the target value vectors \( V_E \) corresponding to the new facts to be inserted. In the \spread stage, it solves a constrained optimization problem to update the MLP weight matrix \( W_0 \) to a new matrix \( \hat{W} \) that satisfies two goals: (1) preserving the projections of a set of pre-existing keys \( K_0 \), and (2) exactly mapping a batch of new keys \( K_E \) to their corresponding values \( V_E \). Formally:
\begin{align*}
&\hat{W} = \arg\min_{\hat{W}} \left\| \hat{W} K_0 - W_0 K_0 \right\|^2 
\quad \text{s.t.} \\
&\quad \hat{W} k^{(e)}_i = v^{(e)}_i \quad \forall i \in [1, E],
\end{align*}
where the first term enforces knowledge preservation and the constraints enforce exact memorization of the new facts.

This preservation-memorization objective is solved using Lagrange multipliers, yielding the closed-form update:
\[
\hat{W} = W_0 + (V_E - W_0 K_E)(K_E^\top C_0^{-1} K_E)^{-1} K_E^\top C_0^{-1},
\]
where \( C_0 = K_0 K_0^\top \) is the uncentered covariance matrix of the preserved keys.

By design, EMMET enables high-precision edits with support for batch sizes up to 10,000, effectively bridging the capabilities of ROME and MEMIT within a unified formulation. Empirically, it achieves performance comparable to MEMIT while retaining the theoretical rigor of equality-constrained updates. As such, EMMET offers a principled and scalable approach to batch model editing that fits cleanly into the \opti/\spread framework.

\paragraph{ENCORE}\cite{gupta2025lifelongsequentialknowledgeediting} addresses the long-term instability of locate-and-edit methods under large-scale sequential editing. Prior work has shown that repeated edits with methods like MEMIT and ROME lead to overfitting on edited facts and model collapse due to uncontrolled growth in the norm of updated weights. ENCORE introduces two key enhancements—\textit{Most-Probable Early Stopping (MPES)} and a \textit{norm constraint}—to enable robust, long-horizon editing.

In the \opti stage, ENCORE adopts MPES, an early stopping strategy that halts optimization once the target object becomes the most probable prediction across all optimization queries. This prevents overfitting by stopping the editing process before excessive memorization occurs, similar to how early stopping in training halts based on validation loss.

In the \spread stage, ENCORE augments the standard preservation-memorization objective with a Frobenius norm penalty that discourages large deviations from the original weights:
\begin{align*}
\mathcal{L}(\hat{W}) = 
&\;\; \lambda_p \| \hat{W}K_0 - W_0K_0 \|^2 \\
&+ \| \hat{W}K_1 - V_1 \|^2 \\
&+ \lambda_n \| \hat{W} - W_0 \|_F^2,
\end{align*}
where the first two terms represent knowledge preservation and memorization (as in MEMIT), and the third term explicitly controls norm growth. This yields a closed-form solution:
\begin{align*}
\hat{W} = W_0 + (V_1 - W_0K_1)&K_1^\top 
\Big(\lambda_p K_0 K_0^\top \\
&+ K_1 K_1^\top + \lambda_n I\Big)^{-1}.
\end{align*}

Like other methods in this space, ENCORE cleanly fits into the two-stage locate-and-edit framework, using MPES for target identification in \opti and a norm-constrained update formulation in \spread.

\paragraph{EVOKE (LTI)}\cite{zhang2025uncovering} introduces a plug-and-play optimization strategy called \textit{Learn the Inference (LTI)} to mitigate \overedit in complex reasoning tasks such as multi-hop inference. While prior locate-and-edit methods like ROME and MEMIT often overfit to edit targets, assigning disproportionately high probabilities to them, EVOKE attributes this \textit{Editing Overfit} to the strong coupling between the edit prompt and the target object. To address this, LTI regularizes the optimization process by incorporating auxiliary constraints inspired by how unedited LLMs recall knowledge via in-context learning.

EVOKE operates within the standard two-stage editing framework. In the \opti stage, it modifies the optimization objective used to compute the target value vector \( v^* \), introducing three additional constraints:

\begin{itemize}
    \item \textbf{Subject Representation Constraint (SRC):} aligns the intermediate representation of the subject token between the edited and unedited models to avoid overfitting on narrow context.
    \item \textbf{Output Distribution Constraint (ODC):} matches the output distributions of the edited and unedited models to preserve global behavior.
    \item \textbf{New Knowledge Constraint (NKC):} ensures the edited model correctly predicts the new target object across randomly prefixed contexts.
\end{itemize}

These constraints are jointly optimized via a weighted objective:
\[
\mathcal{L} = \lambda \, \mathcal{L}_{\mathrm{SRC}} + \beta \, \mathcal{L}_{\mathrm{ODC}} + \alpha \, \mathcal{L}_{\mathrm{NKC}},
\]
where \( \lambda, \beta, \alpha \) are tunable weights. Once the optimal residual vector \( h \) is learned, the final update in the \spread stage follows the standard weight update (e.g., as used in ROME).

While EVOKE aims to mitigate \overedit through more informed optimization, its approach differs from our neighbor-assisted editing strategy in key ways. EVOKE constrains the model’s internal representations by comparing to unedited inference with prepended context, whereas we use neighboring samples that share the same relation to guide editing. Additionally, EVOKE perturbs the value vector \( v^* \) while holding the subject fixed and varying the relation; in contrast, we vary the subject and fix the relation to ensure the original object remains unchanged. EVOKE also uses significantly more prompts per edit sample, which may affect scalability.

Despite these differences, EVOKE still adheres to the two-stage locate-and-edit framework. As such, our proposed iterative and neighbor-assisted methods are compatible with it and could further enhance its performance by addressing \underedit and reinforcing relation-consistent specificity.

\section{Implementation details}
\label{sec:implement}
\subsection{Iterative model editing}
\label{sec:iteration}

Currently, our implementation requires running an additional iteration to compute $p(\theta_{k+1}, \hat h^{l_c}_{t,k})$ for iteration $k$. As a result, $p(\theta_{k+1}, \hat h^{l_c}_{t,k})$ for the 5th iteration is not reported in Tables~\ref{tab:mcf-results-all}, \ref{tab:zsre-results}, \ref{tab:neighbor-all}, and \ref{tab:nap-neighbor-results-all}. We are updating our code to compute $p(\theta_{k+1}, \hat h^{l_c}_{t,k})$ at the end of the $k$th iteration without requiring the next iteration. This update involves running only the \opti stage with a single gradient step, using the initialization vector, before proceeding to the next iteration.
\begin{figure*}[t]
\centering
\begin{minipage}{0.95\textwidth}
\begin{lstlisting}
{
  "prompt": "The original language of Face Dances was",
  "subject": "Face Dances",
  "ground_truth": "English",
  "target_new": "Italian",
  "locality": {
    "neighborhood": {
      "prompt": [
        "The original language of Ghost Rider is",
        "The original language of 42nd Street was",
        "The original language of The Fox and the Hound was",
        "The language of Titanic is"
      ],
      "ground_truth": [
        "English",
        "English",
        "English",
        "English"
      ]
    }
  },
  "neighborhood_samples": [
    {
      "subject": "Star Wars: Episode III 2013 Revenge of the Sith",
      "prompt": "The original language of {} was"
    }
  ]
}
\end{lstlisting}
\end{minipage}
\caption{C{\smaller OUNTER}F{\smaller ACT} data sample for neighbor-assited model editing.}
\label{fig:json-example}
\end{figure*}
\begin{table*}
\resizebox{\textwidth}{!}{%
    \centering
    \begin{tabular}{cllc}\toprule
         \textbf{Model} &\textbf{\# Edited Examples} &\textbf{Neighborhood Examples / Edit}& A\textbf{verage evaluation neighboring samples (ceil) / Edit}\\\midrule
         \textbf{GPT-2 XL} &739 &1& 2\\
         \textbf{GPT-J} &690 &1& 4\\
         \textbf{Llama-2-7B} &1340 &1& 4\\ \bottomrule
    \end{tabular}
    }
    \caption{Average neighboring sample used to evaluate each edit in neighbor-assisted editing}
    \label{tab:avg_neighbor}
\end{table*}

\subsection{Neighbor-assisted model editing}
\label{sec:neighbor}
We observed that different models often produce varying objects for the same neighboring knowledge. To calculate specificity, we used the model’s actual output as the object ($o$), which should remain unchanged during editing. However, this introduced a challenge when employing neighbor-assisted model editing to guide the finding of the ideal hidden state $\bar h^{l_c}_t$ during \opti, as models produced inconsistent outputs for neighboring knowledge used for evaluation. This inconsistency caused conflicts among neighboring knowledge when selecting a single instance for neighbor-assisted editing.

To address this, we filtered out neighboring knowledge samples that did not yield the same model output as the original ground truth reported in the dataset. This strategy ensured that any remaining neighboring knowledge sample could be randomly selected for neighbor-assisted editing, resolving the conflict.

This approach also introduced an additional constraint on the edit data points: each data point must have at least two neighboring knowledge samples—one for editing and the others for evaluation. Unfortunately, the ZsRE dataset contains only one neighboring knowledge sample per data point, restricting us to the C{\smaller OUNTER}F{\smaller ACT} dataset. Even within C{\smaller OUNTER}F{\smaller ACT}, only a limited number of samples met the required constraints. The number of qualifying samples varied depending on the model, as shown in Table~\ref{tab:avg_neighbor}. For clarity, we have added an illustrative example drawn from the C{\smaller OUNTER}F{\smaller ACT} dataset (Figure~\ref{fig:json-example}). In our neighbor-assisted model-editing setup, each edit consists of (i) a prompt—formed by the subject + relation pair—and (ii) a target\_new (the new object). In addition, a single neighborhood\_sample is always provided to guide the edit. Edit specificity is then assessed with the metric locality[neighborhood][prompt], whose neighborhood size varies as reported in Table~\ref{tab:avg_neighbor}. Therefore, in our neighbor-assisted model editing, exactly one neighboring example was used to assist the edit. Additionally, an average of 2, 4, and 4 neighboring examples (as reported in the Table~\ref{tab:avg_neighbor}) were used for evaluating each edit. Generating multiple neighbor knowledge using ChatGPT~\citep{meng2022locating} is straightforward and convenient. However, scaling this to all possible neighbors for a single piece of knowledge is a highly challenging research problem that falls beyond the scope of our paper.

\begin{table}[h]
\resizebox{\columnwidth}{!}{%
\begin{tabular}{@{}llll@{}}
\toprule
\textbf{Model}                                                                      & \textbf{Algo}                   & \textbf{Dataset} & \textbf{GPU} \\ \midrule
\multirow{6}{*}{\textbf{\begin{tabular}[c]{@{}l@{}}GPT-2 XL\\ (1.5B)\end{tabular}}} & \multirow{2}{*}{\textbf{ROME}}  & C{\smaller OUNTER}F{\smaller ACT}      & L40S (48GB)  \\
                                                                                    &                                 & ZsRE             & L40S (48GB)  \\ \cmidrule(l){2-4} 
                                                                                    & \multirow{2}{*}{\textbf{MEMIT}} & C{\smaller OUNTER}F{\smaller ACT}      & L40S (48GB)  \\
                                                                                    &                                 & ZsRE             & L40S (48GB)  \\ \cmidrule(l){2-4} 
                                                                                    & \multirow{2}{*}{\textbf{PMET}}  & C{\smaller OUNTER}F{\smaller ACT}      & L40S (48GB)  \\
                                                                                    &                                 & ZsRE      & L40S (48GB)  \\ \midrule
\multirow{6}{*}{\textbf{\begin{tabular}[c]{@{}l@{}}GPT-J\\ (6B)\end{tabular}}}      & \multirow{2}{*}{\textbf{ROME}}  & C{\smaller OUNTER}F{\smaller ACT}      & A100 (80GB)  \\
                                                                                    &                                 & ZsRE             & A100 (80GB)  \\ \cmidrule(l){2-4} 
                                                                                    & \multirow{2}{*}{\textbf{MEMIT}} & C{\smaller OUNTER}F{\smaller ACT}      & L40S (48GB)  \\
                                                                                    &                                 & ZsRE             & A100 (80GB)  \\ \cmidrule(l){2-4} 
                                                                                    & \multirow{2}{*}{\textbf{PMET}}  & C{\smaller OUNTER}F{\smaller ACT}      & A100 (80GB)  \\
                                                                                    &                                 & ZsRE             & A100 (80GB)  \\ \midrule
\multirow{6}{*}{\textbf{\begin{tabular}[c]{@{}l@{}}Llama 2\\ (7B)\end{tabular}}}    & \multirow{2}{*}{\textbf{ROME}}  & C{\smaller OUNTER}F{\smaller ACT}      & A100 (80GB)  \\
                                                                                    &                                 & ZsRE             & A100 (80GB)  \\ \cmidrule(l){2-4} 
                                                                                    & \multirow{2}{*}{\textbf{MEMIT}} & C{\smaller OUNTER}F{\smaller ACT}      & L40S (48GB)  \\
                                                                                    &                                 & ZsRE             & A100 (80GB)  \\ \cmidrule(l){2-4} 
                                                                                    & \multirow{2}{*}{\textbf{PMET}}  & C{\smaller OUNTER}F{\smaller ACT}      & L40S (48GB)  \\
                                                                                    &                                 & ZsRE             & A100 (80GB)  \\ \bottomrule
\end{tabular}
}
\caption{GPU requirements to conduct 1000 edits}
\label{tab:gpu}
\end{table}
\section{Hardware Details}
\label{sec:hardware}
Table~\ref{tab:gpu} outlines the GPU resources utilized to conduct edits of batch size 1000 across various models, algorithms, and datasets. It highlights the specific hardware configurations, such as GPU type (e.g., NVIDIA L40S with 48GB memory or NVIDIA A100 with 80GB memory), used for each experiment.

\begin{table*}[]
\resizebox{\textwidth}{!}{%
\begin{tabular}{@{}rrrrrrrrrlll@{}}
\toprule
\multicolumn{1}{l}{\multirow{2}{*}{\textbf{k}}} & \multicolumn{2}{l}{\textbf{Efficacy ($\uparrow$)}} & \multicolumn{2}{l}{\textbf{Generalization ($\uparrow$)}} & \multicolumn{2}{l}{\textbf{Specificity ($\uparrow$)}} & \multicolumn{2}{l}{\textbf{Score ($\uparrow$)}}   & \multicolumn{3}{l}{\textbf{Perplexity ($\downarrow$)}}                                   \\ \cmidrule(l){2-12}
\multicolumn{1}{l}{}                            & \textbf{Accuracy}  & \textbf{Success} & \textbf{Accuracy}     & \textbf{Success}    & \textbf{Accuracy}   & \textbf{Success}   & \textbf{Accuracy} & \textbf{Success} & \textbf{ME-PPL-50} & \textbf{$p(\theta_k, \bar h_{t,k}^{l_c})$} & \textbf{$p(\theta_{k+1}, \hat h^{l_c}_{t,k})$} \\ \midrule
\multicolumn{1}{r|}{1}                          & 79.00\%                               & \multicolumn{1}{r|}{93.00\%}         & 22.00\%                               & \multicolumn{1}{r|}{65.00\%}         & 75.00\%                               & \multicolumn{1}{r|}{99.00\%}         & 42.00\%                               & \multicolumn{1}{r|}{83.00\%}         & 57.98              & 1.02                     & 10816.41                  \\
\multicolumn{1}{r|}{2}                          & 89.00\%                               & \multicolumn{1}{r|}{98.00\%}         & 35.00\%                               & \multicolumn{1}{r|}{76.00\%}         & 61.00\%                               & \multicolumn{1}{r|}{98.00\%}         & 53.00\%                               & \multicolumn{1}{r|}{89.00\%}         & 84.28              & 1.02                     & 3700.56                   \\
\multicolumn{1}{r|}{3}                          & 68.00\%                               & \multicolumn{1}{r|}{95.00\%}         & 28.00\%                               & \multicolumn{1}{r|}{72.00\%}         & 54.00\%                               & \multicolumn{1}{r|}{97.00\%}         & 44.00\%                               & \multicolumn{1}{r|}{86.00\%}         & 174.62             & 1.02                     & 64858.83                  \\
\multicolumn{1}{r|}{4}                          & 41.00\%                               & \multicolumn{1}{r|}{85.00\%}         & 19.00\%                               & \multicolumn{1}{r|}{68.00\%}         & 47.00\%                               & \multicolumn{1}{r|}{97.00\%}         & 31.00\%                               & \multicolumn{1}{r|}{81.00\%}         & 348.5              & 1.02                     & 2254146.29                \\
\multicolumn{1}{r|}{5}                          & 20.00\%                               & \multicolumn{1}{r|}{76.00\%}         & 11.00\%                               & \multicolumn{1}{r|}{64.00\%}         & 37.00\%                               & \multicolumn{1}{r|}{97.00\%}         & 17.00\%                               & \multicolumn{1}{r|}{76.00\%}         & 602.14             & 1.02                     & 5827805.82   \\ \bottomrule                      
\end{tabular}
}
\caption{Iterative \spread model editing results on C{\smaller OUNTER}F{\smaller ACT} with MEMIT on GPT-2 XL for 5 iterations. Unlike \spread, \opti was run only in first iteration.
} 
\label{tab:spreadonly}
\end{table*}
\section{Iterative \spread}
\label{sec:spreadonly}
In Figure~\ref{fig:perplexity-example}, we observe little to no change in the perplexity of $\bar{h}^{l_c}_{t,k}$ across iterations. However, we argue that despite similar perplexity values, the actual representations of $\bar{h}^{l_c}_{t,k}$ differ at each step. This is because $\theta_{k+1}$ is computed based on $\bar{h}^{l_c}_{t,k}$, which itself is derived from $\theta_k$, as described in Section~\ref{sec:iterative-edit}. Due to this recursive dependency, every update to $\theta_k$ induces a corresponding change in $\bar{h}^{l_c}_{t,k}$. Since the overall editing goal remains consistent across iterations, the magnitude of change required in $\theta_k$ naturally diminishes over time, resulting in a sequence of distinct yet converging hidden states.

To further explore this, consider a scenario where $\bar{h}^{l_c}_{t,k}$ remains constant across iterations. In such a case, one might skip repeated calls to \opti and only perform \spread, potentially saving computation without sacrificing performance. However, we hypothesize that this would lead to overfitting due to repeated use of the same update. To test this, we conducted an additional experiment using iterative \spread with MEMIT on GPT-2 XL (C{\smaller OUNTER}F{\smaller ACT} dataset), where \opti was run only in the first iteration. The results, shown in Table~\ref{tab:spreadonly}, reveal performance degradation and an increase in ME-PPL-50. Although we observe a temporary performance improvement in iteration 2, likely due to overfitting, it is followed by continued degradation, and the editing process fails to meet its stopping criterion. This underscores the importance of running \opti in every iteration and indicates that it generates distinct target hidden states at each step.

\section{Factors Contributing Model Collapse}
\label{sec:model_collapse}
Existing literature attributes model collapse during editing to two main factors: (1) the sequential nature of edits~\citep{gupta-etal-2024-model, yang-etal-2024-butterfly}, and (2) characteristics of the specific examples being edited~\citep{yang-etal-2024-butterfly}. Since iterative model editing falls under the first category, one might expect it to be susceptible. However, prior work shows that collapse typically occurs after thousands of edits, e.g., up to 3k in AlphaEdit~\cite{fang2025alphaedit} and 10k in ENCORE~\cite{gupta2025lifelongsequentialknowledgeediting}, whereas our iterative approach involves only a few single-digit iterations. This makes collapse due to iteration count alone unlikely.

Given that all models in our study were evaluated on the same dataset, we can reasonably rule out the second factor. Furthermore, since no GPT models, and not even LLaMA-3.1 (8B), exhibited similar collapse behavior, we hypothesize a third contributing factor: model-specific characteristics that may stem from differences in training strategies. Finally, the editing algorithm itself may serve as a fourth contributing factor. For example, MEMIT’s less targeted weight updates (compared to PMET) may introduce excessive parameter shifts, as discussed in Appendix~\ref{sec:MPR}.

In summary, while iterative model editing is unlikely to cause collapse on its own, interactions with model-specific or algorithm-specific factors may trigger instability. Although investigating collapse is not the primary focus of this work, our findings offer a useful foundation for future research in this area.

\begin{figure}[h]
    \centering
    \includegraphics[width=\columnwidth]{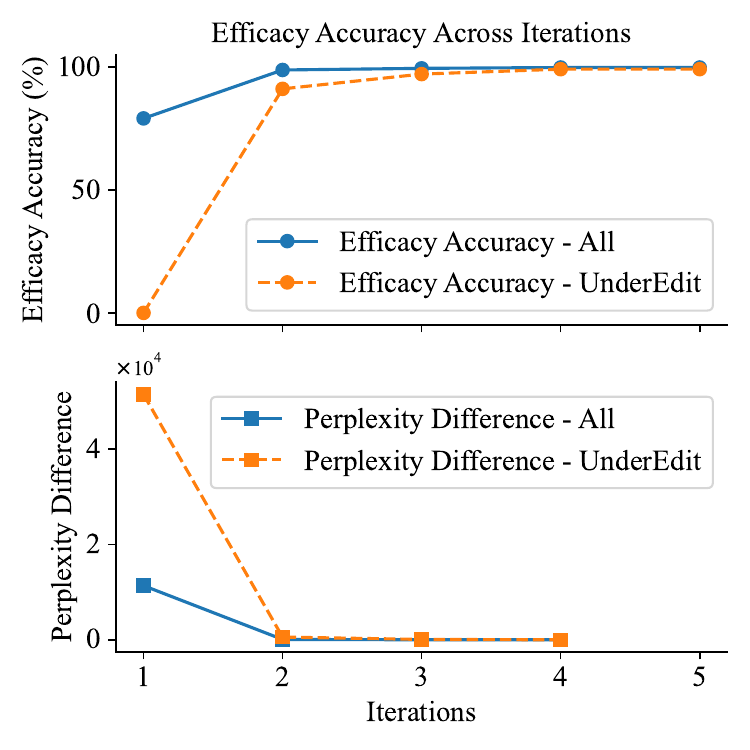}
    \caption{Improvement in efficacy accuracy and reduction in $|\Delta p_k|$ for \underedit examples over iterative model editing. The results show that iterative editing mitigates \underedit cases in GPT-2 XL edited with MEMIT on C{\smaller OUNTER}F{\smaller ACT}, contributing to overall performance gains.}
    \label{fig1:hard}
\end{figure}
\section{Additional Analysis}
\label{sec:AA}
This section dives deeper into the proposed methods. Specifically, we aim to understand how iterative model editing addresses \underedit, the impact of prefixing in neighbor-assisted model editing for resolving \overedit, and the effectiveness of the stopping criteria.
\subsection{How does iterative model editing address \texorpdfstring{\underedit}{UnderEdit}?}
\label{sec:analysis_underedit}
We hypothesize that the iterative model editing approach can reduce the number of \underedit cases. To test this hypothesis, we identified \underedit examples in GPT-2 XL edited with MEMIT on C{\smaller OUNTER}F{\smaller ACT} after the first iteration, i.e, edits $(s_i,r_i,o_i,o^*_i)$ where $f_{\theta_2}(x_i)[o_i^*] < f_{\theta_2}(x_i)[o_i]$. We then tracked $|\Delta p_k|$ and efficacy accuracy across subsequent iterations. Figure  \ref{fig1:hard} illustrates the results of the analysis. 
We observe that the rate of $|\Delta p_k|$ decrease and accuracy improving over iterations is much more pronounced for the \underedit examples. This observation tells us that multiple iterations of $\spread$ is the larger contributors to getting higher performance. 

\begin{table}[]
\resizebox{\columnwidth}{!}{%
\begin{tabular}{@{}lrrrrr@{}}
\toprule
\multirow{2}{*}{\textbf{Algo}} & \multicolumn{1}{l}{\multirow{2}{*}{\textbf{k}}} & \multicolumn{1}{l}{\textbf{Efficacy ($\uparrow$)}} & \multicolumn{1}{l}{\textbf{Generalization ($\uparrow$)}} & \multicolumn{1}{l}{\textbf{Specificity ($\uparrow$)}} & \multicolumn{1}{l}{\textbf{Score ($\uparrow$)}}    \\ \cmidrule(l){3-6} 
                               & \multicolumn{1}{l}{}                            & \multicolumn{1}{l}{\textbf{Accuracy}} & \multicolumn{1}{l}{\textbf{Accuracy}}       & \multicolumn{1}{l}{\textbf{Accuracy}}    & \multicolumn{1}{l}{\textbf{Accuracy}} \\ \midrule
\textbf{Unedited}              & 0                                               & 1.00\%                                & 1.00\%                                      & \multicolumn{1}{l}{}                     & \multicolumn{1}{l}{}                  \\ \midrule
\textbf{MEMIT}                 & 4                                               & \textbf{99.00\%}                      & \textbf{38.00\%}                            & 52.00\%                                  & 54.00\%                               \\
\textbf{NA\_MEMIT}             & 4                                               & \textbf{99.00\%}                      & 36.00\%                                     & \textbf{86.00\%}                         & 60.00\%                               \\
\textbf{NAP\_MEMIT}            & 4                                               & \textbf{99.00\%}                      & 37.00\%                                     & 84.00\%                                  & \textbf{61.00\%}                      \\ \midrule
\textbf{PMET}                  & 8                                               & \textbf{99.00\%}                      & \textbf{31.00\%}                            & 67.00\%                                  & 52.00\%                               \\
\textbf{NA\_PMET}              & 9                                               & 98.00\%                               & 29.00\%                                     & \textbf{85.00\%}                         & 53.00\%                               \\
\textbf{NAP\_PMET}             & 8                                               & 98.00\%                               & 30.00\%                                     & 84.00\%                                  & \textbf{54.00\%}                      \\ \bottomrule
\end{tabular}
}
\caption{Results of prefix-free (NA\_) and with prefix (NAP\_) neighbor-assisted model editing on GPT-2 XL on 739 samples of C{\smaller OUNTER}F{\smaller ACT}. We compare their evaluation metrics and \textbf{bold} the higher value. Full results with success and perplexity performance for all iterations are reported in Table~\ref{tab:nap-neighbor-results-all}.}
\label{tab:nap-short}
\end{table}
\subsection{How prefixes influence neighbor-assisted model editing behavior?}
\label{sec:analysis_nap}

Using random prefixes aid generalization across contexts in the memory editing process when only the target knowledge edit is known. So we pose the question, does adding random prefixes to the neighbor knowledge prompts help prevent \overedit? To answer this question we run an experiment by adding random prefixes to the neighboring knowledge used during edit. Table \ref{tab:nap-short} shows an increase in specificity accuracy\footnote{The full results with success on each measurement and perplexity performance are in Table \ref{tab:nap-neighbor-results-all}}. However, this increase is less compared to the improvement of using neighbor-assisted editing versus no neighbor-assist.
Regardless, the prefixed neighbor-assisted edits (NAP\_) achieved better overall performance, denoted by Accuracy in Score, due to a boost in generalization.

\section{Iterative model editing results}
\label{sec:izsre}
We present the complete results of all iterations of iterative model editing in Table~\ref{tab:mcf-results-all} for the C{\smaller OUNTER}F{\smaller ACT} dataset and Table~\ref{tab:zsre-results} for the ZsRE dataset. For experiments that did not meet our proposed stopping criteria within 5 iterations, we extended the runs by an additional 5 iterations and included those results as well.
\begin{table*}[h]
\resizebox{\textwidth}{!}{%
\begin{tabular}{@{}llrrrrrrrrrrrrr@{}}
\toprule
                                                                                    &                                                         & \multicolumn{1}{l}{}                             & \multicolumn{2}{l}{\textbf{Efficacy ($\uparrow$)}}                                                & \multicolumn{2}{l}{\textbf{Generalization ($\uparrow$)}}                                         & \multicolumn{2}{l}{\textbf{Specificity ($\uparrow$)}}                                            & \multicolumn{2}{l}{\textbf{Score ($\uparrow$)}}                                                  & \multicolumn{3}{l}{\textbf{Perplexity ($\downarrow$)}}                                                                                 & \multicolumn{1}{l}{\textbf{$|\Delta p_k|$ ($\downarrow$)}}              \\ \cmidrule(l){4-15} 
\multirow{-2}{*}{\textbf{Model}}                                                    & \multirow{-2}{*}{\textbf{Algo}}                         & \multicolumn{1}{l}{\multirow{-2}{*}{\textbf{k}}} & \multicolumn{1}{l}{\textbf{Accuracy}}    & \multicolumn{1}{l}{\textbf{Success}}       & \multicolumn{1}{l}{\textbf{Accuracy}}    & \multicolumn{1}{l}{\textbf{Success}}      & \multicolumn{1}{l}{\textbf{Accuracy}}    & \multicolumn{1}{l}{\textbf{Success}}      & \multicolumn{1}{l}{\textbf{Accuracy}}    & \multicolumn{1}{l}{\textbf{Success}}      & \multicolumn{1}{l}{\textbf{ME-PPL-50}} & \multicolumn{1}{l}{\textbf{$p(\theta_k, \bar h_{t,k}^{l_c})$}}  & \multicolumn{1}{l}{\textbf{$p(\theta_{k+1}, \hat h^{l_c}_{t,k})$}}              & \multicolumn{1}{l}{\textbf{}}         \\ \midrule
                                                                                    & \textbf{Unedited}                                       & 0                                                & 0.00\%                                   & 21.67\%                                   & 0.00\%                                   & 31.33\%                                  & \multicolumn{1}{l}{}                     & 56.67\%                                  & \multicolumn{1}{l}{}                     & 31.33\%                                  & 54.66                                 & \multicolumn{1}{l}{}             & 3.26E+05                                     & \multicolumn{1}{l}{}                         \\ \cmidrule(l){2-15} 
                                                                                    & \cellcolor[HTML]{FFCCC9}                                & \cellcolor[HTML]{FFCCC9}1                        & \cellcolor[HTML]{FFCCC9}1.00\%           & \cellcolor[HTML]{FFCCC9}56.00\%           & \cellcolor[HTML]{FFCCC9}1.00\%           & \cellcolor[HTML]{FFCCC9}55.00\%          & \cellcolor[HTML]{FFCCC9}0.00\%           & \cellcolor[HTML]{FFCCC9}92.00\%          & \cellcolor[HTML]{FFCCC9}1.00\%           & \cellcolor[HTML]{FFCCC9}64.00\%          & \cellcolor[HTML]{FFCCC9}7.07E+03      & \cellcolor[HTML]{FFCCC9}5.29E+03 & \cellcolor[HTML]{FFCCC9}7.20E+05             & \cellcolor[HTML]{FFCCC9}7.15E+05             \\
                                                                                    & \cellcolor[HTML]{FFCCC9}                                & \cellcolor[HTML]{FFCCC9}2                        & \cellcolor[HTML]{FFCCC9}0.00\%           & \cellcolor[HTML]{FFCCC9}55.00\%           & \cellcolor[HTML]{FFCCC9}0.00\%           & \cellcolor[HTML]{FFCCC9}51.00\%          & \cellcolor[HTML]{FFCCC9}0.00\%           & \cellcolor[HTML]{FFCCC9}93.00\%          & \cellcolor[HTML]{FFCCC9}0.00\%           & \cellcolor[HTML]{FFCCC9}62.00\%          & \cellcolor[HTML]{FFCCC9}1.36E+04      & \cellcolor[HTML]{FFCCC9}9.10E+04 & \cellcolor[HTML]{FFCCC9}1.78E+06             & \cellcolor[HTML]{FFCCC9}1.69E+06             \\
                                                                                    & \cellcolor[HTML]{FFCCC9}                                & \cellcolor[HTML]{FFCCC9}3                        & \cellcolor[HTML]{FFCCC9}0.00\%           & \cellcolor[HTML]{FFCCC9}55.00\%           & \cellcolor[HTML]{FFCCC9}0.00\%           & \cellcolor[HTML]{FFCCC9}52.00\%          & \cellcolor[HTML]{FFCCC9}0.00\%           & \cellcolor[HTML]{FFCCC9}95.00\%          & \cellcolor[HTML]{FFCCC9}0.00\%           & \cellcolor[HTML]{FFCCC9}63.00\%          & \cellcolor[HTML]{FFCCC9}1.26E+04      & \cellcolor[HTML]{FFCCC9}7.91E+05 & \cellcolor[HTML]{FFCCC9}3.68E+06             & \cellcolor[HTML]{FFCCC9}2.89E+06             \\
                                                                                    & \cellcolor[HTML]{FFCCC9}                                & \cellcolor[HTML]{FFCCC9}4                        & \cellcolor[HTML]{FFCCC9}0.00\%           & \cellcolor[HTML]{FFCCC9}55.00\%           & \cellcolor[HTML]{FFCCC9}0.00\%           & \cellcolor[HTML]{FFCCC9}50.00\%          & \cellcolor[HTML]{FFCCC9}0.00\%           & \cellcolor[HTML]{FFCCC9}88.00\%          & \cellcolor[HTML]{FFCCC9}0.00\%           & \cellcolor[HTML]{FFCCC9}60.00\%          & \cellcolor[HTML]{FFCCC9}3.13E+04      & \cellcolor[HTML]{FFCCC9}2.10E+06 & \cellcolor[HTML]{FFCCC9}1.93E+05             & \cellcolor[HTML]{FFCCC9}-1.91E+06            \\
                                                                                    & \multirow{-5}{*}{\cellcolor[HTML]{FFCCC9}\textbf{ROME}} & \cellcolor[HTML]{FFCCC9}5                        & \cellcolor[HTML]{FFCCC9}3.00\%           & \cellcolor[HTML]{FFCCC9}64.00\%           & \cellcolor[HTML]{FFCCC9}1.00\%           & \cellcolor[HTML]{FFCCC9}58.00\%          & \cellcolor[HTML]{FFCCC9}0.00\%           & \cellcolor[HTML]{FFCCC9}91.00\%          & \cellcolor[HTML]{FFCCC9}1.00\%           & \cellcolor[HTML]{FFCCC9}68.00\%          & \cellcolor[HTML]{FFCCC9}1.14E+04      & \cellcolor[HTML]{FFCCC9}7.86E+04 & \multicolumn{1}{l}{\cellcolor[HTML]{FFCCC9}} & \multicolumn{1}{l}{\cellcolor[HTML]{FFCCC9}} \\ \cmidrule(l){2-15} 
                                                                                    &                                                         & 1                                                & 79.00\%                                  & 92.67\%                                   & 22.00\%                                  & 65.67\%                                  & \textbf{75.00\%}                         & \textbf{99.00\%}                         & 42.67\%                                  & 83.00\%                                  & 58.66                                 & 1.04                             & 11360.64                                     & 11359.60                                     \\
                                                                                    &                                                         & 2                                                & 98.67\%                                  & 99.67\%                                   & 33.67\%                                  & 75.67\%                                  & 70.33\%                                  & 99.00\%                                  & 55.67\%                                  & 89.67\%                                  & 60.47                                 & 1.03                             & 78.15                                        & 77.12                                        \\
                                                                                    &                                                         & 3                                                & 99.33\%                                  & 99.67\%                                   & 35.33\%                                  & 76.00\%                                  & 69.33\%                                  & 99.00\%                                  & 56.67\%                                  & 90.00\%                                  & 61.58                                 & 1.02                             & 10.14                                        & 9.11                                         \\
                                                                                    &                                                         & \cellcolor[HTML]{9AFF99}4                        & \cellcolor[HTML]{9AFF99}\textbf{99.67\%} & \cellcolor[HTML]{9AFF99}\textbf{99.67\%}  & \cellcolor[HTML]{9AFF99}\textbf{35.67\%} & \cellcolor[HTML]{9AFF99}\textbf{76.00\%} & \cellcolor[HTML]{9AFF99}68.67\%          & \cellcolor[HTML]{9AFF99}\textbf{99.00\%} & \cellcolor[HTML]{9AFF99}\textbf{56.67\%} & \cellcolor[HTML]{9AFF99}\textbf{90.00\%} & \cellcolor[HTML]{9AFF99}62.44         & \cellcolor[HTML]{9AFF99}1.02     & \cellcolor[HTML]{9AFF99}1.49                 & \cellcolor[HTML]{9AFF99}0.47                 \\
                                                                                    & \multirow{-5}{*}{\textbf{MEMIT}}                        & 5                                                & 99.67\%                                  & 99.67\%                                   & 35.67\%                                  & 76.00\%                                  & 68.67\%                                  & 99.00\%                                  & 56.67\%                                  & 90.00\%                                  & 63.28                                 & 1.02                             & \multicolumn{1}{l}{}                         & \multicolumn{1}{l}{}                         \\ \cmidrule(l){2-15} 
                                                                                    &                                                         & 1                                                & 21.67\%                                  & 57.00\%                                   & 3.00\%                                   & 42.00\%                                  & \textbf{91.67\%}                                  & \textbf{100.00\%}                                 & 8.33\%                                   & 58.33\%                                  & 55.60                                 & 12598.80                         & 116384.51                                    & 103785.71                                    \\
                                                                                    &                                                         & 2                                                & 65.67\%                                  & 85.33\%                                   & 14.67\%                                  & 58.67\%                                  & 84.00\%                                  & 99.33\%                                  & 31.67\%                                  & 77.33\%                                  & 57.01                                 & 1333.32                          & 26852.75                                     & 25519.43                                     \\
                                                                                    &                                                         & 3                                                & 86.00\%                                  & 93.67\%                                   & 20.00\%                                  & 65.33\%                                  & 80.67\%                                  & 99.00\%                                  & 40.67\%                                  & 83.33\%                                  & 58.00                                 & 97.51                            & 5123.07                                      & 5025.56                                      \\
                                                                                    &                                                         & 4                                                & 93.00\%                                  & 97.00\%                                   & 21.67\%                                  & 68.00\%                                  & 78.67\%                                  & 99.00\%                                  & 43.33\%                                  & 85.33\%                                  & 58.82                                 & 15.96                            & 2045.77                                      & 2029.80                                      \\
\multirow{-16}{*}{\textbf{\begin{tabular}[c]{@{}l@{}}GPT2XL\\ (1.5B)\end{tabular}}} & \multirow{-5}{*}{\textbf{PMET}}                         & 5                                                & 96.00\%                                  & 98.33\%                                   & 22.67\%                                  & 69.00\%                                  & 77.67\%                                  & 99.00\%                                  & 44.00\%                                  & 86.67\%                                  & 59.47                                 & 3.90                             & \multicolumn{1}{l}{228.44}                         & \multicolumn{1}{l}{224.53}                         \\
 & & 6& 97.33\%& 99.67\%& 22.67\%& 69.67\%& 77.00\%& 99.00\%	& 44.67\%& 86.67\%& 60.46& 1.81& 50.77&48.96\\
 & & 7& 98.67\%& 99.67\%& 23.67\%& 70.00\%& 77.00\%& 99.00\%& 45.67\%& 87.00\%& 61.52& 1.42& 22.44&21.02\\
 & & 8& 98.67\%& 99.67\%& 24.00\%& 70.00\%& 76.00\%& 99.00\%& 46.00\%& 87.00\%& 62.58& 1.35& 12.10&10.75\\
 & & 9& 99.33\%& 99.67\%& 24.00\%& 70.33\%& 76.00\%& 99.00\%& 46.00\%& 87.00\%& 63.20& 1.32& 3.03&1.71\\
 & & \cellcolor[HTML]{9AFF99}10& \cellcolor[HTML]{9AFF99}\textbf{99.33\%}& \cellcolor[HTML]{9AFF99}\textbf{99.67\%}& \cellcolor[HTML]{9AFF99}\textbf{24.67\%}& \cellcolor[HTML]{9AFF99}\textbf{70.33\%}& \cellcolor[HTML]{9AFF99}75.33\%& \cellcolor[HTML]{9AFF99}99.00\%& \cellcolor[HTML]{9AFF99}\textbf{47.00\%}& \cellcolor[HTML]{9AFF99}\textbf{87.33\%}& \cellcolor[HTML]{9AFF99}63.92& \cellcolor[HTML]{9AFF99}1.31& \cellcolor[HTML]{9AFF99}1.73&\cellcolor[HTML]{9AFF99}0.42\\ \midrule
                                                                                    & \textbf{Unedited}                                       & 0                                                & 9.33\%                                   & 38.00\%                                   & 9.00\%                                   & 38.33\%                                  & \multicolumn{1}{l}{}                     & 82.00\%                                  & \multicolumn{1}{l}{}                     & 37.33\%                                  & 39.80                                 & \multicolumn{1}{l}{}             & 5.98E+05                                     & \multicolumn{1}{l}{}                         \\ \cmidrule(l){2-15} 
                                                                                    & \cellcolor[HTML]{FFCCC9}                                & \cellcolor[HTML]{FFCCC9}1                        & \cellcolor[HTML]{FFCCC9}1.00\%           & \cellcolor[HTML]{FFCCC9}57.00\%           & \cellcolor[HTML]{FFCCC9}1.00\%           & \cellcolor[HTML]{FFCCC9}55.00\%          & \cellcolor[HTML]{FFCCC9}0.00\%           & \cellcolor[HTML]{FFCCC9}76.00\%          & \cellcolor[HTML]{FFCCC9}1.00\%           & \cellcolor[HTML]{FFCCC9}61.00\%          & \cellcolor[HTML]{FFCCC9}2.15E+05      & \cellcolor[HTML]{FFCCC9}1.19E+08 & \cellcolor[HTML]{FFCCC9}1.21E+14             & \cellcolor[HTML]{FFCCC9}1.21E+14             \\
                                                                                    & \cellcolor[HTML]{FFCCC9}                                & \cellcolor[HTML]{FFCCC9}2                        & \cellcolor[HTML]{FFCCC9}1.00\%           & \cellcolor[HTML]{FFCCC9}67.00\%           & \cellcolor[HTML]{FFCCC9}2.00\%           & \cellcolor[HTML]{FFCCC9}62.00\%          & \cellcolor[HTML]{FFCCC9}0.00\%           & \cellcolor[HTML]{FFCCC9}71.00\%          & \cellcolor[HTML]{FFCCC9}1.00\%           & \cellcolor[HTML]{FFCCC9}66.00\%          & \cellcolor[HTML]{FFCCC9}8.51E+05      & \cellcolor[HTML]{FFCCC9}1.75E+12 & \cellcolor[HTML]{FFCCC9}4.17E+06             & \cellcolor[HTML]{FFCCC9}-1.75E+12            \\
                                                                                    & \cellcolor[HTML]{FFCCC9}                                & \cellcolor[HTML]{FFCCC9}3                        & \cellcolor[HTML]{FFCCC9}1.00\%           & \cellcolor[HTML]{FFCCC9}71.00\%           & \cellcolor[HTML]{FFCCC9}1.00\%           & \cellcolor[HTML]{FFCCC9}63.00\%          & \cellcolor[HTML]{FFCCC9}0.00\%           & \cellcolor[HTML]{FFCCC9}71.00\%          & \cellcolor[HTML]{FFCCC9}1.00\%           & \cellcolor[HTML]{FFCCC9}68.00\%          & \cellcolor[HTML]{FFCCC9}9.00E+05      & \cellcolor[HTML]{FFCCC9}8.09E+05 & \cellcolor[HTML]{FFCCC9}2.44E+06             & \cellcolor[HTML]{FFCCC9}1.64E+06             \\
                                                                                    & \cellcolor[HTML]{FFCCC9}                                & \cellcolor[HTML]{FFCCC9}4                        & \cellcolor[HTML]{FFCCC9}1.00\%           & \cellcolor[HTML]{FFCCC9}72.00\%           & \cellcolor[HTML]{FFCCC9}1.00\%           & \cellcolor[HTML]{FFCCC9}64.00\%          & \cellcolor[HTML]{FFCCC9}0.00\%           & \cellcolor[HTML]{FFCCC9}73.00\%          & \cellcolor[HTML]{FFCCC9}1.00\%           & \cellcolor[HTML]{FFCCC9}69.00\%          & \cellcolor[HTML]{FFCCC9}8.29E+05      & \cellcolor[HTML]{FFCCC9}7.36E+05 & \cellcolor[HTML]{FFCCC9}8.36E+05             & \cellcolor[HTML]{FFCCC9}9.97E+04             \\
                                                                                    & \multirow{-5}{*}{\cellcolor[HTML]{FFCCC9}\textbf{ROME}} & \cellcolor[HTML]{FFCCC9}5                        & \cellcolor[HTML]{FFCCC9}1.00\%           & \cellcolor[HTML]{FFCCC9}72.00\%           & \cellcolor[HTML]{FFCCC9}1.00\%           & \cellcolor[HTML]{FFCCC9}65.00\%          & \cellcolor[HTML]{FFCCC9}0.00\%           & \cellcolor[HTML]{FFCCC9}71.00\%          & \cellcolor[HTML]{FFCCC9}1.00\%           & \cellcolor[HTML]{FFCCC9}69.00\%          & \cellcolor[HTML]{FFCCC9}7.10E+05      & \cellcolor[HTML]{FFCCC9}3.36E+05 & \multicolumn{1}{l}{\cellcolor[HTML]{FFCCC9}} & \multicolumn{1}{l}{\cellcolor[HTML]{FFCCC9}} \\ \cmidrule(l){2-15} 
                                                                                    &                                                         & 1                                                & 99.00\%                                  & 100.00\%                                  & 75.00\%                                  & 95.67\%                                  & \textbf{69.33\%}                         & \textbf{89.33\%}                         & 77.67\%                                  & 94.33\%                                  & 42.20                                 & 1.03                             & 2.25                                         & 1.22                                         \\
                                                                                    &                                                         & \cellcolor[HTML]{9AFF99}2                        & \cellcolor[HTML]{9AFF99}\textbf{99.33\%} & \cellcolor[HTML]{9AFF99}\textbf{100.00\%} & \cellcolor[HTML]{9AFF99}\textbf{80.67\%} & \cellcolor[HTML]{9AFF99}\textbf{98.00\%} & \cellcolor[HTML]{9AFF99}66.33\%          & \cellcolor[HTML]{9AFF99}88.33\%          & \cellcolor[HTML]{9AFF99}\textbf{79.00\%} & \cellcolor[HTML]{9AFF99}\textbf{95.00\%} & \cellcolor[HTML]{9AFF99}43.92         & \cellcolor[HTML]{9AFF99}1.02     & \cellcolor[HTML]{9AFF99}1.05                 & \cellcolor[HTML]{9AFF99}0.03                 \\
                                                                                    &                                                         & 3                                                & 99.67\%                                  & 100.00\%                                  & 82.00\%                                  & 98.33\%                                  & 65.33\%                                  & 88.33\%                                  & 79.00\%                                  & 95.00\%                                  & 46.33                                 & 1.02                             & 2.88                                         & 1.86                                         \\
                                                                                    &                                                         & 4                                                & 99.67\%                                  & 100.00\%                                  & 83.67\%                                  & 98.33\%                                  & 64.67\%                                  & 88.00\%                                  & 79.33\%                                  & 95.00\%                                  & 46.95                                 & 1.01                             & 1.04                                         & 0.03                                         \\
                                                                                    & \multirow{-5}{*}{\textbf{MEMIT}}                        & 5                                                & 99.67\%                                  & 100.00\%                                  & 83.67\%                                  & 98.33\%                                  & 64.33\%                                  & 88.00\%                                  & 79.33\%                                  & 95.00\%                                  & 47.20                                 & 1.01                             & \multicolumn{1}{l}{}                         & \multicolumn{1}{l}{}                         \\ \cmidrule(l){2-15} 
                                                                                    &                                                         & 1                                                & 98.00\%                                  & \textbf{99.67\%}                          & 76.00\%                                  & 95.00\%                                  & \textbf{68.33\%}                         & \textbf{88.67\%}                         & 77.67\%                                  & 93.67\%                                  & 41.27                                 & 1.08                             & 2.24                                         & 1.15                                         \\
                                                                                    &                                                         & 2                                                & 98.67\%                                  & 99.67\%                                   & 75.67\%                                  & 95.00\%                                  & 68.33\%                                  & 89.00\%                                  & 78.00\%                                  & 94.33\%                                  & 41.16                                 & 1.06                             & 5.29                                         & 4.23                                         \\
                                                                                    &                                                         & \cellcolor[HTML]{9AFF99}3                        & \cellcolor[HTML]{9AFF99}\textbf{99.00\%} & \cellcolor[HTML]{9AFF99}\textbf{99.67\%}  & \cellcolor[HTML]{9AFF99}\textbf{76.67\%} & \cellcolor[HTML]{9AFF99}\textbf{95.67\%} & \cellcolor[HTML]{9AFF99}\textbf{68.33\%} & \cellcolor[HTML]{9AFF99}\textbf{88.67\%} & \cellcolor[HTML]{9AFF99}\textbf{78.33\%} & \cellcolor[HTML]{9AFF99}\textbf{94.00\%} & \cellcolor[HTML]{9AFF99}41.15         & \cellcolor[HTML]{9AFF99}1.06     & \cellcolor[HTML]{9AFF99}1.11                 & \cellcolor[HTML]{9AFF99}0.05                 \\
                                                                                    &                                                         & 4                                                & 99.33\%                                  & 99.67\%                                   & 77.00\%                                  & 95.67\%                                  & 68.33\%                                  & 88.67\%                                  & 78.33\%                                  & 94.00\%                                  & 41.19                                 & 1.06                             & 1.08                                         & 0.02                                         \\
\multirow{-16}{*}{\textbf{\begin{tabular}[c]{@{}l@{}}GPTJ\\ (6B)\end{tabular}}}     & \multirow{-5}{*}{\textbf{PMET}}                         & 5                                                & 99.33\%                                  & 99.67\%                                   & 77.00\%                                  & 95.67\%                                  & 68.00\%                                  & 89.00\%                                  & 78.33\%                                  & 94.33\%                                  & 41.28                                 & 1.06                             & \multicolumn{1}{l}{}                         & \multicolumn{1}{l}{}                         \\
 & & & & & & & & & & & & & &\\ \midrule
                                                                                    & \textbf{Unedited}                                       & 0                                                & 15.00\%                                  & 13.67\%                                   & 15.00\%                                  & 15.00\%                                  & \multicolumn{1}{l}{}                     & 84.33\%                                  & \multicolumn{1}{l}{}                     & 19.67\%                                  & 30.63                                 & \multicolumn{1}{l}{}             & 2789.16                                      & \multicolumn{1}{l}{}                         \\ \cmidrule(l){2-15} 
                                                                                    & \cellcolor[HTML]{FFCCC9}                                & \cellcolor[HTML]{FFCCC9}1                        & \cellcolor[HTML]{FFCCC9}0.00\%           & \cellcolor[HTML]{FFCCC9}48.00\%           & \cellcolor[HTML]{FFCCC9}0.00\%           & \cellcolor[HTML]{FFCCC9}49.00\%          & \cellcolor[HTML]{FFCCC9}0.00\%           & \cellcolor[HTML]{FFCCC9}76.00\%          & \cellcolor[HTML]{FFCCC9}0.00\%           & \cellcolor[HTML]{FFCCC9}55.00\%          & \cellcolor[HTML]{FFCCC9}1.45E+04      & \cellcolor[HTML]{FFCCC9}3.80E+03 & \cellcolor[HTML]{FFCCC9}8.14E+05             & \cellcolor[HTML]{FFCCC9}8.10E+05             \\
                                                                                    & \cellcolor[HTML]{FFCCC9}                                & \cellcolor[HTML]{FFCCC9}2                        & \cellcolor[HTML]{FFCCC9}0.00\%           & \cellcolor[HTML]{FFCCC9}56.00\%           & \cellcolor[HTML]{FFCCC9}0.00\%           & \cellcolor[HTML]{FFCCC9}54.00\%          & \cellcolor[HTML]{FFCCC9}0.00\%           & \cellcolor[HTML]{FFCCC9}64.00\%          & \cellcolor[HTML]{FFCCC9}0.00\%           & \cellcolor[HTML]{FFCCC9}58.00\%          & \cellcolor[HTML]{FFCCC9}7.27E+04      & \cellcolor[HTML]{FFCCC9}6.81E+05 & \cellcolor[HTML]{FFCCC9}2.34E+08             & \cellcolor[HTML]{FFCCC9}2.33E+08             \\
                                                                                    & \cellcolor[HTML]{FFCCC9}                                & \cellcolor[HTML]{FFCCC9}3                        & \cellcolor[HTML]{FFCCC9}0.00\%           & \cellcolor[HTML]{FFCCC9}55.00\%           & \cellcolor[HTML]{FFCCC9}0.00\%           & \cellcolor[HTML]{FFCCC9}53.00\%          & \cellcolor[HTML]{FFCCC9}0.00\%           & \cellcolor[HTML]{FFCCC9}57.00\%          & \cellcolor[HTML]{FFCCC9}0.00\%           & \cellcolor[HTML]{FFCCC9}55.00\%          & \cellcolor[HTML]{FFCCC9}3.86E+05      & \cellcolor[HTML]{FFCCC9}1.55E+08 & \cellcolor[HTML]{FFCCC9}7.55E+09             & \cellcolor[HTML]{FFCCC9}7.39E+09             \\
                                                                                    & \cellcolor[HTML]{FFCCC9}                                & \cellcolor[HTML]{FFCCC9}4                        & \cellcolor[HTML]{FFCCC9}0.00\%           & \cellcolor[HTML]{FFCCC9}57.00\%           & \cellcolor[HTML]{FFCCC9}0.00\%           & \cellcolor[HTML]{FFCCC9}55.00\%          & \cellcolor[HTML]{FFCCC9}0.00\%           & \cellcolor[HTML]{FFCCC9}54.00\%          & \cellcolor[HTML]{FFCCC9}0.00\%           & \cellcolor[HTML]{FFCCC9}55.00\%          & \cellcolor[HTML]{FFCCC9}1.26E+06      & \cellcolor[HTML]{FFCCC9}6.04E+09 & \cellcolor[HTML]{FFCCC9}5.81E+10             & \cellcolor[HTML]{FFCCC9}5.20E+10             \\
                                                                                    & \multirow{-5}{*}{\cellcolor[HTML]{FFCCC9}\textbf{ROME}} & \cellcolor[HTML]{FFCCC9}5                        & \cellcolor[HTML]{FFCCC9}0.00\%           & \cellcolor[HTML]{FFCCC9}56.00\%           & \cellcolor[HTML]{FFCCC9}0.00\%           & \cellcolor[HTML]{FFCCC9}52.00\%          & \cellcolor[HTML]{FFCCC9}0.00\%           & \cellcolor[HTML]{FFCCC9}57.00\%          & \cellcolor[HTML]{FFCCC9}0.00\%           & \cellcolor[HTML]{FFCCC9}55.00\%          & \cellcolor[HTML]{FFCCC9}1.38E+06      & \cellcolor[HTML]{FFCCC9}4.36E+10 & \multicolumn{1}{l}{\cellcolor[HTML]{FFCCC9}} & \multicolumn{1}{l}{\cellcolor[HTML]{FFCCC9}} \\ \cmidrule(l){2-15} 
                                                                                    &                                                         & 1                                                & 91.67\%                                  & 98.00\%                                   & 70.33\%                                  & 93.33\%                                  & 29.33\%                                  & 67.33\%                                  & 50.67\%                                  & 83.67\%                                  & 42.10                                 & 1.01                             & 199.80                                       & 198.79                                       \\
                                                                                    &                                                         & \cellcolor[HTML]{FFCCC9}2                        & \cellcolor[HTML]{FFCCC9}14.33\%          & \cellcolor[HTML]{FFCCC9}79.00\%           & \cellcolor[HTML]{FFCCC9}9.67\%           & \cellcolor[HTML]{FFCCC9}73.67\%          & \cellcolor[HTML]{FFCCC9}6.67\%           & \cellcolor[HTML]{FFCCC9}70.67\%          & \cellcolor[HTML]{FFCCC9}9.00\%           & \cellcolor[HTML]{FFCCC9}74.67\%          & \cellcolor[HTML]{FFCCC9}9.37E+03      & \cellcolor[HTML]{FFCCC9}16.77    & \cellcolor[HTML]{FFCCC9}4681.01              & \cellcolor[HTML]{FFCCC9}4664.24              \\
                                                                                    &                                                         & \cellcolor[HTML]{FFCCC9}3                        & \cellcolor[HTML]{FFCCC9}26.67\%          & \cellcolor[HTML]{FFCCC9}89.67\%           & \cellcolor[HTML]{FFCCC9}12.33\%          & \cellcolor[HTML]{FFCCC9}82.67\%          & \cellcolor[HTML]{FFCCC9}5.67\%           & \cellcolor[HTML]{FFCCC9}66.67\%          & \cellcolor[HTML]{FFCCC9}9.67\%           & \cellcolor[HTML]{FFCCC9}78.33\%          & \cellcolor[HTML]{FFCCC9}4.35E+04      & \cellcolor[HTML]{FFCCC9}8.40     & \cellcolor[HTML]{FFCCC9}884.21               & \cellcolor[HTML]{FFCCC9}875.80               \\
                                                                                    &                                                         & \cellcolor[HTML]{FFCCC9}4                        & \cellcolor[HTML]{FFCCC9}22.00\%          & \cellcolor[HTML]{FFCCC9}94.67\%           & \cellcolor[HTML]{FFCCC9}5.67\%           & \cellcolor[HTML]{FFCCC9}82.00\%          & \cellcolor[HTML]{FFCCC9}5.67\%           & \cellcolor[HTML]{FFCCC9}68.67\%          & \cellcolor[HTML]{FFCCC9}7.00\%           & \cellcolor[HTML]{FFCCC9}80.33\%          & \cellcolor[HTML]{FFCCC9}8.63E+04      & \cellcolor[HTML]{FFCCC9}1.89     & \cellcolor[HTML]{FFCCC9}1381.59              & \cellcolor[HTML]{FFCCC9}1379.71              \\
                                                                                    & \multirow{-5}{*}{\textbf{MEMIT}}                        & \cellcolor[HTML]{FFCCC9}5                        & \cellcolor[HTML]{FFCCC9}38.33\%          & \cellcolor[HTML]{FFCCC9}95.67\%           & \cellcolor[HTML]{FFCCC9}10.67\%          & \cellcolor[HTML]{FFCCC9}77.67\%          & \cellcolor[HTML]{FFCCC9}4.67\%           & \cellcolor[HTML]{FFCCC9}66.67\%          & \cellcolor[HTML]{FFCCC9}8.67\%           & \cellcolor[HTML]{FFCCC9}78.33\%          & \cellcolor[HTML]{FFCCC9}7.60E+04      & \cellcolor[HTML]{FFCCC9}2.37     & \multicolumn{1}{l}{\cellcolor[HTML]{FFCCC9}} & \multicolumn{1}{l}{\cellcolor[HTML]{FFCCC9}} \\ \cmidrule(l){2-15} 
                                                                                    &                                                         & 1                                                & 94.33\%                                  & 97.00\%                                   & 68.33\%                                  & 86.67\%                                  & \textbf{76.33\%}                         & \textbf{89.00\%}                         & 77.33\%                                  & 90.33\%                                  & 30.73                                 & 1.09                             & 4.41                                         & 3.32                                         \\
                                                                                    &                                                         & \cellcolor[HTML]{9AFF99}2                        & \cellcolor[HTML]{9AFF99}\textbf{95.33\%} & \cellcolor[HTML]{9AFF99}\textbf{98.33\%}  & \cellcolor[HTML]{9AFF99}\textbf{70.00\%} & \cellcolor[HTML]{9AFF99}\textbf{88.67\%} & \cellcolor[HTML]{9AFF99}75.33\%          & \cellcolor[HTML]{9AFF99}88.67\%          & \cellcolor[HTML]{9AFF99}\textbf{78.00\%} & \cellcolor[HTML]{9AFF99}\textbf{91.67\%} & \cellcolor[HTML]{9AFF99}30.76         & \cellcolor[HTML]{9AFF99}1.14     & \cellcolor[HTML]{9AFF99}1.23                 & \cellcolor[HTML]{9AFF99}0.09                 \\
                                                                                    &                                                         & 3                                                & 95.33\%                                  & 98.33\%                                   & 69.67\%                                  & 88.67\%                                  & 75.33\%                                  & 88.67\%                                  & 78.33\%                                  & 91.67\%                                  & 30.78                                 & 1.14                             & 1.30                                         & 0.16                                         \\
                                                                                    &                                                         & 4                                                & 95.33\%                                  & 98.33\%                                   & 70.00\%                                  & 89.00\%                                  & 75.33\%                                  & 88.67\%                                  & 78.33\%                                  & 91.67\%                                  & 30.79                                 & 1.14                             & 1.14                                         & 0.00                                         \\
\multirow{-16}{*}{\textbf{\begin{tabular}[c]{@{}l@{}}Llama2\\ (7B)\end{tabular}}}   & \multirow{-5}{*}{\textbf{PMET}}                         & 5                                                & 95.33\%                                  & 98.33\%                                   & 70.00\%                                  & 89.00\%                                  & 75.33\%                                  & 88.67\%                                  & 78.33\%                                  & 91.67\%                                  & 30.79                                 & 1.14                             & \multicolumn{1}{l}{}                         & \multicolumn{1}{l}{}                         \\ \bottomrule
\end{tabular}
}
\caption{Iterative model editing results on C{\smaller OUNTER}F{\smaller ACT} for at most 10 iterations (denoted by k). We compare the evaluation metrics of iteration that met stopping criterion $|\Delta p_k |\leq 1$ (green rows) to that of their corresponding first iteration and \textbf{bold} the higher value. PMET on GPT-2 XL require more than 5 iterations to achieve our stopping criteria. While ROME is known to collapse (red rows), we observed a unique case of collapse with Llama-2 (7B) specifically when using MEMIT. We discuss this in Section \ref{sec:result}.
} 
\label{tab:mcf-results-all}
\end{table*}
\begin{table*}[h]
\resizebox{\textwidth}{!}{%
\begin{tabular}{@{}llrrrrrrrrrrrrr@{}}
\toprule
                                                                                      &                                                         & \multicolumn{1}{l}{}                             & \multicolumn{2}{l}{\textbf{Efficacy ($\uparrow$)}}                                                 & \multicolumn{2}{l}{\textbf{Generalization ($\uparrow$)}}                                          & \multicolumn{2}{l}{\textbf{Specificity ($\uparrow$)}}                                            & \multicolumn{2}{l}{\textbf{Score ($\uparrow$)}}                                                  & \multicolumn{3}{l}{\textbf{Perplexity ($\downarrow$)}}                                                                                 & \multicolumn{1}{l}{\textbf{$|\Delta p_k|$ ($\downarrow$)}}              \\ \cmidrule(l){4-15} 
\multirow{-2}{*}{\textbf{Model}}                                                      & \multirow{-2}{*}{\textbf{Algo}}                         & \multicolumn{1}{l}{\multirow{-2}{*}{\textbf{k}}} & \multicolumn{1}{l}{\textbf{Accuracy}}     & \multicolumn{1}{l}{\textbf{Success}}      & \multicolumn{1}{l}{\textbf{Accuracy}}    & \multicolumn{1}{l}{\textbf{Success}}      & \multicolumn{1}{l}{\textbf{Accuracy}}    & \multicolumn{1}{l}{\textbf{Success}}     & \multicolumn{1}{l}{\textbf{Accuracy}}    & \multicolumn{1}{l}{\textbf{Success}}     & \multicolumn{1}{l}{\textbf{ME-PPL-50}} & \multicolumn{1}{l}{\textbf{$p(\theta_k, \bar h_{t,k}^{l_c})$}}  & \multicolumn{1}{l}{\textbf{$p(\theta_{k+1}, \hat h^{l_c}_{t,k})$}}              & \multicolumn{1}{l}{\textbf{}}         \\ \midrule
                                                                                      & \textbf{Unedited}                                       & 0                                                & 22.00\%                                   & 87.33\%                                   & 21.00\%                                  & 86.33\%                                   & \multicolumn{1}{l}{}                     & 56.33\%                                  & \multicolumn{1}{l}{}                     & 73.67\%                                  & 54.66                                 & \multicolumn{1}{l}{}             & 195351.10                                    & \multicolumn{1}{l}{}                         \\ \cmidrule(l){2-15} 
                                                                                      & \cellcolor[HTML]{FFCCC9}                                & \cellcolor[HTML]{FFCCC9}1                        & \cellcolor[HTML]{FFCCC9}3.00\%            & \cellcolor[HTML]{FFCCC9}40.00\%           & \cellcolor[HTML]{FFCCC9}3.00\%           & \cellcolor[HTML]{FFCCC9}37.00\%           & \cellcolor[HTML]{FFCCC9}1.00\%           & \cellcolor[HTML]{FFCCC9}77.00\%          & \cellcolor[HTML]{FFCCC9}2.00\%           & \cellcolor[HTML]{FFCCC9}46.00\%          & \cellcolor[HTML]{FFCCC9}8.59E+03      & \cellcolor[HTML]{FFCCC9}1.21E+04 & \cellcolor[HTML]{FFCCC9}2.64E+05             & \cellcolor[HTML]{FFCCC9}2.52E+05             \\
                                                                                      & \cellcolor[HTML]{FFCCC9}                                & \cellcolor[HTML]{FFCCC9}2                        & \cellcolor[HTML]{FFCCC9}0.00\%            & \cellcolor[HTML]{FFCCC9}97.00\%           & \cellcolor[HTML]{FFCCC9}0.00\%           & \cellcolor[HTML]{FFCCC9}97.00\%           & \cellcolor[HTML]{FFCCC9}29.00\%          & \cellcolor[HTML]{FFCCC9}71.00\%          & \cellcolor[HTML]{FFCCC9}0.00\%           & \cellcolor[HTML]{FFCCC9}86.00\%          & \cellcolor[HTML]{FFCCC9}5.20E+03      & \cellcolor[HTML]{FFCCC9}5.75E+04 & \cellcolor[HTML]{FFCCC9}2.67E+05             & \cellcolor[HTML]{FFCCC9}2.10E+05             \\
                                                                                      & \cellcolor[HTML]{FFCCC9}                                & \cellcolor[HTML]{FFCCC9}3                        & \cellcolor[HTML]{FFCCC9}0.00\%            & \cellcolor[HTML]{FFCCC9}21.00\%           & \cellcolor[HTML]{FFCCC9}0.00\%           & \cellcolor[HTML]{FFCCC9}18.00\%           & \cellcolor[HTML]{FFCCC9}0.00\%           & \cellcolor[HTML]{FFCCC9}76.00\%          & \cellcolor[HTML]{FFCCC9}0.00\%           & \cellcolor[HTML]{FFCCC9}25.00\%          & \cellcolor[HTML]{FFCCC9}1.21E+04      & \cellcolor[HTML]{FFCCC9}9.44E+04 & \cellcolor[HTML]{FFCCC9}3.27E+05             & \cellcolor[HTML]{FFCCC9}2.32E+05             \\
                                                                                      & \cellcolor[HTML]{FFCCC9}                                & \cellcolor[HTML]{FFCCC9}4                        & \cellcolor[HTML]{FFCCC9}0.00\%            & \cellcolor[HTML]{FFCCC9}100.00\%          & \cellcolor[HTML]{FFCCC9}0.00\%           & \cellcolor[HTML]{FFCCC9}100.00\%          & \cellcolor[HTML]{FFCCC9}0.00\%           & \cellcolor[HTML]{FFCCC9}68.00\%          & \cellcolor[HTML]{FFCCC9}0.00\%           & \cellcolor[HTML]{FFCCC9}86.00\%          & \cellcolor[HTML]{FFCCC9}6.01E+03      & \cellcolor[HTML]{FFCCC9}8.27E+04 & \cellcolor[HTML]{FFCCC9}1.02E+06             & \cellcolor[HTML]{FFCCC9}9.36E+05             \\
                                                                                      & \multirow{-5}{*}{\cellcolor[HTML]{FFCCC9}\textbf{ROME}} & \cellcolor[HTML]{FFCCC9}5                        & \cellcolor[HTML]{FFCCC9}0.00\%            & \cellcolor[HTML]{FFCCC9}0.00\%            & \cellcolor[HTML]{FFCCC9}0.00\%           & \cellcolor[HTML]{FFCCC9}0.00\%            & \cellcolor[HTML]{FFCCC9}0.00\%           & \cellcolor[HTML]{FFCCC9}67.00\%          & \cellcolor[HTML]{FFCCC9}0.00\%           & \cellcolor[HTML]{FFCCC9}67.00\%          & \cellcolor[HTML]{FFCCC9}1.85E+04      & \cellcolor[HTML]{FFCCC9}4.38E+05 & \multicolumn{1}{l}{\cellcolor[HTML]{FFCCC9}} & \multicolumn{1}{l}{\cellcolor[HTML]{FFCCC9}} \\ \cmidrule(l){2-15} 
                                                                                      &                                                         & 1                                                & 69.00\%                                   & \textbf{100.00\%}                         & 58.67\%                                  & 99.67\%                                   & \textbf{32.33\%}                         & \textbf{85.00\%}                         & \textbf{48.33\%}                         & \textbf{94.00\%}                         & 61.74                                 & 1.02                             & 2035.34                                      & 2034.31                                      \\
                                                                                      &                                                         & 2                                                & 98.33\%                                   & 100.00\%                                  & 87.00\%                                  & 100.00\%                                  & 24.33\%                                  & 84.33\%                                  & 47.67\%                                  & 94.00\%                                  & 67.68                                 & 1.02                             & 40.41                                        & 39.39                                        \\
                                                                                      &                                                         & \cellcolor[HTML]{9AFF99}3                        & \cellcolor[HTML]{9AFF99}\textbf{100.00\%} & \cellcolor[HTML]{9AFF99}\textbf{100.00\%} & \cellcolor[HTML]{9AFF99}\textbf{88.00\%} & \cellcolor[HTML]{9AFF99}\textbf{100.00\%} & \cellcolor[HTML]{9AFF99}23.33\%          & \cellcolor[HTML]{9AFF99}84.00\%          & \cellcolor[HTML]{9AFF99}46.67\%          & \cellcolor[HTML]{9AFF99}\textbf{94.00\%} & \cellcolor[HTML]{9AFF99}69.40         & \cellcolor[HTML]{9AFF99}1.02     & \cellcolor[HTML]{9AFF99}1.05                 & \cellcolor[HTML]{9AFF99}0.03                 \\
                                                                                      &                                                         & 4                                                & 100.00\%                                  & 100.00\%                                  & 89.00\%                                  & 100.00\%                                  & 22.00\%                                  & 84.00\%                                  & 45.00\%                                  & 94.00\%                                  & 70.02                                 & 1.02                             & 1.03                                         & 0.01                                         \\
                                                                                      & \multirow{-5}{*}{\textbf{MEMIT}}                        & 5                                                & 100.00\%                                  & 100.00\%                                  & 89.67\%                                  & 100.00\%                                  & 21.67\%                                  & 84.00\%                                  & 44.33\%                                  & 94.00\%                                  & 70.16                                 & 1.02                             & \multicolumn{1}{l}{}                         & \multicolumn{1}{l}{}                         \\ \cmidrule(l){2-15} 
                                                                                      &                                                         & 1                                                & 34.67\%                                   & 97.67\%                                   & 30.33\%                                  & 95.67\%                                   & 45.00\%                                  & 85.67\%                                  & 35.33\%                                  & 93.00\%                                  & 56.77                                 & 5375.18                          & 116427.63                                    & 111052.45                                    \\
                                                                                      &                                                         & 2                                                & 67.00\%                                   & 100.00\%                                  & 52.00\%                                  & 99.33\%                                   & 38.33\%                                  & 85.33\%                                  & 49.67\%                                  & 94.33\%                                  & 60.44                                 & 15.17                            & 3109.90                                      & 3094.74                                      \\
                                                                                      &                                                         & 3                                                & 89.67\%                                   & 100.00\%                                  & 66.33\%                                  & 99.67\%                                   & 35.00\%                                  & 85.00\%                                  & 55.00\%                                  & 94.33\%                                  & 62.37                                 & 2.16                             & 491.43                                       & 489.28                                       \\
                                                                                      &                                                         & 4                                                & 94.33\%                                   & 100.00\%                                  & 69.67\%                                  & 100.00\%                                  & 33.33\%                                  & 84.67\%                                  & 54.67\%                                  & 94.00\%                                  & 64.42                                 & 1.34                             & 50.04                                        & 48.69                                        \\
                                                                                      &                                                         & 5                                                & 98.00\%                                   & 100.00\%                                  & 73.00\%                                  & 100.00\%                                  & 32.00\%                                  & 84.67\%                                  & 54.00\%                                  & 94.00\%                                  & 65.22                                 & 1.25                             & 2.70                                         & 1.45                                         \\
                                                                                      &                                                         & \cellcolor[HTML]{9AFF99}6                        & \cellcolor[HTML]{9AFF99}\textbf{99.33\%}  & \cellcolor[HTML]{9AFF99}\textbf{100.00\%} & \cellcolor[HTML]{9AFF99}\textbf{73.67\%} & \cellcolor[HTML]{9AFF99}\textbf{100.00\%} & \cellcolor[HTML]{9AFF99}31.00\%          & \cellcolor[HTML]{9AFF99}84.00\%          & \cellcolor[HTML]{9AFF99}\textbf{54.00\%} & \cellcolor[HTML]{9AFF99}\textbf{94.00\%} & \cellcolor[HTML]{9AFF99}65.93         & \cellcolor[HTML]{9AFF99}1.23     & \cellcolor[HTML]{9AFF99}1.41                 & \cellcolor[HTML]{9AFF99}0.19                 \\
                                                                                      &                                                         & 7                                                & 99.00\%                                   & 100.00\%                                  & 75.00\%                                  & 100.00\%                                  & 30.33\%                                  & 84.00\%                                  & 53.33\%                                  & 94.00\%                                  & 66.41                                 & 1.22                             & 1.30                                         & 0.08                                         \\
                                                                                      &                                                         & 8                                                & 99.67\%                                   & 100.00\%                                  & 74.67\%                                  & 100.00\%                                  & 30.00\%                                  & 84.00\%                                  & 53.00\%                                  & 94.00\%                                  & 66.71                                 & 1.21                             & 1.25                                         & 0.04                                         \\
                                                                                      &                                                         & 9                                                & 99.67\%                                   & 100.00\%                                  & 75.00\%                                  & 100.00\%                                  & 29.33\%                                  & 84.00\%                                  & 52.33\%                                  & 94.00\%                                  & 66.77                                 & 1.22                             & 1.22                                         & 0.01                                         \\
\multirow{-21}{*}{\textbf{\begin{tabular}[c]{@{}l@{}}GPT-2 XL\\ (1.5B)\end{tabular}}} & \multirow{-10}{*}{\textbf{PMET}}                        & 10                                               & 100.00\%                                  & 100.00\%                                  & 75.33\%                                  & 100.00\%                                  & 29.33\%                                  & 84.00\%                                  & 52.33\%                                  & 94.00\%                                  & 66.89                                 & 1.20                             & 1.21                                         & 0.01                                         \\ \midrule
                                                                                      & \textbf{Unedited}                                       & 0                                                & 27.33\%                                   & 91.00\%                                   & 26.33\%                                  & 90.00\%                                   & \multicolumn{1}{l}{}                     & 60.00\%                                  & \multicolumn{1}{l}{}                     & 77.33\%                                  & 39.80                                 & \multicolumn{1}{l}{}             & 5.02E+04                                     & 5.02E+04                                     \\ \cmidrule(l){2-15} 
                                                                                      & \cellcolor[HTML]{FFCCC9}                                & \cellcolor[HTML]{FFCCC9}1                        & \cellcolor[HTML]{FFCCC9}5.00\%            & \cellcolor[HTML]{FFCCC9}90.00\%           & \cellcolor[HTML]{FFCCC9}5.00\%           & \cellcolor[HTML]{FFCCC9}89.00\%           & \cellcolor[HTML]{FFCCC9}0.00\%           & \cellcolor[HTML]{FFCCC9}65.00\%          & \cellcolor[HTML]{FFCCC9}5.00\%           & \cellcolor[HTML]{FFCCC9}80.00\%          & \cellcolor[HTML]{FFCCC9}3.93E+05      & \cellcolor[HTML]{FFCCC9}1.63E+04 & \cellcolor[HTML]{FFCCC9}1.33E+05             & \cellcolor[HTML]{FFCCC9}1.17E+05             \\
                                                                                      & \cellcolor[HTML]{FFCCC9}                                & \cellcolor[HTML]{FFCCC9}2                        & \cellcolor[HTML]{FFCCC9}13.00\%           & \cellcolor[HTML]{FFCCC9}95.00\%           & \cellcolor[HTML]{FFCCC9}10.00\%          & \cellcolor[HTML]{FFCCC9}94.00\%           & \cellcolor[HTML]{FFCCC9}0.00\%           & \cellcolor[HTML]{FFCCC9}67.00\%          & \cellcolor[HTML]{FFCCC9}11.00\%          & \cellcolor[HTML]{FFCCC9}83.00\%          & \cellcolor[HTML]{FFCCC9}6.19E+05      & \cellcolor[HTML]{FFCCC9}8.72E+03 & \cellcolor[HTML]{FFCCC9}1.58E+04             & \cellcolor[HTML]{FFCCC9}7.04E+03             \\
                                                                                      & \cellcolor[HTML]{FFCCC9}                                & \cellcolor[HTML]{FFCCC9}3                        & \cellcolor[HTML]{FFCCC9}17.00\%           & \cellcolor[HTML]{FFCCC9}99.00\%           & \cellcolor[HTML]{FFCCC9}12.00\%          & \cellcolor[HTML]{FFCCC9}97.00\%           & \cellcolor[HTML]{FFCCC9}0.00\%           & \cellcolor[HTML]{FFCCC9}66.00\%          & \cellcolor[HTML]{FFCCC9}14.00\%          & \cellcolor[HTML]{FFCCC9}84.00\%          & \cellcolor[HTML]{FFCCC9}6.88E+05      & \cellcolor[HTML]{FFCCC9}3.75E+03 & \cellcolor[HTML]{FFCCC9}9.70E+03             & \cellcolor[HTML]{FFCCC9}5.94E+03             \\
                                                                                      & \cellcolor[HTML]{FFCCC9}                                & \cellcolor[HTML]{FFCCC9}4                        & \cellcolor[HTML]{FFCCC9}14.00\%           & \cellcolor[HTML]{FFCCC9}100.00\%          & \cellcolor[HTML]{FFCCC9}11.00\%          & \cellcolor[HTML]{FFCCC9}99.00\%           & \cellcolor[HTML]{FFCCC9}0.00\%           & \cellcolor[HTML]{FFCCC9}64.00\%          & \cellcolor[HTML]{FFCCC9}12.00\%          & \cellcolor[HTML]{FFCCC9}84.00\%          & \cellcolor[HTML]{FFCCC9}1.39E+06      & \cellcolor[HTML]{FFCCC9}4.61E+03 & \cellcolor[HTML]{FFCCC9}1.28E+04             & \cellcolor[HTML]{FFCCC9}8.24E+03             \\
                                                                                      & \multirow{-5}{*}{\cellcolor[HTML]{FFCCC9}\textbf{ROME}} & \cellcolor[HTML]{FFCCC9}5                        & \cellcolor[HTML]{FFCCC9}10.00\%           & \cellcolor[HTML]{FFCCC9}99.00\%           & \cellcolor[HTML]{FFCCC9}8.00\%           & \cellcolor[HTML]{FFCCC9}99.00\%           & \cellcolor[HTML]{FFCCC9}0.00\%           & \cellcolor[HTML]{FFCCC9}63.00\%          & \cellcolor[HTML]{FFCCC9}9.00\%           & \cellcolor[HTML]{FFCCC9}83.00\%          & \cellcolor[HTML]{FFCCC9}3.32E+06      & \cellcolor[HTML]{FFCCC9}7.89E+03 & \multicolumn{1}{l}{\cellcolor[HTML]{FFCCC9}} & \multicolumn{1}{l}{\cellcolor[HTML]{FFCCC9}} \\ \cmidrule(l){2-15} 
                                                                                      &                                                         & 1                                                & 98.67\%                                   & 100.00\%                                  & 89.33\%                                  & \textbf{100.00\%}                         & \textbf{52.67\%}                         & 80.00\%                                  & \textbf{74.67\%}                         & 92.00\%                                  & 41.56                                 & 1.01                             & 2.68                                         & 1.67                                         \\
                                                                                      &                                                         & \cellcolor[HTML]{9AFF99}2                        & \cellcolor[HTML]{9AFF99}\textbf{99.33\%}  & \cellcolor[HTML]{9AFF99}\textbf{100.00\%} & \cellcolor[HTML]{9AFF99}\textbf{92.67\%} & \cellcolor[HTML]{9AFF99}\textbf{100.00\%} & \cellcolor[HTML]{9AFF99}51.67\%          & \cellcolor[HTML]{9AFF99}\textbf{80.33\%} & \cellcolor[HTML]{9AFF99}\textbf{74.67\%} & \cellcolor[HTML]{9AFF99}\textbf{92.33\%} & \cellcolor[HTML]{9AFF99}41.71         & \cellcolor[HTML]{9AFF99}1.02     & \cellcolor[HTML]{9AFF99}1.03                 & \cellcolor[HTML]{9AFF99}0.01                 \\
                                                                                      &                                                         & 3                                                & 100.00\%                                  & 100.00\%                                  & 94.00\%                                  & 100.00\%                                  & 51.33\%                                  & 80.00\%                                  & 75.00\%                                  & 92.33\%                                  & 41.88                                 & 1.02                             & 1.02                                         & 0.00                                         \\
                                                                                      &                                                         & 4                                                & 100.00\%                                  & 100.00\%                                  & 93.67\%                                  & 100.00\%                                  & 51.33\%                                  & 80.00\%                                  & 75.00\%                                  & 92.33\%                                  & 41.88                                 & 1.01                             & 1.01                                         & 0.00                                         \\
                                                                                      & \multirow{-5}{*}{\textbf{MEMIT}}                        & 5                                                & 100.00\%                                  & 100.00\%                                  & 93.67\%                                  & 100.00\%                                  & 51.33\%                                  & 80.00\%                                  & 75.00\%                                  & 92.33\%                                  & 41.87                                 & 1.01                             & \multicolumn{1}{l}{}                         & \multicolumn{1}{l}{}                         \\ \cmidrule(l){2-15} 
                                                                                      &                                                         & 1                                                & 95.67\%                                   & \textbf{100.00\%}                         & 87.33\%                                  & \textbf{100.00\%}                         & 52.00\%                                  & \textbf{80.00\%}                         & 72.67\%                                  & 92.00\%                                  & 41.94                                 & 1.10                             & 5.20                                         & 4.10                                         \\
                                                                                      &                                                         & \cellcolor[HTML]{9AFF99}2                        & \cellcolor[HTML]{9AFF99}\textbf{98.67\%}  & \cellcolor[HTML]{9AFF99}\textbf{100.00\%} & \cellcolor[HTML]{9AFF99}\textbf{88.00\%} & \cellcolor[HTML]{9AFF99}99.67\%           & \cellcolor[HTML]{9AFF99}\textbf{52.00\%} & \cellcolor[HTML]{9AFF99}\textbf{80.00\%} & \cellcolor[HTML]{9AFF99}\textbf{74.00\%} & \cellcolor[HTML]{9AFF99}\textbf{92.00\%} & \cellcolor[HTML]{9AFF99}41.61         & \cellcolor[HTML]{9AFF99}1.06     & \cellcolor[HTML]{9AFF99}1.14                 & \cellcolor[HTML]{9AFF99}0.09                 \\
                                                                                      &                                                         & 3                                                & 99.67\%                                   & 100.00\%                                  & 89.67\%                                  & 100.00\%                                  & 52.00\%                                  & 80.00\%                                  & 74.33\%                                  & 92.00\%                                  & 41.62                                 & 1.06                             & 1.08                                         & 0.02                                         \\
                                                                                      &                                                         & 4                                                & 100.00\%                                  & 100.00\%                                  & 89.67\%                                  & 100.00\%                                  & 52.00\%                                  & 80.00\%                                  & 74.33\%                                  & 92.00\%                                  & 41.59                                 & 1.06                             & 1.06                                         & 0.00                                         \\
\multirow{-16}{*}{\textbf{\begin{tabular}[c]{@{}l@{}}GPT-J\\ (6B)\end{tabular}}}      & \multirow{-5}{*}{\textbf{PMET}}                         & 5                                                & 100.00\%                                  & 100.00\%                                  & 89.67\%                                  & 100.00\%                                  & 52.00\%                                  & 80.00\%                                  & 74.33\%                                  & 92.00\%                                  & 41.58                                 & 1.06                             & \multicolumn{1}{l}{}                         & \multicolumn{1}{l}{}                         \\ \midrule
                                                                                      & \textbf{Unedited}                                       & 0                                                & 38.33\%                                   & 57.00\%                                   & 37.00\%                                  & 56.00\%                                   & \multicolumn{1}{l}{}                     & 59.67\%                                  & \multicolumn{1}{l}{}                     & 55.67\%                                  & 33.69                                 & \multicolumn{1}{l}{}             & 2.01E+04                                     & 2.01E+04                                     \\ \cmidrule(l){2-15} 
                                                                                      & \cellcolor[HTML]{FFCCC9}                                & \cellcolor[HTML]{FFCCC9}1                        & \cellcolor[HTML]{FFCCC9}9.00\%            & \cellcolor[HTML]{FFCCC9}93.00\%           & \cellcolor[HTML]{FFCCC9}8.00\%           & \cellcolor[HTML]{FFCCC9}92.00\%           & \cellcolor[HTML]{FFCCC9}0.00\%           & \cellcolor[HTML]{FFCCC9}72.00\%          & \cellcolor[HTML]{FFCCC9}9.00\%           & \cellcolor[HTML]{FFCCC9}84.00\%          & \cellcolor[HTML]{FFCCC9}2.49E+04      & \cellcolor[HTML]{FFCCC9}7.73E+01 & \cellcolor[HTML]{FFCCC9}7.05E+04             & \cellcolor[HTML]{FFCCC9}7.04E+04             \\
                                                                                      & \cellcolor[HTML]{FFCCC9}                                & \cellcolor[HTML]{FFCCC9}2                        & \cellcolor[HTML]{FFCCC9}13.00\%           & \cellcolor[HTML]{FFCCC9}99.00\%           & \cellcolor[HTML]{FFCCC9}11.00\%          & \cellcolor[HTML]{FFCCC9}99.00\%           & \cellcolor[HTML]{FFCCC9}1.00\%           & \cellcolor[HTML]{FFCCC9}73.00\%          & \cellcolor[HTML]{FFCCC9}3.00\%           & \cellcolor[HTML]{FFCCC9}88.00\%          & \cellcolor[HTML]{FFCCC9}4.49E+04      & \cellcolor[HTML]{FFCCC9}4.85E+02 & \cellcolor[HTML]{FFCCC9}9.17E+03             & \cellcolor[HTML]{FFCCC9}8.69E+03             \\
                                                                                      & \cellcolor[HTML]{FFCCC9}                                & \cellcolor[HTML]{FFCCC9}3                        & \cellcolor[HTML]{FFCCC9}22.00\%           & \cellcolor[HTML]{FFCCC9}100.00\%          & \cellcolor[HTML]{FFCCC9}16.00\%          & \cellcolor[HTML]{FFCCC9}99.00\%           & \cellcolor[HTML]{FFCCC9}0.00\%           & \cellcolor[HTML]{FFCCC9}72.00\%          & \cellcolor[HTML]{FFCCC9}18.00\%          & \cellcolor[HTML]{FFCCC9}88.00\%          & \cellcolor[HTML]{FFCCC9}3.73E+04      & \cellcolor[HTML]{FFCCC9}1.07E+03 & \cellcolor[HTML]{FFCCC9}6.18E+03             & \cellcolor[HTML]{FFCCC9}5.11E+03             \\
                                                                                      & \cellcolor[HTML]{FFCCC9}                                & \cellcolor[HTML]{FFCCC9}4                        & \cellcolor[HTML]{FFCCC9}24.00\%           & \cellcolor[HTML]{FFCCC9}100.00\%          & \cellcolor[HTML]{FFCCC9}17.00\%          & \cellcolor[HTML]{FFCCC9}99.00\%           & \cellcolor[HTML]{FFCCC9}0.00\%           & \cellcolor[HTML]{FFCCC9}74.00\%          & \cellcolor[HTML]{FFCCC9}20.00\%          & \cellcolor[HTML]{FFCCC9}89.00\%          & \cellcolor[HTML]{FFCCC9}3.53E+04      & \cellcolor[HTML]{FFCCC9}1.78E+03 & \cellcolor[HTML]{FFCCC9}1.04E+04             & \cellcolor[HTML]{FFCCC9}8.61E+03             \\
                                                                                      & \multirow{-5}{*}{\cellcolor[HTML]{FFCCC9}\textbf{ROME}} & \cellcolor[HTML]{FFCCC9}5                        & \cellcolor[HTML]{FFCCC9}24.00\%           & \cellcolor[HTML]{FFCCC9}100.00\%          & \cellcolor[HTML]{FFCCC9}17.00\%          & \cellcolor[HTML]{FFCCC9}98.00\%           & \cellcolor[HTML]{FFCCC9}0.00\%           & \cellcolor[HTML]{FFCCC9}74.00\%          & \cellcolor[HTML]{FFCCC9}20.00\%          & \cellcolor[HTML]{FFCCC9}89.00\%          & \cellcolor[HTML]{FFCCC9}3.50E+04      & \cellcolor[HTML]{FFCCC9}2.47E+03 & \multicolumn{1}{l}{\cellcolor[HTML]{FFCCC9}} & \multicolumn{1}{l}{\cellcolor[HTML]{FFCCC9}} \\ \cmidrule(l){2-15} 
                                                                                      &                                                         & 1                                                & 79.50\%                                   & 99.50\%                                   & 76.50\%                                  & 99.00\%                                   & 31.00\%                                  & 74.50\%                                  & 51.50\%                                  & 89.00\%                                  & 41.04                                 & 1.05                             & 23.10                                        & 22.06                                        \\
                                                                                      &                                                         & \cellcolor[HTML]{FFCCC9}2                        & \cellcolor[HTML]{FFCCC9}6.50\%            & \cellcolor[HTML]{FFCCC9}88.00\%           & \cellcolor[HTML]{FFCCC9}6.00\%           & \cellcolor[HTML]{FFCCC9}86.00\%           & \cellcolor[HTML]{FFCCC9}5.50\%           & \cellcolor[HTML]{FFCCC9}80.00\%          & \cellcolor[HTML]{FFCCC9}6.00\%           & \cellcolor[HTML]{FFCCC9}84.50\%          & \cellcolor[HTML]{FFCCC9}4435.72       & \cellcolor[HTML]{FFCCC9}1.07     & \cellcolor[HTML]{FFCCC9}2.38E+04             & \cellcolor[HTML]{FFCCC9}2.38E+04             \\
                                                                                      &                                                         & \cellcolor[HTML]{FFCCC9}3                        & \cellcolor[HTML]{FFCCC9}13.50\%           & \cellcolor[HTML]{FFCCC9}96.00\%           & \cellcolor[HTML]{FFCCC9}11.00\%          & \cellcolor[HTML]{FFCCC9}95.00\%           & \cellcolor[HTML]{FFCCC9}3.50\%           & \cellcolor[HTML]{FFCCC9}70.00\%          & \cellcolor[HTML]{FFCCC9}6.50\%           & \cellcolor[HTML]{FFCCC9}85.00\%          & \cellcolor[HTML]{FFCCC9}120046.73     & \cellcolor[HTML]{FFCCC9}1.68     & \cellcolor[HTML]{FFCCC9}9.45E+03             & \cellcolor[HTML]{FFCCC9}9.45E+03             \\
                                                                                      &                                                         & \cellcolor[HTML]{FFCCC9}4                        & \cellcolor[HTML]{FFCCC9}6.00\%            & \cellcolor[HTML]{FFCCC9}94.00\%           & \cellcolor[HTML]{FFCCC9}4.50\%           & \cellcolor[HTML]{FFCCC9}91.50\%           & \cellcolor[HTML]{FFCCC9}4.50\%           & \cellcolor[HTML]{FFCCC9}72.00\%          & \cellcolor[HTML]{FFCCC9}5.00\%           & \cellcolor[HTML]{FFCCC9}84.50\%          & \cellcolor[HTML]{FFCCC9}34779.75      & \cellcolor[HTML]{FFCCC9}1.65     & \cellcolor[HTML]{FFCCC9}1.74E+04             & \cellcolor[HTML]{FFCCC9}1.74E+04             \\
                                                                                      & \multirow{-5}{*}{\textbf{MEMIT}}                        & \cellcolor[HTML]{FFCCC9}5                        & \cellcolor[HTML]{FFCCC9}6.50\%            & \cellcolor[HTML]{FFCCC9}86.00\%           & \cellcolor[HTML]{FFCCC9}6.00\%           & \cellcolor[HTML]{FFCCC9}83.00\%           & \cellcolor[HTML]{FFCCC9}1.00\%           & \cellcolor[HTML]{FFCCC9}68.00\%          & \cellcolor[HTML]{FFCCC9}2.00\%           & \cellcolor[HTML]{FFCCC9}78.00\%          & \cellcolor[HTML]{FFCCC9}40680.54      & \cellcolor[HTML]{FFCCC9}1.70     & \multicolumn{1}{l}{\cellcolor[HTML]{FFCCC9}} & \multicolumn{1}{l}{\cellcolor[HTML]{FFCCC9}} \\ \cmidrule(l){2-15} 
                                                                                      &                                                         & 1                                                & 90.00\%                                   & 98.67\%                                   & \textbf{83.00\%}                         & 96.33\%                                   & \textbf{66.00\%}                         & \textbf{74.67\%}                         & 77.33\%                                  & \textbf{88.33\%}                         & 34.63                                 & 1.07                             & 7.40                                         & 6.33                                         \\
                                                                                      &                                                         & \cellcolor[HTML]{9AFF99}2                        & \cellcolor[HTML]{9AFF99}\textbf{92.00\%}  & \cellcolor[HTML]{9AFF99}\textbf{99.00\%}  & \cellcolor[HTML]{9AFF99}\textbf{83.33\%} & \cellcolor[HTML]{9AFF99}\textbf{96.67\%}  & \cellcolor[HTML]{9AFF99}\textbf{66.00\%} & \cellcolor[HTML]{9AFF99}\textbf{74.67\%} & \cellcolor[HTML]{9AFF99}\textbf{78.00\%} & \cellcolor[HTML]{9AFF99}\textbf{88.33\%} & \cellcolor[HTML]{9AFF99}34.47         & \cellcolor[HTML]{9AFF99}1.05     & \cellcolor[HTML]{9AFF99}1.12                 & \cellcolor[HTML]{9AFF99}0.07                 \\
                                                                                      &                                                         & 3                                                & 92.33\%                                   & 99.00\%                                   & 84.33\%                                  & 96.67\%                                   & 66.00\%                                  & 74.67\%                                  & 78.67\%                                  & 88.33\%                                  & 34.44                                 & 1.05                             & 1.07                                         & 0.02                                         \\
                                                                                      &                                                         & 4                                                & 93.00\%                                   & 99.00\%                                   & 84.67\%                                  & 96.67\%                                   & 65.67\%                                  & 74.67\%                                  & 78.67\%                                  & 88.33\%                                  & 34.42                                 & 1.05                             & 1.06                                         & 0.00                                         \\
\multirow{-16}{*}{\textbf{\begin{tabular}[c]{@{}l@{}}Llama2\\ (7B)\end{tabular}}}    & \multirow{-5}{*}{\textbf{PMET}}                         & 5                                                & 93.00\%                                   & 99.00\%                                   & 84.67\%                                  & 97.00\%                                   & 66.00\%                                  & 74.67\%                                  & 79.00\%                                  & 88.33\%                                  & 34.44                                 & 1.05                             & \multicolumn{1}{l}{}                         & \multicolumn{1}{l}{}                         \\ \bottomrule
\end{tabular}
}
\caption{Iterative model editing results on ZsRE for at most 10 iterations (denoted by k). We compare the evaluation metrics of iteration that met stopping criterion $|\Delta p_k |\leq 1$ (green rows) to that of their corresponding first iteration and \textbf{bold} the higher value. PMET on GPT-2 XL require more than 5 iterations to achieve our stopping criteria. While ROME is known to collapse (red rows), we observed a unique case of collapse with Llama-2 (7B) specifically when using MEMIT. We discuss this in Section \ref{sec:result}.}
\label{tab:zsre-results}
\end{table*}

\section{Neighbor-assisted model editing results}
\label{sec:neighbor-result}
We present the complete results of all iterations of neighbor-assisted model editing in Table~\ref{tab:neighbor-all}. As the ZsRE dataset was unsuitable for this experiment (see Appendix~\ref{sec:neighbor} for details), results are reported only for the C{\smaller OUNTER}F{\smaller ACT} dataset. Furthermore, since only a subset of samples from C{\smaller OUNTER}F{\smaller ACT} qualified for this experiment, we also include the performance of iterative model editing on these samples for comparison with neighbor-assisted model editing.

Our current implementation of neighbor-assisted model editing does not use prefixes. To analyze its behavior with prefixes, we conducted an additional set of experiments. The results, presented in Table~\ref{tab:nap-neighbor-results-all}, compare prefix-based neighbor-assisted model editing with its prefix-free counterpart and iterative model editing. A detailed analysis is provided in Section~\ref{sec:analysis_nap}.

\begin{table*}[h]
\resizebox{\textwidth}{!}{%
}} & \multirow{-10}{*}{\textbf{NA\_PMET}} & 10                                               & 0.98                                     & 0.98                                      & 0.3                                      & 0.65                                     & 0.83                                     & 0.96                                     & 0.54                                     & 0.84                                     & 67.86                                 & 1.24                            & 1.51                            & 0.27                                 \\ \midrule
                                                                                              & \textbf{Unedited}                    & 0                                                & 0.00\%                                   & 8.00\%                                    & 1.00\%                                   & 10.00\%                                  & \multicolumn{1}{l}{}                     & 100.00\%                                 & \multicolumn{1}{l}{}                     & 12.00\%                                  & 39.80                                 & \multicolumn{1}{l}{}            & \multicolumn{1}{l}{}            & \multicolumn{1}{l}{}                 \\ \cmidrule(l){2-15} 
                                                                                              &                                      & 1                                                & 99.00\%                                  & 100.00\%                                  & 71.00\%                                  & 93.00\%                                  & 67.00\%                                  & 86.00\%                                  & 77.00\%                                  & 93.00\%                                  & 42.79                                 & 1.03                            & 4.24                            & 3.21                                 \\
                                                                                              &                                      & \cellcolor[HTML]{9AFF99}2                        & \cellcolor[HTML]{9AFF99}\textbf{99.00\%} & \cellcolor[HTML]{9AFF99}\textbf{100.00\%} & \cellcolor[HTML]{9AFF99}\textbf{79.00\%} & \cellcolor[HTML]{9AFF99}\textbf{97.00\%} & \cellcolor[HTML]{9AFF99}63.00\%          & \cellcolor[HTML]{9AFF99}83.00\%          & \cellcolor[HTML]{9AFF99}78.00\%          & \cellcolor[HTML]{9AFF99}93.00\%          & \cellcolor[HTML]{9AFF99}44.84         & \cellcolor[HTML]{9AFF99}1.02    & \cellcolor[HTML]{9AFF99}1.66    & \cellcolor[HTML]{9AFF99}0.64         \\
                                                                                              &                                      & 3                                                & 99.00\%                                  & 100.00\%                                  & 79.00\%                                  & 98.00\%                                  & 62.00\%                                  & 83.00\%                                  & 77.00\%                                  & 93.00\%                                  & 48.58                                 & 1.01                            & 17.10                           & 16.09                                \\
                                                                                              &                                      & 4                                                & 100.00\%                                 & 100.00\%                                  & 79.00\%                                  & 98.00\%                                  & 59.00\%                                  & 82.00\%                                  & 76.00\%                                  & 93.00\%                                  & 49.43                                 & 1.01                            & 1.74                            & 0.73                                 \\
                                                                                              & \multirow{-5}{*}{\textbf{MEMIT}}     & 5                                                & 100.00\%                                 & 100.00\%                                  & 81.00\%                                  & 98.00\%                                  & 58.00\%                                  & 81.00\%                                  & 76.00\%                                  & 92.00\%                                  & 50.50                                 & 1.01                            & \multicolumn{1}{l}{}            & \multicolumn{1}{l}{}                 \\ \cmidrule(l){2-15} 
                                                                                              &                                      & 1                                                & 98.00\%                                  & 99.00\%                                   & 63.00\%                                  & 82.00\%                                  & 84.00\%                                  & 95.00\%                                  & 79.00\%                                  & 91.00\%                                  & 42.71                                 & 1.04                            & 2.56                            & 1.52                                 \\
                                                                                              &                                      & \cellcolor[HTML]{9AFF99}2                        & \cellcolor[HTML]{9AFF99}\textbf{99.00\%} & \cellcolor[HTML]{9AFF99}99.00\%           & \cellcolor[HTML]{9AFF99}75.00\%          & \cellcolor[HTML]{9AFF99}92.00\%          & \cellcolor[HTML]{9AFF99}\textbf{81.00\%} & \cellcolor[HTML]{9AFF99}\textbf{95.00\%} & \cellcolor[HTML]{9AFF99}\textbf{84.00\%} & \cellcolor[HTML]{9AFF99}\textbf{95.00\%} & \cellcolor[HTML]{9AFF99}45.24         & \cellcolor[HTML]{9AFF99}1.02    & \cellcolor[HTML]{9AFF99}1.35    & \cellcolor[HTML]{9AFF99}0.33         \\
                                                                                              &                                      & 3                                                & 99.00\%                                  & 100.00\%                                  & 74.00\%                                  & 91.00\%                                  & 83.00\%                                  & 95.00\%                                  & 84.00\%                                  & 95.00\%                                  & 47.19                                 & 1.02                            & 3.29                            & 2.27                                 \\
                                                                                              &                                      & 4                                                & 99.00\%                                  & 100.00\%                                  & 74.00\%                                  & 90.00\%                                  & 80.00\%                                  & 95.00\%                                  & 83.00\%                                  & 95.00\%                                  & 49.77                                 & 1.02                            & 1.43                            & 0.41                                 \\
                                                                                              & \multirow{-5}{*}{\textbf{NA\_MEMIT}} & 5                                                & 100.00\%                                 & 100.00\%                                  & 73.00\%                                  & 91.00\%                                  & 79.00\%                                  & 95.00\%                                  & 82.00\%                                  & 95.00\%                                  & 50.94                                 & 1.02                            & \multicolumn{1}{l}{}            & \multicolumn{1}{l}{}                 \\ \cmidrule(l){2-15} 
                                                                                              &                                      & \cellcolor[HTML]{9AFF99}1                        & \cellcolor[HTML]{9AFF99}\textbf{99.00\%} & \cellcolor[HTML]{9AFF99}\textbf{100.00\%} & \cellcolor[HTML]{9AFF99}\textbf{72.00\%} & \cellcolor[HTML]{9AFF99}\textbf{93.00\%} & \cellcolor[HTML]{9AFF99}65.00\%          & \cellcolor[HTML]{9AFF99}84.00\%          & \cellcolor[HTML]{9AFF99}76.00\%          & \cellcolor[HTML]{9AFF99}92.00\%          & \cellcolor[HTML]{9AFF99}40.79         & \cellcolor[HTML]{9AFF99}1.06    & \cellcolor[HTML]{9AFF99}1.31    & \cellcolor[HTML]{9AFF99}0.25         \\
                                                                                              &                                      & 2                                                & 99.00\%                                  & 100.00\%                                  & 73.00\%                                  & 93.00\%                                  & 65.00\%                                  & 84.00\%                                  & 77.00\%                                  & 92.00\%                                  & 40.90                                 & 1.06                            & 219.70                          & 218.64                               \\
                                                                                              &                                      & 3                                                & 99.00\%                                  & 100.00\%                                  & 73.00\%                                  & 93.00\%                                  & 65.00\%                                  & 84.00\%                                  & 77.00\%                                  & 92.00\%                                  & 40.92                                 & 1.06                            & 2.25                            & 1.19                                 \\
                                                                                              &                                      & 4                                                & 99.00\%                                  & 100.00\%                                  & 73.00\%                                  & 94.00\%                                  & 65.00\%                                  & 84.00\%                                  & 77.00\%                                  & 92.00\%                                  & 41.08                                 & 1.05                            & 1.06                            & 0.01                                 \\
                                                                                              & \multirow{-5}{*}{\textbf{PMET}}      & 5                                                & 99.00\%                                  & 100.00\%                                  & 74.00\%                                  & 94.00\%                                  & 65.00\%                                  & 84.00\%                                  & 77.00\%                                  & 92.00\%                                  & 41.28                                 & 1.05                            & \multicolumn{1}{l}{}            & \multicolumn{1}{l}{}                 \\ \cmidrule(l){2-15} 
                                                                                              &                                      & 1                                                & 99.00\%                                  & 100.00\%                                  & 72.00\%                                  & 91.00\%                                  & 75.00\%                                  & 90.00\%                                  & 80.00\%                                  & 93.00\%                                  & 41.01                                 & 1.06                            & 2.38                            & 1.32                                 \\
                                                                                              &                                      & 2                                                & 98.00\%                                  & 99.00\%                                   & 71.00\%                                  & 90.00\%                                  & 82.00\%                                  & 94.00\%                                  & 82.00\%                                  & 94.00\%                                  & 41.23                                 & 1.12                            & 4.91                            & 3.79                                 \\
                                                                                              &                                      & 3                                                & 99.00\%                                  & 99.00\%                                   & 70.00\%                                  & 89.00\%                                  & 83.00\%                                  & 95.00\%                                  & 82.00\%                                  & 94.00\%                                  & 41.66                                 & 1.13                            & 24.94                           & 23.81                                \\
                                                                                              &                                      & 4                                                & 97.00\%                                  & 98.00\%                                   & 68.00\%                                  & 88.00\%                                  & 84.00\%                                  & 95.00\%                                  & 82.00\%                                  & 93.00\%                                  & 41.99                                 & 1.13                            & 4.13                            & 3.00                                 \\
                                                                                              &                                      & 5                                                & 98.00\%                                  & 98.00\%                                   & 69.00\%                                  & 89.00\%                                  & 81.00\%                                  & 94.00\%                                  & 81.00\%                                  & 94.00\%                                  & 42.71                                 & 1.12                            & 273.05                          & 271.93                               \\
                                                                                              &                                      & \cellcolor[HTML]{9AFF99}6                        & \cellcolor[HTML]{9AFF99}98.00\%          & \cellcolor[HTML]{9AFF99}99.00\%           & \cellcolor[HTML]{9AFF99}69.00\%          & \cellcolor[HTML]{9AFF99}89.00\%          & \cellcolor[HTML]{9AFF99}\textbf{80.00\%} & \cellcolor[HTML]{9AFF99}\textbf{94.00\%} & \cellcolor[HTML]{9AFF99}\textbf{81.00\%} & \cellcolor[HTML]{9AFF99}\textbf{94.00\%} & \cellcolor[HTML]{9AFF99}43.54         & \cellcolor[HTML]{9AFF99}1.12    & \cellcolor[HTML]{9AFF99}1.56    & \cellcolor[HTML]{9AFF99}0.44         \\
                                                                                              &                                      & 7                                                & 99.00\%                                  & 99.00\%                                   & 68.00\%                                  & 89.00\%                                  & 76.00\%                                  & 93.00\%                                  & 79.00\%                                  & 94.00\%                                  & 44.62                                 & 1.13                            & 1.57                            & 0.44                                 \\
                                                                                              &                                      & 8                                                & 98.00\%                                  & 99.00\%                                   & 67.00\%                                  & 88.00\%                                  & 75.00\%                                  & 92.00\%                                  & 78.00\%                                  & 93.00\%                                  & 45.70                                 & 1.13                            & 1.64                            & 0.51                                 \\
                                                                                              &                                      & 9                                                & 98.00\%                                  & 99.00\%                                   & 67.00\%                                  & 88.00\%                                  & 72.00\%                                  & 92.00\%                                  & 77.00\%                                  & 93.00\%                                  & 47.48                                 & 1.14                            & 2.11                            & 0.97                                 \\
\multirow{-26}{*}{\begin{tabular}[c]{@{}l@{}}GPT-J\\ (6B)\\ \#960\end{tabular}}               & \multirow{-10}{*}{\textbf{NA\_PMET}} & 10                                               & 99.00\%                                  & 99.00\%                                   & 68.00\%                                  & 88.00\%                                  & 70.00\%                                  & 91.00\%                                  & 76.00\%                                  & 93.00\%                                  & 50.68                                 & 1.14                            & \multicolumn{1}{l}{}            & -1.14                                \\ \midrule
                                                                                              & \textbf{Unedited}                    & 0                                                & 38.33\%                                  & 57.00\%                                   & 37.00\%                                  & 56.00\%                                  & \multicolumn{1}{l}{}                     & 59.67\%                                  & \multicolumn{1}{l}{}                     & 55.67\%                                  & 33.69                                 & \multicolumn{1}{l}{}            & \multicolumn{1}{l}{}            & \multicolumn{1}{l}{}                 \\ \cmidrule(l){2-15} 
                                                                                              &                                      & 1                                                & 94.00\%                                  & 97.00\%                                   & 70.00\%                                  & 87.00\%                                  & 71.00\%                                  & 92.00\%                                  & 77.00\%                                  & 92.00\%                                  & 30.95                                 & 1.09                            & 2.29                            & 1.2                                  \\
                                                                                              &                                      & 2                                                & 96.00\%                                  & 98.00\%                                   & 72.00\%                                  & 89.00\%                                  & 71.00\%                                  & 92.00\%                                  & 78.00\%                                  & 93.00\%                                  & 31.03                                 & 1.14                            & 25.72                           & 24.58                                \\
                                                                                              &                                      & \cellcolor[HTML]{9AFF99}3                        & \cellcolor[HTML]{9AFF99}96.00\%          & \cellcolor[HTML]{9AFF99}98.00\%           & \cellcolor[HTML]{9AFF99}72.00\%          & \cellcolor[HTML]{9AFF99}89.00\%          & \cellcolor[HTML]{9AFF99}70.00\%          & \cellcolor[HTML]{9AFF99}92.00\%          & \cellcolor[HTML]{9AFF99}78.00\%          & \cellcolor[HTML]{9AFF99}93.00\%          & \cellcolor[HTML]{9AFF99}31            & \cellcolor[HTML]{9AFF99}1.14    & \cellcolor[HTML]{9AFF99}1.58    & \cellcolor[HTML]{9AFF99}0.44         \\
                                                                                              &                                      & 4                                                & 96.00\%                                  & 98.00\%                                   & 72.00\%                                  & 89.00\%                                  & 71.00\%                                  & 92.00\%                                  & 78.00\%                                  & 93.00\%                                  & 31.01                                 & 1.14                            & 1.14                            & 0                                    \\
                                                                                              & \multirow{-5}{*}{\textbf{PMET}}      & 5                                                & 96.00\%                                  & 98.00\%                                   & 72.00\%                                  & 89.00\%                                  & 71.00\%                                  & 92.00\%                                  & 78.00\%                                  & 93.00\%                                  & 31                                    & 1.14                            & \multicolumn{1}{l}{}            & \multicolumn{1}{l}{}                 \\ \cmidrule(l){2-15} 
                                                                                              &                                      & 1                                                & 92.00\%                                  & 95.00\%                                   & 67.00\%                                  & 80.00\%                                  & 76.00\%                                  & 94.00\%                                  & 77.00\%                                  & 89.00\%                                  & 30.43                                 & 1.14                            & 15.27                           & 14.13                                \\
                                                                                              &                                      & 2                                                & 96.00\%                                  & 97.00\%                                   & 75.00\%                                  & 87.00\%                                  & 73.00\%                                  & 93.00\%                                  & 80.00\%                                  & 92.00\%                                  & 30.52                                 & 1.02                            & 2.86                            & 1.84                                 \\
                                                                                              &                                      & 3                                                & 96.00\%                                  & 96.00\%                                   & 76.00\%                                  & 87.00\%                                  & 71.00\%                                  & 94.00\%                                  & 80.00\%                                  & 92.00\%                                  & 31.19                                 & 1.04                            & 2.61                            & 1.57                                 \\
                                                                                              &                                      & 4                                                & 95.00\%                                  & 95.00\%                                   & 76.00\%                                  & 86.00\%                                  & 66.00\%                                  & 92.00\%                                  & 77.00\%                                  & 91.00\%                                  & 32.19                                 & 1.04                            & 2.27                            & 1.23                                 \\
                                                                                              &                                      & 5                                                & 94.00\%                                  & 94.00\%                                   & 75.00\%                                  & 85.00\%                                  & 60.00\%                                  & 91.00\%                                  & 74.00\%                                  & 90.00\%                                  & 35.28                                 & 1.05                            & 186                             & 184.95                               \\
                                                                                              &                                      & 6                                                & 90.00\%                                  & 91.00\%                                   & 74.00\%                                  & 84.00\%                                  & 55.00\%                                  & 89.00\%                                  & 70.00\%                                  & 88.00\%                                  & 40.1                                  & 1.06                            & 13.09                           & 12.03                                \\
                                                                                              &                                      & 7                                                & 87.00\%                                  & 88.00\%                                   & 70.00\%                                  & 81.00\%                                  & 47.00\%                                  & 86.00\%                                  & 64.00\%                                  & 85.00\%                                  & 53.95                                 & 1.06                            & 30.27                           & 29.21                                \\
                                                                                              &                                      & 8                                                & 80.00\%                                  & 83.00\%                                   & 66.00\%                                  & 80.00\%                                  & 42.00\%                                  & 83.00\%                                  & 58.00\%                                  & 82.00\%                                  & 104.98                                & 1.06                            & 84.27                           & 83.21                                \\
                                                                                              &                                      & 9                                                & 73.00\%                                  & 83.00\%                                   & 58.00\%                                  & 78.00\%                                  & 30.00\%                                  & 79.00\%                                  & 47.00\%                                  & 80.00\%                                  & 373.87                                & 1.07                            & 358.05                          & 356.98                               \\
\multirow{-16}{*}{\textbf{\begin{tabular}[c]{@{}l@{}}Llama 2\\ (7B)\\ \#1340\end{tabular}}}   & \multirow{-10}{*}{\textbf{NA\_PMET}} & 10                                               & 62.00\%                                  & 79.00\%                                   & 47.00\%                                  & 75.00\%                                  & 26.00\%                                  & 76.00\%                                  & 40.00\%                                  & 76.00\%                                  & 1414.01                               & 1.07                            & \multicolumn{1}{l}{}            & \multicolumn{1}{l}{}                 \\ \bottomrule
\end{tabular}
}
\caption{Neighbor-Assisted model editing results on C{\smaller OUNTER}F{\smaller ACT}. We compare evaluation metrics for both neighbor-assisted (NA\_) and without neighbor runs of the model editing algorithms where $|\Delta p_k |\leq 1$ (green rows)
and \textbf{bold} the higher value. Results among models and from Table~\ref{tab:mcf-results-all} are not comparable due to difference in neighboring samples (Appendix.~\ref{sec:neighbor}). Hence, we report the no. of examples (\#) used to run experiment for each model. NA\_PMET on Llama-2 (7B) stands as an exception that didn't achieved the stopping criteria within 10 iteration and showed a performance decrease.
} 
\label{tab:neighbor-all}
\end{table*}

\begin{table*}[h]
\resizebox{\textwidth}{!}{%
}} & \multirow{-10}{*}{\textbf{NAP\_PMET}}          & 10                                               & 98.00\%                                  & 99.00\%                                  & 30.00\%                                  & 65.00\%                                  & 82.00\%                                  & 94.00\%                                  & 53.00\%                                  & 83.00\%                                  & 64.04                                 & 1.21                            & 1.47                            & 0.26                                 \\ \bottomrule
\end{tabular}
}
\caption{Results of prefix-free (NA\_) and with prefix(NAP\_) neighbor-assisted model editing on C{\smaller OUNTER}F{\smaller ACT}. We compare their evaluation metrics when our stopping criteria $|\Delta p_k |\leq 1$ (green rows) is met and \textbf{bold} the higher value. Results among models and from Table~\ref{tab:mcf-results-all} are not comparable due to difference in neighboring samples as explained in Appendix~\ref{sec:neighbor}. Hence, we report the no. of examples (\#) used to run experiment for each model.} 
\label{tab:nap-neighbor-results-all}
\end{table*}

\begin{table*}[h]
\resizebox{\textwidth}{!}{%
\begin{tabular}{@{}llrrrrrrrrr@{}}
\toprule
\multicolumn{2}{l}{\textbf{Dataset}}                                                                                     & \multicolumn{1}{l}{\textbf{}}  & \multicolumn{4}{l}{\textbf{C{\smaller OUNTER}F{\smaller ACT}}}                                                                                                                                                      & \multicolumn{4}{l}{\textbf{ZsRE}}                                                                                                                                         \\ \midrule
                                                                                      &                                  & \multicolumn{1}{l}{\textbf{k}} & \multicolumn{1}{l}{\textbf{Score ($\uparrow$)}}       & \multicolumn{1}{l}{\textbf{}}                                & \multicolumn{1}{l}{\textbf{$|\Delta p_k|$ ($\downarrow$)}}      & \multicolumn{1}{l}{\textbf{$\Delta_{\text{p2}}$ ($\downarrow$)}}            & \multicolumn{2}{l}{\textbf{Score ($\uparrow$)}}                                                  & \multicolumn{1}{l}{\textbf{$|\Delta p_k|$ ($\downarrow$)}}      & \multicolumn{1}{l}{\textbf{$\Delta_{p2}$ ($\downarrow$)}}            \\ \cmidrule(l){3-11} 
\multirow{-2}{*}{\textbf{Model}}                                                      & \multirow{-2}{*}{\textbf{Algo}}  & \multicolumn{1}{l}{\textbf{}}  & \multicolumn{1}{l}{\textbf{Accuracy}}    & \multicolumn{1}{l}{\textbf{Success}}                         & \multicolumn{1}{l}{\textbf{}} & \multicolumn{1}{l}{\textbf{}} & \multicolumn{1}{l}{\textbf{Accuracy}}    & \multicolumn{1}{l}{\textbf{Success}}     & \multicolumn{1}{l}{\textbf{}} & \multicolumn{1}{l}{\textbf{}} \\ \midrule
                                                                                      & \textbf{Unedited}                & 0                              & \multicolumn{1}{l}{}                     & 31.33\%                                                      & \multicolumn{1}{l}{}                 & \multicolumn{1}{l}{}                         & \multicolumn{1}{l}{}                     & 73.67\%                                  & \multicolumn{1}{l}{}                 & \multicolumn{1}{l}{}                         \\ \cmidrule(l){2-11} 
                                                                                      &                                  & 1                              & 42.67\%                                  & 83.00\%                                                      & 11359.60                             & \multicolumn{1}{l}{}                         & 48.33\%                                  & 94.00\%                                  & 2034.31                              & \multicolumn{1}{l}{}                         \\
                                                                                      &                                  & 2                              & 55.67\%                                  & 89.67\%                                                      & 77.12                                & 1.13E+04                                     & 47.67\%                                  & 94.00\%                                  & 39.39                                & 1994.93                                      \\
                                                                                      &                                  & 3                              & 56.67\%                                  & 90.00\%                                                      & 9.11                                 & 68.01                                        & \cellcolor[HTML]{9AFF99}\textbf{46.67\%} & \cellcolor[HTML]{9AFF99}\textbf{94.00\%} & \cellcolor[HTML]{9AFF99}0.03         & 39.36                                        \\
                                                                                      &                                  & 4                              & \cellcolor[HTML]{9AFF99}\textbf{56.67\%} & \cellcolor[HTML]{9AFF99}\textbf{90.00\%}                     & \cellcolor[HTML]{9AFF99}0.47         & 8.65                                         & \cellcolor[HTML]{CBCEFB}45.00\%          & \cellcolor[HTML]{CBCEFB}94.00\%          & 0.01                                 & \cellcolor[HTML]{CBCEFB}0.02                 \\
                                                                                      & \multirow{-5}{*}{\textbf{MEMIT}} & 5                              & 56.67\%                                  & 90.00\%                                                      & \multicolumn{1}{l}{}                 & \multicolumn{1}{l}{}                         & 44.33\%                                  & 94.00\%                                  & \multicolumn{1}{l}{}                 & \multicolumn{1}{l}{}                         \\ \cmidrule(l){2-11} 
                                                                                      &                                  & 1                              & 8.33\%                                   & 58.33\%                                                      & 103785.71                            & \multicolumn{1}{l}{}                         & 35.33\%                                  & 93.00\%                                  & 111052.45                            & \multicolumn{1}{l}{}                         \\
                                                                                      &                                  & 2                              & 31.67\%                                  & 77.33\%                                                      & 25519.43                             & 8.95E+04                                     & 49.67\%                                  & 94.33\%                                  & 3094.74                              & 113317.73                                    \\
                                                                                      &                                  & 3                              & 40.67\%                                  & 83.33\%                                                      & 5025.56                              & 2.17E+04                                     & 55.00\%                                  & 94.33\%                                  & 489.28                               & 2618.47                                      \\
                                                                                      &                                  & 4                              & 43.33\%                                  & 85.33\%                                                      & 2029.80                              & 3077.3                                       & 54.67\%                                  & 94.00\%                                  & 48.69                                & 441.40                                       \\
                                                                                      &                                  & 5                              & 44.00\%                                  & 86.67\%                                                      & 224.53                               & 1805.25                                      & 54.00\%                                  & 94.00\%                                  & 1.45                                 & 47.33                                        \\
                                                                                      &                                  & \multicolumn{1}{l}{6}          & 44.67\%                                  & \multicolumn{1}{l}{86.67\%}                                  & 48.96                                & 175.57                                       & \cellcolor[HTML]{9AFF99}\textbf{54.00\%} & \cellcolor[HTML]{9AFF99}\textbf{94.00\%} & \cellcolor[HTML]{9AFF99}0.19         & 1.29                                         \\
                                                                                      &                                  & \multicolumn{1}{l}{7}          & 45.67\%                                  & \multicolumn{1}{l}{87.00\%}                                  & 21.02                                & 27.94                                        & \cellcolor[HTML]{CBCEFB}53.33\%          & \cellcolor[HTML]{CBCEFB}\textbf{94.00\%} & 0.08                                 & \cellcolor[HTML]{CBCEFB}0.11                 \\
                                                                                      &                                  & \multicolumn{1}{l}{8}          & 46.00\%                                  & \multicolumn{1}{l}{87.00\%}                                  & 10.75                                & 10.27                                        & 53.00\%                                  & 94.00\%                                  & 0.04                                 & 0.05                                         \\
                                                                                      &                                  & \multicolumn{1}{l}{9}          & 46.00\%                                  & \multicolumn{1}{l}{87.00\%}                                  & 1.71                                 & 9.04                                         & 52.33\%                                  & 94.00\%                                  & 0.01                                 & 0.03                                         \\
\multirow{-16}{*}{\textbf{\begin{tabular}[c]{@{}l@{}}GPT-2 XL\\ (1.5B)\end{tabular}}} & \multirow{-10}{*}{\textbf{PMET}} & \multicolumn{1}{l}{10}         & \cellcolor[HTML]{9AFF99}\textbf{47.00\%} & \multicolumn{1}{l}{\cellcolor[HTML]{9AFF99}\textbf{87.33\%}} & \cellcolor[HTML]{9AFF99}0.42         & 1.29                                         & 52.33\%                                  & 94.00\%                                  & 0.01                                 & 0.01                                         \\ \midrule
                                                                                      & \textbf{Unedited}                & 0                              & \multicolumn{1}{l}{}                     & 37.33\%                                                      & \multicolumn{1}{l}{}                 & \multicolumn{1}{l}{}                         & \multicolumn{1}{l}{}                     & 77.33\%                                  & 5.02E+04                             & \multicolumn{1}{l}{}                         \\ \cmidrule(l){2-11} 
                                                                                      &                                  & 1                              & 77.67\%                                  & 94.33\%                                                      & 1.22                                 & \multicolumn{1}{l}{}                         & 74.67\%                                  & 92.00\%                                  & 1.67                                 & \multicolumn{1}{l}{}                         \\
                                                                                      &                                  & 2                              & \cellcolor[HTML]{9AFF99}\textbf{79.00\%} & \cellcolor[HTML]{9AFF99}\textbf{95.00\%}                     & \cellcolor[HTML]{9AFF99}0.03         & 1.20                                         & \cellcolor[HTML]{9AFF99}74.67\%          & \cellcolor[HTML]{9AFF99}\textbf{92.33\%} & \cellcolor[HTML]{9AFF99}0.01         & 1.65                                         \\
                                                                                      &                                  & 3                              & 79.00\%                                  & 95.00\%                                                      & 1.86                                 & 1.83                                         & \cellcolor[HTML]{CBCEFB}\textbf{75.00\%} & \cellcolor[HTML]{CBCEFB}\textbf{92.33\%} & 0.00                                 & \cellcolor[HTML]{CBCEFB}0.02                 \\
                                                                                      &                                  & 4                              & 79.33\%                                  & 95.00\%                                                      & 0.03                                 & 1.84                                         & 75.00\%                                  & 92.33\%                                  & 0.00                                 & 0.00                                         \\
                                                                                      & \multirow{-5}{*}{\textbf{MEMIT}} & 5                              & 79.33\%                                  & 95.00\%                                                      & \multicolumn{1}{l}{}                 & \multicolumn{1}{l}{}                         & 75.00\%                                  & 92.33\%                                  & \multicolumn{1}{l}{}                 & \multicolumn{1}{l}{}                         \\ \cmidrule(l){2-11} 
                                                                                      &                                  & 1                              & \cellcolor[HTML]{FFCE93}77.67\%          & \cellcolor[HTML]{FFCE93}93.67\%                              & \cellcolor[HTML]{FFCE93}1.15         & \multicolumn{1}{l}{}                         & 72.67\%                                  & 92.00\%                                  & 4.10                                 & \multicolumn{1}{l}{}                         \\
                                                                                      &                                  & 2                              & 78.00\%                                  & 94.33\%                                                      & 4.23                                 & 3.05                                         & \cellcolor[HTML]{9AFF99}74.00\%          & \cellcolor[HTML]{9AFF99}\textbf{92.00\%} & \cellcolor[HTML]{9AFF99}0.09         & 4.06                                         \\
                                                                                      &                                  & 3                              & \cellcolor[HTML]{9AFF99}\textbf{78.33\%} & \cellcolor[HTML]{9AFF99}\textbf{94.00\%}                     & \cellcolor[HTML]{9AFF99}0.05         & 4.18                                         & \cellcolor[HTML]{CBCEFB}\textbf{74.33\%} & \cellcolor[HTML]{CBCEFB}\textbf{92.00\%} & 0.02                                 & \cellcolor[HTML]{CBCEFB}0.07                 \\
                                                                                      &                                  & 4                              & \cellcolor[HTML]{CBCEFB}78.33\%          & \cellcolor[HTML]{CBCEFB}94.00\%                              & 0.02                                 & \cellcolor[HTML]{CBCEFB}0.03                 & 74.33\%                                  & 92.00\%                                  & 0.00                                 & 0.01                                         \\
\multirow{-11}{*}{\textbf{\begin{tabular}[c]{@{}l@{}}GPT-J\\ (6B)\end{tabular}}}      & \multirow{-5}{*}{\textbf{PMET}}  & 5                              & 78.33\%                                  & 94.33\%                                                      & \multicolumn{1}{l}{}                 & \multicolumn{1}{l}{}                         & 74.33\%                                  & 92.00\%                                  & \multicolumn{1}{l}{}                 & \multicolumn{1}{l}{}                         \\ \midrule
                                                                                      & \textbf{Unedited}                & 0                              & \multicolumn{1}{l}{}                     & 19.67\%                                                      & \multicolumn{1}{l}{}                 & \multicolumn{1}{l}{}                         & \multicolumn{1}{l}{}                     & 55.67\%                                  & 2.01E+04                             & \multicolumn{1}{l}{}                         \\ \cmidrule(l){2-11} 
                                                                                      &                                  & 1                              & 77.33\%                                  & 90.33\%                                                      & 3.32                                 & \multicolumn{1}{l}{}                         & 77.33\%                                  & 88.33\%                                  & 6.33                                 & \multicolumn{1}{l}{}                         \\
                                                                                      &                                  & 2                              & \cellcolor[HTML]{9AFF99}78.00\%          & \cellcolor[HTML]{9AFF99}\textbf{91.67\%}                     & \cellcolor[HTML]{9AFF99}0.09         & 3.18                                         & \cellcolor[HTML]{9AFF99}78.00\%          & \cellcolor[HTML]{9AFF99}\textbf{88.33\%} & \cellcolor[HTML]{9AFF99}0.07         & 6.28                                         \\
                                                                                      &                                  & 3                              & \cellcolor[HTML]{CBCEFB}\textbf{78.33\%} & \cellcolor[HTML]{CBCEFB}\textbf{91.67\%}                     & 0.16                                 & \cellcolor[HTML]{CBCEFB}0.07                 & \cellcolor[HTML]{CBCEFB}\textbf{78.67\%} & \cellcolor[HTML]{CBCEFB}\textbf{88.33\%} & 0.02                                 & \cellcolor[HTML]{CBCEFB}0.05                 \\
                                                                                      &                                  & 4                              & 78.33\%                                  & 91.67\%                                                      & 0.00                                 & 0.16                                         & 78.67\%                                  & 88.33\%                                  & 0.00                                 & 0.01                                         \\
\multirow{-6}{*}{\textbf{\begin{tabular}[c]{@{}l@{}}Llama-2\\ (7B)\end{tabular}}}     & \multirow{-5}{*}{\textbf{PMET}}  & 5                              & 78.33\%                                  & 91.67\%                                                      & \multicolumn{1}{l}{}                 & \multicolumn{1}{l}{}                         & 79.00\%                                  & 88.33\%                                  & \multicolumn{1}{l}{}                 & \multicolumn{1}{l}{}                         \\ \bottomrule
\end{tabular}
}
\caption{Comparing stopping criteria. We compare our proposed stopping criteria $|\Delta p_k|\le 1$ (green) to the two alternate stopping criteria, monotonic decrease i.e. $|\Delta p_{k+1}| < |\Delta p_{k}|$, otherwise stop and use $\theta_k$ (orange), and small change, i.e., $\Delta_{p2} = |p(\theta_{k+1}, h^{l_c}_{t,k+1}) - p(\theta_{k}, h^{l_c}_{t,k})| \leq 1$ (purple). We \textbf{bold} the higher scores among them.
}
\label{tab:stopcompare}
\end{table*}

\end{document}